\documentclass[10pt]{article}
\usepackage[utf8]{inputenc}
\usepackage{fullpage}
\usepackage{amsmath}
\usepackage{amsthm}
\usepackage{amsfonts}
\usepackage{amssymb}
\usepackage{graphicx}
\usepackage[pdftex]{hyperref}
\usepackage{color}
\usepackage{authblk}
\usepackage{algorithm}
\usepackage{algorithmic}
\usepackage{multirow}
\usepackage{array}
\usepackage{todonotes}
\usepackage{soul}
\usepackage{subfigure}
\usepackage[numbers,sort&compress]{natbib}
\usepackage[titletoc,title]{appendix}

\theoremstyle{definition}
\newtheorem{defn}{Definition}
\newtheorem*{remark}{Remark}

\DeclareMathOperator{\sgn}{sgn}

\definecolor{darkgreen}{rgb}{0,0.6,0.2}

\title{Interpreting recurrent neural networks behaviour via excitable network attractors}

\author[1]{Andrea Ceni\thanks{ac860@exeter.ac.uk}}
\author[2]{Peter Ashwin\thanks{p.ashwin@exeter.ac.uk}}
\author[3,1]{Lorenzo Livi\thanks{lorenz.livi@gmail.com}\thanks{Corresponding author}}
\affil[1]{Department of Computer Science, University of Exeter, Exeter EX4 4QF, UK}
\affil[2]{Department of Mathematics, University of Exeter, Exeter EX4 4QF, UK}
\affil[3]{Departments of Computer Science and Mathematics, University of Manitoba, Winnipeg, MB R3T 2N2, Canada}

\providecommand{\keywords}[1]{\textbf{\textit{Keywords---}} #1}

\begin{document}

\maketitle

\begin{abstract}
\textbf{Introduction:} Machine learning provides fundamental tools both for scientific research and for the development of technologies with significant impact on society.
It provides methods that facilitate the discovery of regularities in data and that give predictions without explicit knowledge of the rules governing a system.
However, a price is paid for exploiting such flexibility: machine learning methods are typically black-boxes where it is difficult to fully understand what the machine is doing or how it is operating. This poses constraints on the applicability and explainability of such methods.
\textbf{Methods:} Our research aims to open the black-box of recurrent neural networks, an important family of neural networks used for processing sequential data. We propose a novel methodology that provides a mechanistic interpretation of behaviour when solving a computational task.
Our methodology uses mathematical constructs called excitable network attractors, which are invariant sets in phase space composed of stable attractors and excitable connections between them.
\textbf{Results and Discussion:} As the behaviour of recurrent neural networks depends both on training and on inputs to the system, we introduce an algorithm to extract network attractors directly from the trajectory of a neural network while solving tasks.
Simulations conducted on a controlled benchmark task confirm the relevance of these attractors for interpreting the behaviour of recurrent neural networks, at least for tasks that involve learning a finite number of stable states and transitions between them.\\
\end{abstract}

\section{Introduction}

Artificial recurrent neural networks (RNNs) are widely used to solve tasks involving temporal data, e.g., speech \cite{graves2013speech} and handwriting recognition \cite{6981034}, audio classification \cite{scardapane2017semi,keuninckx2017real} or time series forecasting \cite{bianchi2015prediction}.
RNNs are characterised by the presence of feedback connections in a hidden layer, which allows generating a state-space representation that equips the network with short-term memory capability.
RNNs are universal approximators of dynamical systems \cite{hammer2000approximation,funahashi1993approximation} meaning that, given enough neurons in the hidden layer, it is possible to fine tune the weights to achieve any desired level of accuracy.
Nevertheless, training via back-propagation through time is difficult due to the vanishing/exploding gradient problem \cite{pascanu2013difficulty,kanai2017preventing}. This has led to the development of new and faster techniques for training RNNs, including a different paradigm known as reservoir computing \cite{lukovsevivcius2009reservoir,maass2007computational}.
Echo state networks (ESNs) \cite{jaeger2004harnessing,lokse2017training} constitute an important example of reservoir computing, where a recurrent layer (called a reservoir) is composed of a large number of neurons with randomly initialised connections that are not fine-tuned via gradient-based optimisation mechanisms.
The main idea behind ESNs is to exploit the rich dynamics generated by the reservoir with an output layer, the read-out that is optimised to solve a specific task.

\subsection{Problem statement and research hypothesis}

The high-dimensional and non-linear nature of RNNs complicates interpretability of their internal dynamics, which are characterised by complex, input-dependent spatio-temporal patterns of activity \cite{sussillo2013opening,rajan2010stimulus}.
This poses constraints on understanding the behaviour of RNNs: they are usually viewed as black-boxes from which it is hard to extract useful knowledge about their inner workings. As highlighted by recent research efforts \cite{7959606,montavon2017methods,castelvecchi2016can}, similar interpretability issues affect many other machine learning methods. Furthermore, an increasing societal need to develop accountability and explainability of decision making by AI \cite{goodman2016european} is driving the development of methodologies for explaining the behaviour of such methods.

Our aim in this paper is to develop effective models that capture the essential dynamical behaviour of RNNs on computational tasks as input-driven responses of a dynamical system, while neglecting microscopic details of the RNN dynamics in phase space (i.e., the space of all possible neuron activations).
To this end, we hypothesise that RNNs can undertake computations by exploiting (i) transient dynamical regimes and (ii) excitable connections to switch between different stable attractors, depending on input and current state.
The RNN behaviour depends on the task at hand: for example, long transients are mostly exploited in time series forecasting problems, while switching between attractors is mostly exploited in classification problems or tasks requiring to learn a finite set of memory states. These mechanisms are not mutually exclusive and can be exploited synergistically to realise more complex computations.

\subsection{Contribution and paper organisation}

In order to test our hypothesis, we develop a theoretical framework based on network attractors of dynamical systems.
An attractor is a subset of the state space of a dynamical system (i.e., the phase space), where the state will asymptotically converge for ``usual'' choices of initial conditions.
Network attractors are special kinds of attractor which can be thought of as directed graphs, where nodes correspond to local attractors and directed edges corresponds to particular trajectories that connect (or almost connect) those local attractors. 
Since for this paper, the local attractors of interest are all stable fixed points, we will use the term local attractor synonymously with stable fixed point. A stable fixed point is the simplest possible attractors: this is a point in phase space where all variables of the dynamical system assume constant values, and all close enough initial conditions converge to this point.
However, fixed points need not be stable: they can be partially stable (saddle points), or totally unstable (repellers).
According to the nature of the trajectories connecting local attractors it is possible to consider both heteroclinic network attractors composed of heteroclinic connections between saddles (i.e. partially stable fixed points), and excitable network attractors (ENAs) composed by connections that are excitable, i.e. that require a small initial perturbation to move between stable attractors \cite{ashwin2016designing,network_attractor_review2018}.

Heteroclinic network attractors have already been suggested as models to describe transient computation \cite{rabinovich2008transient, rabinovich2001dynamical}, here, conversely, we focus attention on ENAs which have the advantage of being more robust to perturbations.
We focus particularly on ENAs between stable fixed points \cite{ashwin2016designing}, although ENAs can be theoretically conceived between any kind of local attractor: limit cycles, limit tori or strange attractors.
More precisely, we show how to construct ENAs describing RNN behaviour for tasks that involve switching between a finite set of attractors.
Interestingly, from a directed graph (representing the underlying network attractor of a dynamical system) it is possible to reverse engineer \cite{ashwin2013designing} a set of ordinary differential equations ruling, in our case, the RNN behaviour.
This, in turn, opens the way to a more analytic description of how RNNs solve computational tasks.

In this paper, we focus on the flip-flop task with two bits \cite{sussillo2013opening}, since it provides a controlled workbench to test our hypothesis. The input to discrete-time RNN is assumed to be a discrete temporal sequence with values from $\{+1,0,-1\}$.
In general, we consider discrete time $k$ varying input, denoted as ${\bf u}[k]$, and output, denoted as ${\bf z}[k]$, of a system related by
\begin{equation}
    ({\bf x}[k] , {\bf z}[k] ) ={\bf R} ({\bf u}[k],{\bf x}[k-1]),\\
\label{eq:task}
\end{equation}
where ${\bf R}$ represents an RNN with evolving internal state, namely ${\bf x}[k]$, and connection weights that are optimised during the training phase.
We consider response to inputs where, independently for any $k$, ${\bf u}[k]$ is $(0,0)$ with probability $1-p$ or otherwise chosen to be one of the four inputs $(\pm 1,0)$ or $(0,\pm 1)$ with equal probability. This generates a sequence of inputs that remain at $(0,0)$ for an exponentially distributed period of time, but occasionally takes one of the four values $(0,\pm 1)$, $(\pm 1,0)$.
The target output to be learned by the RNN is a two-dimensional vector ${\bf z}[k]$ that can assume four possible configurations: $(1, 1)$, $(1,-1)$, $(-1,1)$, and $(-1,-1)$.
The system \eqref{eq:task} is considered to successfully accomplish the task if the network is able to reliably and accurately reproduce all 20 possible actions described in Table \ref{tab:flip-flop-task}.
\begin{table}[ht!]
\centering
\caption{The two-bit flip-flop task. Depending on the output ${\bf z}[k-1]$ and input ${\bf u}[k]$, we expect the output ${\bf z}[k]$ to become as given in the table, where N.C. indicates ``no change'' from the current output value.}
\begin{tabular}{|c|c|c|c|c|c|}
\hline
\textbf{Outputs}$\backslash$\textbf{Inputs} & $(0, 0)$ & $(1, 0)$ & $(-1, 0)$  & $(0, 1)$ & $(0, -1)$\\
\hline
$(1, 1)$ & N.C. & N.C. & $(-1, 1)$ & N.C. & $(1, -1)$\\
\hline
$(1, -1)$ & N.C. & N.C. & $(-1, -1)$ & $(1, 1)$ & N.C.\\
\hline
$(-1, 1)$ & N.C. & $(1, 1)$ & N.C. & N.C. & $(-1, -1)$\\
\hline
$(-1, -1)$ & N.C. & $(1, -1)$ & N.C. & $(-1, 1)$ & N.C.\\
\hline
\end{tabular}
\label{tab:flip-flop-task}
\end{table}

We use RNNs that are ESNs subjected to supervised training with a perturbation matrix obtained by injecting the output into the ESN dynamics.
This choice does not limit the validity of our hypothesis: we claim that our results are general and can be used to describe the behaviour of more general discrete-time, input-driven RNNs. The contributions of this paper can be summarised as follows:
\begin{itemize}
\item We provide a theoretical framework to describe the behaviour of RNNs by means of ENAs. The proposed theoretical framework is general and covers several types of computational tasks. We present a specific instance of such a framework applied to describe RNN behaviour on tasks requiring to learn a finite set of memory states;

\item We use bifurcation analysis to manually design (low-dimensional) ESNs that give rise to ENAs able to solve the flip-flop task with any number of bits. This allows us to justify our choice of modelling framework and suggests its validity in the context of RNNs;

\item As ESNs are driven by inputs and subject to training via perturbation matrices, bifurcation analysis alone is not sufficient to explain changes to their behaviour. In fact, inputs and perturbation matrices can affect ESN behaviour in a non-trivial way.
To this end, we introduce an algorithm to extract the ENA describing the ESN behaviour for the task at hand.
The algorithm analyses an ESN trajectory and constructs a directed graph encoding the underlying ENA on which the dynamics takes place. The vertices (nodes) of such a graph are associated with the fixed points of the dynamics and directed edges describe excitable connections between them. We apply this algorithm to an ESN that is trained to solve the flip-flop task and show that the resulting ENA is able to explain how ESNs perform computations in a detailed and mechanistic way;

\item We define a notion of excitability threshold for this high-dimensional, non-linear dynamical system driven by inputs. We propose a method for computing this excitability threshold that accounts for inputs and can directly be applied to trajectories generated by a neural network while solving a task;

\item Our simulations suggest three important findings.
First, as already noted by recent research \cite{sussillo2013opening,aljadeff2016low}, the dynamics of high-dimensional RNNs takes place in a much lower-dimensional phase space region that is determined by the structure introduced with training and inputs.
Here, we observe that the dynamics is indeed low-dimensional, but highlight the fact that additional dimensions are used occasionally to switch between stable states according to control inputs.
This suggests that, for example, a simplistic use of methods based on explained variance to reduce dimensions needs to be avoided.
Second, we show how ENA models describing RNN behaviour can be exploited to provide a mechanistic interpretation of errors occurring during the computation undertaken by RNNs.
Finally, we show how excitability thresholds of the extracted ENAs allow us to assess the robustness of RNNs to noise-induced perturbations.

\end{itemize}

The remainder of this paper is organised as follows.
Section \ref{sec:background} introduces the essential background material needed in this paper. Section \ref{sec:analysis-design} shows how to design low-dimensional ESN models that give rise to ENAs able to solve the flip-flop task.
In Section \ref{sec:ena-extraction} we propose an algorithm for automatically extracting ENAs directly from an ESN trajectory generated while solving a task.
In Section \ref{sec:experiments}, we present and discuss results of the simulations.
Finally, Section \ref{sec:conclusions} draws conclusions and points to future research directions.
We include three appendices: Appendix~\ref{sec:linear-stability} reviews notions of linear stability used throughout the paper.
Appendix~\ref{sec:bifurcations} discuss bifurcations of fixed points for low-dimensional ESN maps.
Appendix~\ref{sec:fixed_point_aggregation} provides details of the procedure used to determine fixed points.

\section{Background}
\label{sec:background}

\subsection{Echo State Networks}

We consider a specific system of the form \eqref{eq:task}, corresponding to a discrete-time ESN state-update and related output:
\begin{align}
\label{eq:state_update}
\mathbf{x}[k] =& \phi(\mathbf{W}_{r} \mathbf{x}[k-1] + \mathbf{W}_{in}\mathbf{u}[k]), \\
\label{eq:esn_output}  
\mathbf{z}[k] =& \mathbf{W}_{o} \mathbf{x}[k].
\end{align}
$\mathbf{x}[k]\in\mathbb{R}^{N_r}$ is the state, $\mathbf{u}[k]\in\mathbb{R}^{N_i}$ and $\mathbf{z}[k]\in\mathbb{R}^{N_o}$ denote input and output, respectively. The activation function $\phi(\cdot)$ is applied component-wise; without loss of generality, we consider $\phi = \tanh : \mathbb{R} \longrightarrow (-1,1)$.
It is worth mentioning \cite{pascanu2013difficulty} that \eqref{eq:state_update} is often written $\mathbf{x}[k] = \mathbf{W}_{r} \phi(\mathbf{x}[k-1]) + \mathbf{W}_{in}\mathbf{u}[k]$. 
However, \citet{miller2012mathematical} proved that the two formulations are equivalent up to a change of coordinates; they produce the same discrete-time dynamics.
In this paper, we build on \eqref{eq:state_update} and consider a network of discrete-time leaky-integrator neurons \cite{jaeger2007optimization} of the form:
\begin{align}
\label{eq:state_update_leaking}
\mathbf{x}[k] = (1 - \alpha)\mathbf{x}[k-1] + \alpha\phi(\mathbf{W}_{r} \mathbf{x}[k-1] + \mathbf{W}_{in}\mathbf{u}[k] + \boldsymbol{\epsilon}).
\end{align}
Here, $\alpha\in(0, 1]$ is called a leak rate and explicitly sets the time-scale of the ESN \cite{tallec2018can}.
The $\boldsymbol{\epsilon}$ term represents additive white Gaussian noise with spherical covariance matrix and unit standard deviation.

The reservoir $\mathbf{W}_{r}\in \mathbb{R}^{N_r\times N_r}$ and input $ \mathbf{W}_{in} \in \mathbb{R}^{N_r\times N_i}$ matrices are usually random with i.i.d. entries drawn from uniform or Gaussian distributions \cite{lukovsevivcius2009reservoir,Lukosevicius2012}.
However, in the literature it is possible to find reservoirs with different connection patterns, including deterministic topologies \cite{rodan2012simple} and those exploiting the norm-preserving property of orthogonal matrices \cite{mayer2017orthogonal}. In our case the read-out matrix $\mathbf{W}_o \in \mathbb{R}^{N_o\times N_r}$ is optimised for the task at hand.
Relevant hyperparameters directly affecting ESN performance include the number of neurons and sparseness of their connections, the spectral radius of the reservoir matrix, and leak rate $\alpha$ \cite{bianchi2016investigating,esnfish2016}.
The so-called echo-state property (ESP) \cite{gallicchio2017echo,yildiz2012re,manjunath2013echo} guarantees the existence and uniqueness of a global attracting trajectory for any input sequence in a compact set. The ESP, although useful in some tasks like forecasting tasks, is in practice difficult to verify and it is usually formulated only for ESNs with state-update of the form shown in \eqref{eq:state_update} and \eqref{eq:state_update_leaking}.
Therefore, it is not suitable to ESN models and tasks we discuss here; see the following subsection.

\subsection{Training ESNs with low-rank perturbation matrices}
\label{sec:training}

Training of RNNs is typically implemented by means of stochastic gradient descent or variations of thereof \cite{ruder2016overview}.
Learning long-term dependencies in RNNs with gradient descent is known to be problematic, as a consequence of the so-called vanishing/exploding gradient problem \cite{pascanu2013difficulty}. To this end, different approaches have been proposed that can be summarised in two categories: (i) methods using gating mechanisms (such as Long Short-Term Memory \cite{hochreiter1997long} and Gated Recurrent Unit \cite{chung2014empirical} networks) and (ii) those based on unitary matrices and constant-slope activations \cite{arjovsky2016unitary}.
On the other hand, training of the recurrent layer in ESNs is typically realised by perturbing a randomly-initialised reservoir with a low-rank, deterministic matrix.
This is conventionally accomplished by feeding back the ESN output to the recurrent layer \cite{PhysRevLett.118.258101,vincent2016driving,koryakin2012balanced,li2014frequency,reinhart2012regularization} or, as \citet{MASTROGIUSEPPE2018} recently proposed, by designing the reservoir directly as $\mathbf{W}_r=\mathbf{X} + \mathbf{D}$, where $\mathbf{X}$ is a random matrix and $\mathbf{D}$ is a deterministic, low-rank matrix encoding the task of interest.
\begin{figure}[ht!]
\centering
\includegraphics[keepaspectratio=true,scale=0.27]{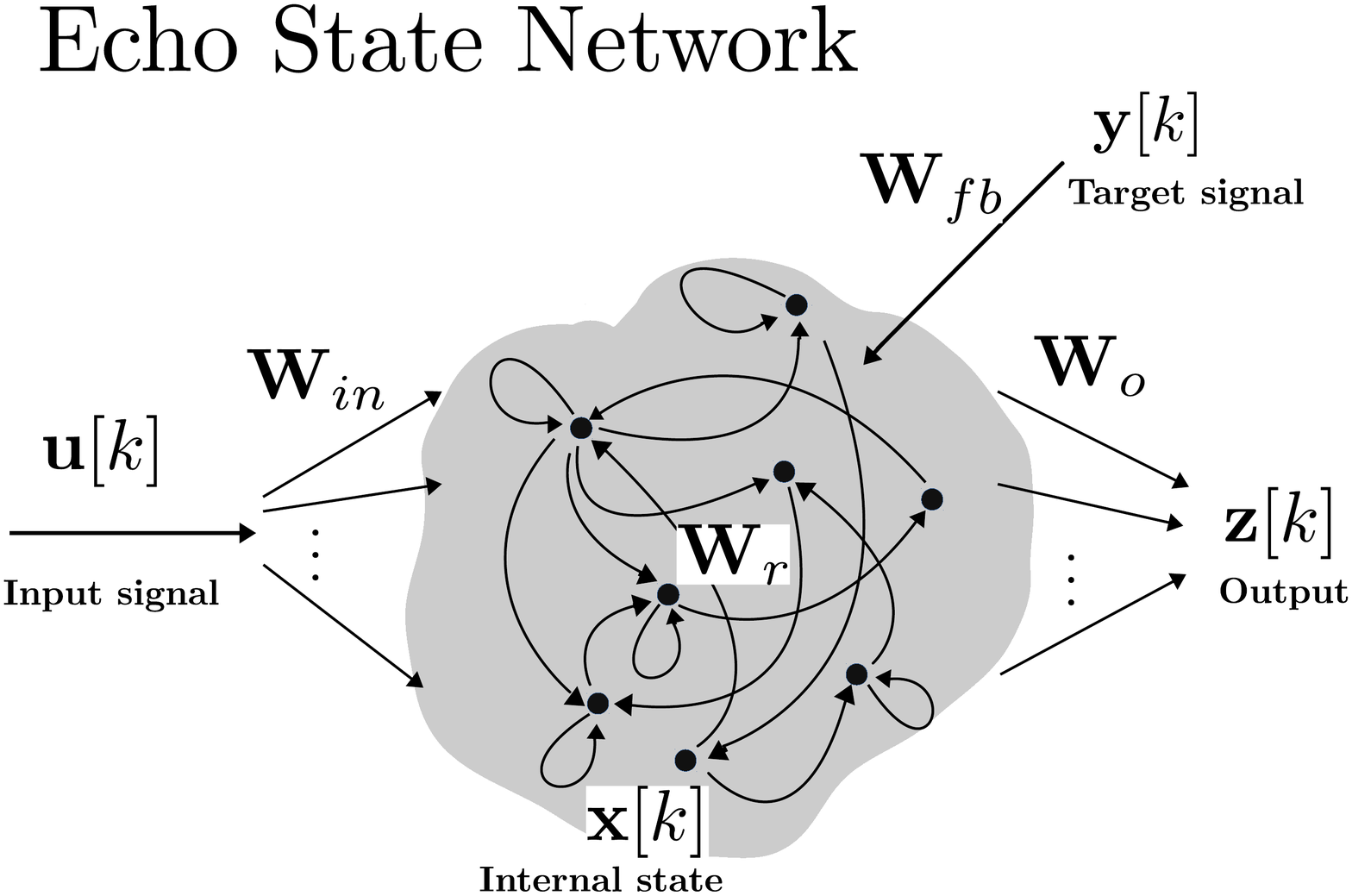}~\includegraphics[keepaspectratio=true,scale=0.27]{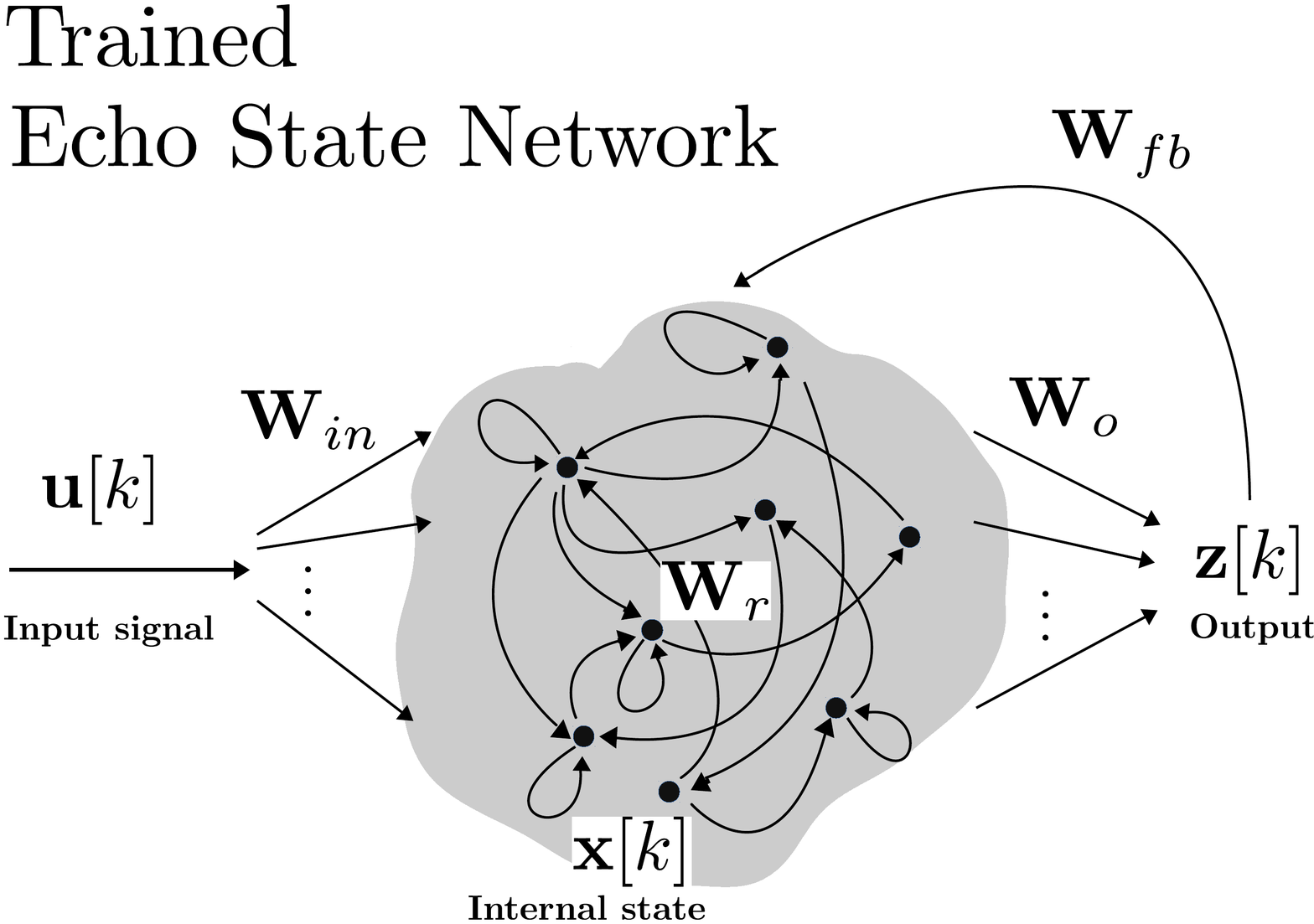}
\caption{ \textbf{Left:} Illustration of a ESN in the open-loop (training) phase \eqref{eq:state_update_leaking_fb}. Training is supervised by means of the target signal $\mathbf{y}$. 
\textbf{Right:} Illustration of an ESN \eqref{eq:activation_update_fb2} after the optimisation of the read-out matrix. Feeding back the output signal into the reservoir  gives the closed-loop (trained) system which self-sustains the dynamics driven by the input signal.}
\label{fig:echostatenetwork}
\end{figure}

In this paper, we use a supervised learning algorithm which exploits the feedback of the ESN output as a mechanism for training the recurrent layer.
The state-update equation \eqref{eq:state_update_leaking} takes the following form:
\begin{equation}
\label{eq:state_update_leaking_fb}
\mathbf{x}[k] = (1 - \alpha)\mathbf{x}[k-1] + \alpha\phi(\mathbf{W}_{r} \mathbf{x}[k-1] + \mathbf{W}_{in}\mathbf{u}[k] + \mathbf{W}_{fb}\mathbf{y}[k-1]  + \boldsymbol{\epsilon}),
\end{equation}
where $\mathbf{W}_{fb}\in\mathbb{R}^{N_r\times N_o}$ is a matrix with i.i.d. random coefficients usually drawn from a uniform or Gaussian distribution.
Depending on whether training is performed batch or online, $\mathbf{y}[k-1]$ in \eqref{eq:state_update_leaking_fb} takes the form of either the target signal or the output produced by the ESN \eqref{eq:esn_output}, respectively.
In batch mode, it is possible to distinguish two main phases; see Fig. \ref{fig:echostatenetwork} for an illustration. 
First, the reservoir is fed with an auxiliary input, i.e. the target signal $\mathbf{y}$.
We construct a matrix $\mathbf{X} \in \mathbb{R}^{N\times N_r}$ containing the states $\mathbf{x}[k]$ of the ESN generated in response to target and input signals. Finally, the weights $\mathbf{W}_o$ of the read-out are determined by solving a regularised least-squares problem, $\mathbf{W}_o = \left(\left( \mathbf{X}^{\top}\mathbf{X} + \lambda^2 \mathbf{I} \right)^{-1}\mathbf{X}^{\top} \mathbf{y}\right)^{\top}$, where $\mathbf{I}$ is an $N_r\times N_r$ identity matrix and $\lambda\geq0$ is a regularisation parameter.
Successively, the target signal is replaced by the generated output $\mathbf{z}$; this ``closed-loop phase'' corresponds to the test phase of the trained ESN. An analysis of the stability of the transition from open- to closed-loop can be found in \cite{PhysRevLett.118.258101}.
\begin{defn}
\label{def:effective_reservoir}
We call the \emph{trained reservoir} the following matrix
\begin{equation}
\label{eq:trained_reservoir}
    \mathbf{M} := \mathbf{W}_{r} + \mathbf{W}_{fb} \mathbf{W}_{o}.
\end{equation}
\end{defn}
Once the read-out matrix $\mathbf{W}_{o}$ is optimised, by imposing $\mathbf{y}[k]=\mathbf{z}[k]$ and expanding in \eqref{eq:state_update_leaking_fb} with \eqref{eq:esn_output}, we obtain:
\begin{align}
\nonumber\mathbf{x}[k] =& (1 - \alpha)\mathbf{x}[k-1] + \alpha\phi(\mathbf{W}_{r}\mathbf{x}[k-1] + \mathbf{W}_{in}\mathbf{u}[k] + \mathbf{W}_{fb} \mathbf{W}_{o}\mathbf{x}[k-1] + \boldsymbol{\epsilon})= \\
\label{eq:activation_update_fb2}  
 =& (1 - \alpha)\mathbf{x}[k-1] + \alpha\phi(\mathbf{M} \mathbf{x}[k-1]  + \mathbf{W}_{in}\mathbf{u}[k] + \boldsymbol{\epsilon}).
\end{align}
\begin{remark}
The trained reservoir \eqref{eq:trained_reservoir} is obtained by adding a matrix $\mathbf{W}_{fb} \mathbf{W}_{o}\in\mathbb{R}^{N_r\times N_r}$, which is low-rank $N_o\ll N_r$ relative to the randomly initialised reservoir matrix $ \mathbf{W}_{r} $. Therefore the reservoir is in some sense trained using output feedback connections.
\end{remark}

Inputs $\mathbf{u}[k]$ in \eqref{eq:activation_update_fb2} play the role of control inputs and are typically constant or impulsive signals.
The ESN read-out matrix $\mathbf{W}_o$ is conventionally determined by solving a regularised least-squares problem.
Nonetheless, we note that also online training schemes have been developed, e.g., the FORCE learning algorithm originally introduced in \cite{sussillo2009generating} and further extended by \citet{10.1371/journal.pone.0191527}.
During the test phase, regardless of the adopted training mechanism, the state-update of ESNs is described by \eqref{eq:activation_update_fb2}.
In this paper, we consider batch training via ridge regression and analyse the trajectory generated by \eqref{eq:activation_update_fb2} during the test phase.

It is worth nothing that our theoretical framework does not rely on a particular training method or a particular RNN architecture. We note that complicated (trained) RNN models may be described using a noisy nonautonomous dynamical system, which in our system is represented by \eqref{eq:activation_update_fb2}. We focus on ESNs as they are the simplest RNN models to test our hypothesis. For this reason, the terms ESN and RNN are used interchangeably.

\subsection{Network attractors}
\label{sec:na}

Many common dynamical systems encountered in nature are dissipative \cite{strogatz2014nonlinear,cencini2010chaos}.
In such systems, the absence of any conservation law means that typically the system evolves towards an attracting set of dimension strictly less than the original phase space dimension; such a set (or a particular subset of it) is commonly called an \emph{attractor}: formal definitions are discussed for example in \cite{milnor1985concept}. The \emph{basin of attraction} of an attractor is the set of all initial conditions from which the system evolves toward the attractor.
Attractors convey crucial information about the behaviour of the dynamical systems which have generated them.  

Here, we consider the following noise-free discrete-time dynamical system with inputs:
\begin{equation}
\label{eq:nonaut_dyn_syst}
    \mathbf{x}[k]=\mathbf{G}(\mathbf{x}[k-1], \mathbf{u}[k]),
\end{equation}
where $\mathbf{G}: \mathbb{R}^{N_r}\times \mathbb{R}^{N_i} \longrightarrow \mathbb{R}^{N_r}$ is related to \eqref{eq:activation_update_fb2} as follows:
\begin{equation}
\label{eq:map}
\mathbf{G}(\mathbf{x}, \mathbf{u})= 
\begin{pmatrix}
(1-\alpha)x_1 + \alpha\phi\bigl(\mathbf{M}_{(1)}\cdot\mathbf{x} + (\mathbf{W}_{in})_{(1)}\cdot\mathbf{u}\bigl) \\
(1-\alpha)x_2 + \alpha\phi\bigl(\mathbf{M}_{(2)}\cdot\mathbf{x} + (\mathbf{W}_{in})_{(2)}\cdot\mathbf{u}\bigl)  \\
\vdots \\
(1-\alpha)x_{N_r} + \alpha\phi\bigl(\mathbf{M}_{(N_r)}\cdot\mathbf{x} + (\mathbf{W}_{in})_{(N_r)}\cdot\mathbf{u}\bigl) 
\end{pmatrix},
\end{equation}
where is $\mathbf{M}$ the trained reservoir matrix \eqref{eq:trained_reservoir}, the subscript $(i)$ denotes the $i$th rows of a matrix and $\cdot$ the usual dot product.
As mentioned before, the input signal for the flip-flop task is null most of the time, at which point it is governed by the autonomous dynamics of
\begin{equation}
\label{eq:autonomous_system}
    \mathbf{x}[k] = \mathbf{F}(\mathbf{x}[k-1]) = \mathbf{G}(\mathbf{x}[k-1], \mathbf{0}).
\end{equation}
Fixed points $\mathbf{p}\in \mathbb{R}^{N_r}$ are solutions of $\mathbf{F}(\mathbf{p}) = \mathbf{p}$.
Related to the notion of attractor is the notion of \emph{limit set} \cite{cencini2010chaos}, thus we introduce the following 
\begin{defn}
\label{def:omega_limitset}
The \emph{$\omega$-limit set} of a point $\mathbf{x}_0 $ under the iterated map \eqref{eq:autonomous_system} is defined by
\begin{equation}
\label{eq:omega_limitset}
    \omega_{\mathbf{F}}(\mathbf{x}_0) := \bigcap_{n \in \mathbb{N}} \overline{\{ \mathbf{F}^h(\mathbf{x}_0) \, | \quad h > n \}}.
\end{equation}
\end{defn}
\begin{remark}
The $\omega$-limit set of a point $\mathbf{x}_0 $ is the set of limit points of the forward trajectory $\{ \mathbf{F}^h(\mathbf{x}_0) \}_{h \in \mathbb{N}} $. For a given fixed point $\mathbf{p}$, its basin of attraction is formed by all points $\mathbf{x} \in \mathbb{R}^{N_r}$ such that $\omega_{\mathbf{F}}(\mathbf{x}) = \{ \mathbf{p} \}$. If such a fixed point is stable, then there exists a neighbourhood of $\mathbf{p}$ whose $\omega$-limit set corresponds to this fixed point.
\end{remark}

%Otherwise, the $\omega$-limit set can be formed by just the fixed point itself (repeller) or a manifold of dimension less than the phase space dimension (saddle).

\begin{figure}[ht!]
\centering
\includegraphics[keepaspectratio=true,scale=0.325]{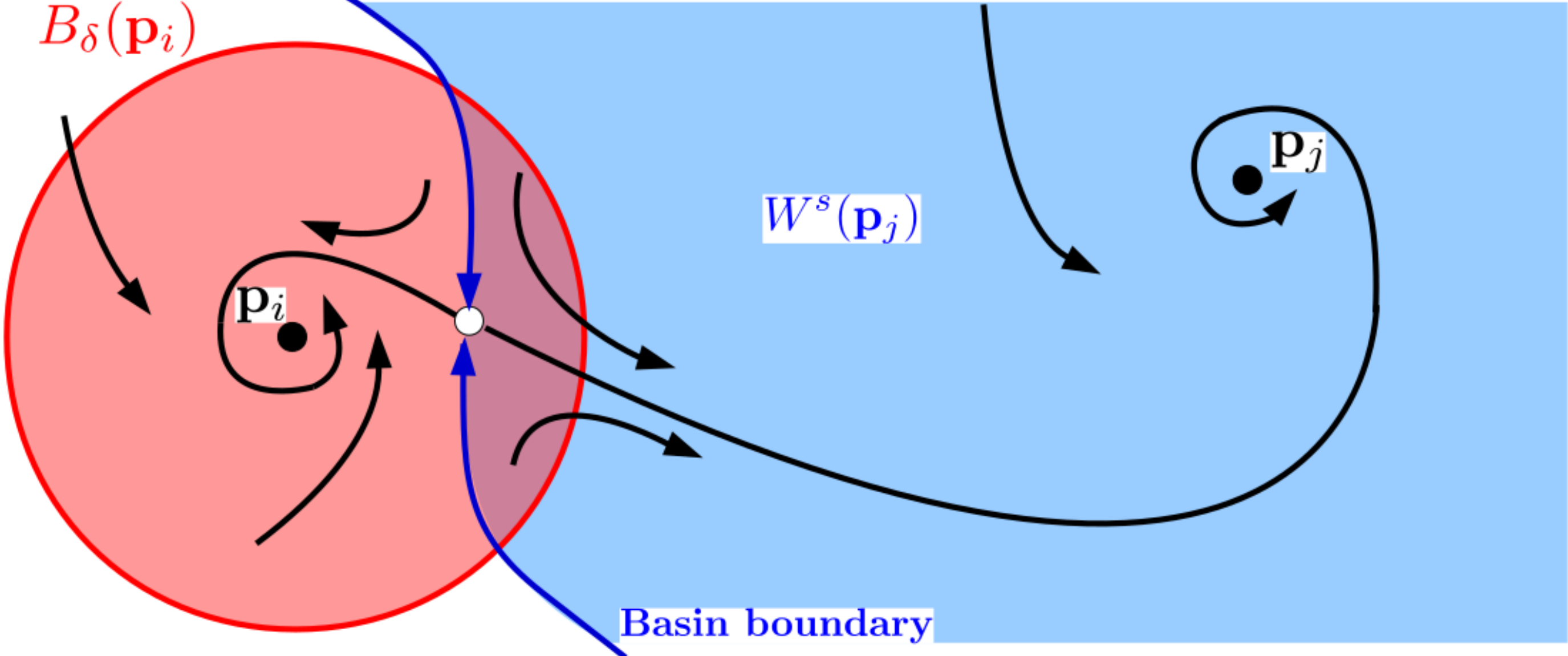}~\includegraphics[keepaspectratio=true,scale=0.355]{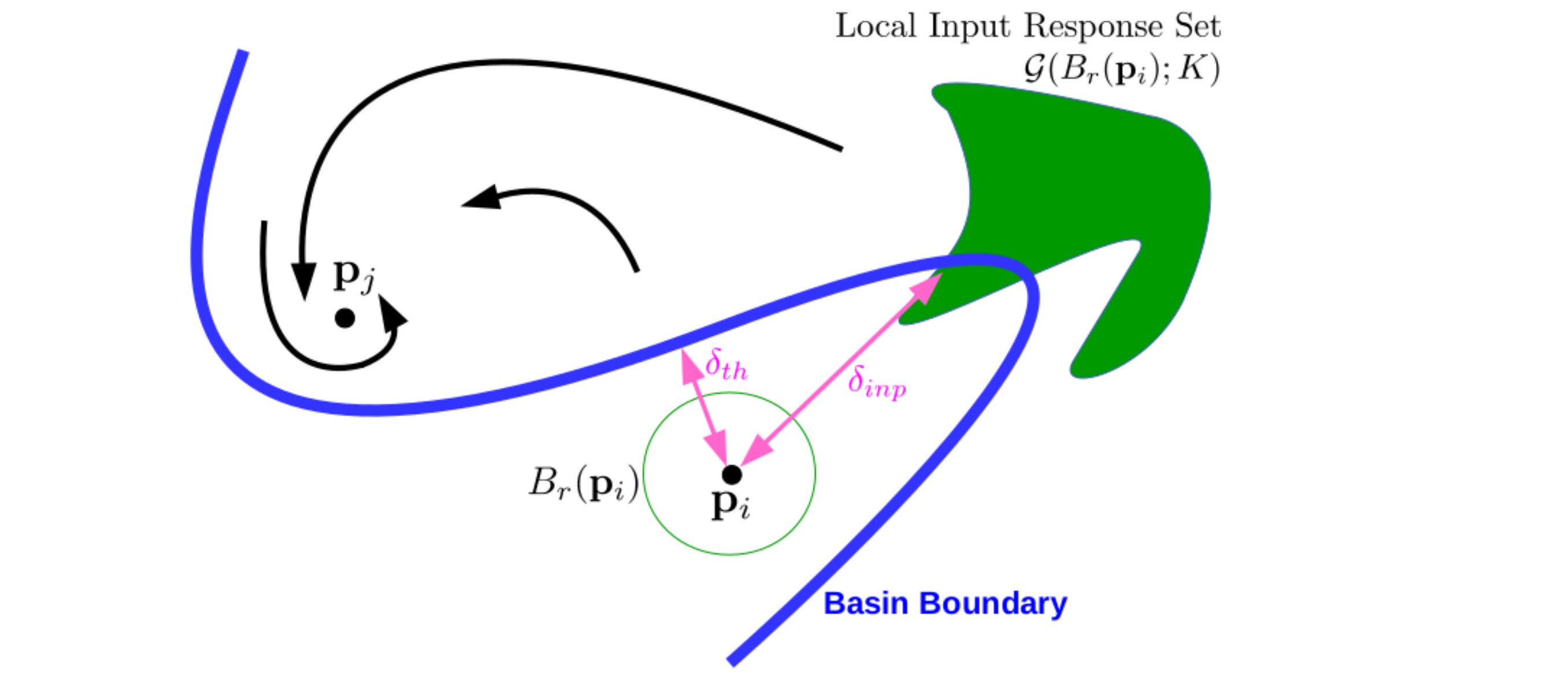}
\caption{ \textbf{Left:} Depiction of an excitable connection from $\mathbf{p}_i$ to $\mathbf{p}_j$. The red area is a closed ball centred on $\mathbf{p}_i$ of radius $\delta$. The blue area represents the stable manifold of $\mathbf{p}_j$, i.e., its basin of attraction. The open circle represents a saddle whose stable manifold (blue curves) denotes the boundary of the basin of attraction of $\mathbf{p}_j$. Some points of $B_{\delta}(\mathbf{p}_i)$ converge to $\mathbf{p}_i$ itself, while those points beyond the basin boundary converge towards $\mathbf{p}_j$, as suggested by the black arrows.
\textbf{Right:} Representation of the activation of an excitable connection through the action of the input which allows to accomplish the switch from the stable point $\mathbf{p}_i$ to the stable point $\mathbf{p}_j$. Firstly, the internal state of the RNN stands nearby the stable point $\mathbf{p}_i$, in the neighbourhood $B_r(\mathbf{p}_i)$. Then, a control input $\mathbf{u}[k+1] \in K$ drives the current internal state $\mathbf{x}[k]$ to the next state $\mathbf{x}[k+1]$ inside the local input response set, represented as the green subregion. Finally, if the state falls beyond the basin boundary then the internal state converges to the stable point $\mathbf{p}_j$. Excitability threshold $ \delta_{\text{th}}(\mathbf{p}_i,\mathbf{p}_j) $, in \eqref{eq:threshold}, is computed along the direction where the distance is shortest in order to escape from the basin of attraction of $\mathbf{p}_i$ and converge to $\mathbf{p}_j$. Nevertheless, input can potentially drive the dynamics towards alternative dimensions for the purpose of achieving the switch from $\mathbf{p}_i$ to $\mathbf{p}_j$. The input-driven excitability threshold $ \delta_{\text{inp}}(\mathbf{p}_i,\mathbf{p}_j)$, in \eqref{eq:inp_driven_threshold}, is computed considering the subspace exploited by the input to solve the task. }
\label{fig:excitable_connection} 
\end{figure}

More complex attractors consisting of networks of invariant sets in phase space have been proposed in the literature \cite{network_attractor_review2018,neves2017noise}.
Such models found renewed interest in neuroscience \cite{miller2016itinerancy,tsuda2015chaotic} and other fields of research, as they provide a fundamental tool to describe dynamic processes occurring on transients that explore excitable connections.
More relevant to our paper, we focus on networks of stable fixed points that are connected by excitable connections.
Following \cite{ashwin2018sensitive,ashwin2016designing}, we say that there exists an \emph{excitable connection} for amplitude $\delta>0$ from stable fixed point $\mathbf{p}_i$ to $\mathbf{p}_j$ whenever
\begin{equation}
\label{eq:excitable_connection}
B_{\delta}(\mathbf{p}_i) \cap W^s(\mathbf{p}_j) \neq \emptyset,
\end{equation}
where $B_{\delta}(\mathbf{p}_i)$ stands for the closed ball centred on $\mathbf{p}_i$ with radius $\delta > 0$ and $W^s(\mathbf{p}_j) = \bigl\{ \mathbf{x} \in \mathbb{R}^{N_r} \,\, | \,\, \omega_{\mathbf{F}}(\mathbf{x}) = \{ \mathbf{p}_j \} \bigl\}$ denotes the basin of attraction\footnote{The basin of attraction corresponds to the stable manifold whenever $\mathbf{p}_j$ is hyperbolic. This has interior if $\mathbf{p}_j$ is an attractor.} of the fixed point $\mathbf{p}_j$. 
\begin{defn}
\label{def:threshold}
We define \emph{excitability threshold} \cite{ashwin2016designing}  (or just \emph{threshold}) of the excitable connection from $\mathbf{p}_i$ to $\mathbf{p}_j$, and denote it as $\delta_{th}(\mathbf{p}_i,\mathbf{p}_j)$, the following nonnegative real number:
\begin{equation}
\label{eq:threshold}
    \delta_{\text{th}}(\mathbf{p}_i,\mathbf{p}_j) := \inf \{  \delta>0 \,\, : \,\, B_{\delta}(\mathbf{p}_i)\cap W^s(\mathbf{p}_j)  \neq \emptyset  \}.
\end{equation}
\end{defn}
\begin{remark}
The quantity \eqref{eq:threshold} can be informally interpreted as the fact that the fixed point $\mathbf{p}_i$ is $\delta_{\text{th}}(\mathbf{p}_i,\mathbf{p}_j)$ away from the basin of $\mathbf{p}_j$; see Fig. \ref{fig:excitable_connection} for a visual explanation.
\end{remark}

\begin{defn}
\label{def:ENA}
A set $X_{\text{exc}} \subset \mathbb{R}^{N_r}$ is called an \emph{excitable network attractor} (ENA) for amplitude $\delta>0$ if there exists a collection of fixed points $\{ \mathbf{p}_i \}_{i=1}^{M}$, such that
    \begin{equation}
    \label{eq:ENA}
        X_{\text{exc}}(\{\mathbf{p}_i \}_{i=1}^{M}, \delta) := \bigcup_{\substack{i,j=1 \\ i\neq j}}^{M} \bigl\{ \mathbf{F}^h( B_{\delta}(\mathbf{p}_i) ) \bigl\}_{h\geq 0} \, \cap  \, W^s(\mathbf{p}_j),
    \end{equation}
where $\mathbf{F}( B_{\delta}(\mathbf{p}_i) ) := \{  \mathbf{F}( \mathbf{x} ) \,\, | \,\, \mathbf{x} \in B_{\delta}(\mathbf{p}_i) \}$ is based on \eqref{eq:autonomous_system} (see \cite{ashwin2018sensitive,ashwin2016designing}).
\end{defn}
The autonomous dynamics on such a set converges to one of the stable fixed points; hence, external inputs are necessary to get interesting dynamics.
If the external input is large enough, then the state will escape from the current basin of attraction and switch to a different one, until another sufficiently large input will lead to another change of basin.

The excitability threshold \eqref{eq:threshold} is defined as the (Euclidean) distance between a given stable fixed point, $\mathbf{p}_i$, and the basin of attraction of another fixed point, say $\mathbf{p}_j$.
Such a quantity measures the minimum distance in phase space necessary to escape from the basin of $\mathbf{p}_i$ and converge towards $\mathbf{p}_j$.
Nevertheless, if the dynamical system is high dimensional, then there will be a large number of possible escaping directions that could be exploited by inputs.
Therefore, excitability thresholds \eqref{eq:threshold} alone may not be representative of nonautonomous systems driven by inputs.
For this purpose, in order to take into account the action of inputs on the dynamics, we introduce the notion of \emph{input-driven excitability threshold} of an excitable connection.
Considering $K$ as a compact subspace of $\mathbb{R}^{N_i}$, we define $\mathcal{G}(\mathbf{x}_0 ; K) := \bigcup_{\mathbf{u} \in K} {\mathbf{G}(\mathbf{x}_0 , \mathbf{u} ) } $, where $\mathbf{G}$ is the function in \eqref{eq:map} defining the nonautonomous dynamical system which describes the trained neural network.
The set $ \mathcal{G}(\mathbf{x}_0 ; K ) $ contains all states reachable from $\mathbf{x}_0$ under the action of input values $\mathbf{u} \in K$.
Importantly, the presence of noise has the effect to perturb away the internal state from the exact location of the stable point in the deterministic counterpart of the dynamics. Therefore, instead of $\mathcal{G}(\mathbf{p}_i; K)$ we rather observe the following set.
\begin{defn}
\label{def:loc_inp_resp_set}
We call \emph{local input response set} of the stable point $\mathbf{p}_i$ the subset of phase space defined by 
    \begin{equation}
    \label{eq:loc_inp_resp_set}
        \mathcal{G}(B_r(\mathbf{p}_i); K ) := \bigcup_{\mathbf{x} \in B_r(\mathbf{p}_i)}{\mathcal{G}(\mathbf{x}; K) },
    \end{equation}
where the radius $r$ can be shrunk according to the amplitude of the noise and $K$ is a subset of admissible input values for the task at hand.
\end{defn}
\begin{remark}
Continuity of $\mathbf{G}$ implies that, if $r \rightarrow 0^+$, then the monotonically decreasing sequence of sets $\{ \mathcal{G}(B_r(\mathbf{p}_i); K ) \}_{r\geq 0} $ converges to $ \mathcal{G}(\mathbf{p}_i; K )$ regardless of $K \subset \mathbb{R}^{N_i}$.
Furthermore, we note that $\mathcal{G}(B_r(\mathbf{p}_i); K ) $, as a subspace of $\mathbb{R}^{N_r}$, is compact if $K\subset \mathbb{R}^{N_i}$ is compact.
\end{remark}
If $K$ represents all possible inputs of a particular task\footnote{For example, in the flip-flop task we have $K=\{ (0,0), \, (1,0), \, (-1,0), \,(0,1), \,(0,-1) \}$}, then we can drop it and denote \eqref{eq:loc_inp_resp_set} simply as $\mathcal{G}(B_r(\mathbf{p}_i) ) $. 
Therefore, the subregion of the phase space represented by the local input response set $\mathcal{G}(B_r(\mathbf{p}_i)) $, encodes the action exercised by the input when the internal state of the RNN is nearby the stable point $ \mathbf{p}_i $.
\begin{defn}
\label{def:inp_driven_threshold}
We define \emph{input-driven excitability threshold} of the excitable connection from $\mathbf{p}_i$ to $\mathbf{p}_j$, and denote it as $\delta_{\text{inp}}(\mathbf{p}_i,\mathbf{p}_j)$, the following nonnegative real number:
\begin{equation}
\label{eq:inp_driven_threshold}
    \delta_{\text{inp}}(\mathbf{p}_i,\mathbf{p}_j) := \inf \{  \delta>0 \,\, : \,\, \mathcal{G}(\mathbf{p}_i)\cap B_{\delta}(\mathbf{p}_i)\cap W^s(\mathbf{p}_j)  \neq \emptyset  \}.
\end{equation}
\end{defn}
\begin{remark}
The excitability threshold in \eqref{eq:inp_driven_threshold} has a similar meaning to the one defined in \eqref{eq:threshold}, although it considers only the subspace exploited by the action of inputs nearby $ \mathbf{p}_i $, where the excitable connection starts from, see Fig. \ref{fig:excitable_connection}.
Finally, from the fact that $\mathcal{G}(\mathbf{p}_i)\cap B_{\delta}(\mathbf{p}_i)\cap W^s(\mathbf{p}_j) \subseteq  B_{\delta}(\mathbf{p}_i)\cap W^s(\mathbf{p}_j)$, it follows that $\delta_{\text{th}}(\mathbf{p}_i,\mathbf{p}_j)  \leq \delta_{\text{inp}}(\mathbf{p}_i,\mathbf{p}_j)$ holds for all pairs of fixed points. 
\end{remark}

%%%%%%%%%%
%%%%%%%%%%
\section{Designing low-dimensional ESNs to solve flip-flop tasks}
\label{sec:analysis-design}

In this section, starting from ESN models we show how to manually design ENAs that realise computations needed for the flip-flop task. This step allows us to justify the choice of the modelling framework adopted here and its validity in the context of RNNs.
For this purpose, we rely on bifurcation theory; see Appendix \ref{sec:bifurcations} for technical notions.
A bifurcation \cite{kuznetsov2013elements} is a qualitative change in the solution set (phase space topology) on changing a parameter of a dynamical system. On varying the parameters of the model, if such qualitative changes appear, then we say the system has undergone a bifurcation.
For this reason, the notion of bifurcation plays an important role in training RNNs and in describing their behaviour. 
Training the recurrent layer of RNNs corresponds to shaping their phase space topology, possibly inducing bifurcations that lead to a qualitative change in behaviour.
We argue that adding a low-rank perturbation matrix to the reservoir \eqref{eq:trained_reservoir} can induce bifurcations that lead to the creation of attracting regions in ESN phase space useful to solve the task at hand. Clearly, this can also happen with more sophisticated training algorithms.
Let us assume the origin as the only global attractor for the autonomous system $ \mathbf{x}[k] = (1 - \alpha)\mathbf{x}[k-1] + \alpha\phi(\mathbf{W}_{r} \mathbf{x}[k-1] ) $ associated with \eqref{eq:state_update_leaking_fb}.
Then, the bifurcation induced by \eqref{eq:trained_reservoir} leads to a transition from such a trivial dynamic to one where the origin becomes unstable\footnote{The origin could in principle remain stable and other attractors appear elsewhere: an example can be found in \cite{yildiz2012re}.} and repels trajectories towards other attracting regions in phase space, where the nonautonomous dynamics actually takes place.

In Sec. \ref{sec:2D-example}, we provide a low-dimensional example of an ESN that is able to solve the flip-flop task; the reservoir of such ESN is formed by two neurons. In Sec. \ref{sec:4D-example}, instead we design an ESN model with a reservoir formed by $2k$ neurons which is able to solve the flip-flop task with $k$ bits. For both examples we show there is an underlying ENA that explains their dynamics.

\subsection{A minimal-dimension example to solve the two-bit flip-flop task}
\label{sec:2D-example}

\begin{figure}[ht!]
\begin{center}
\includegraphics[keepaspectratio=true,scale=0.4]{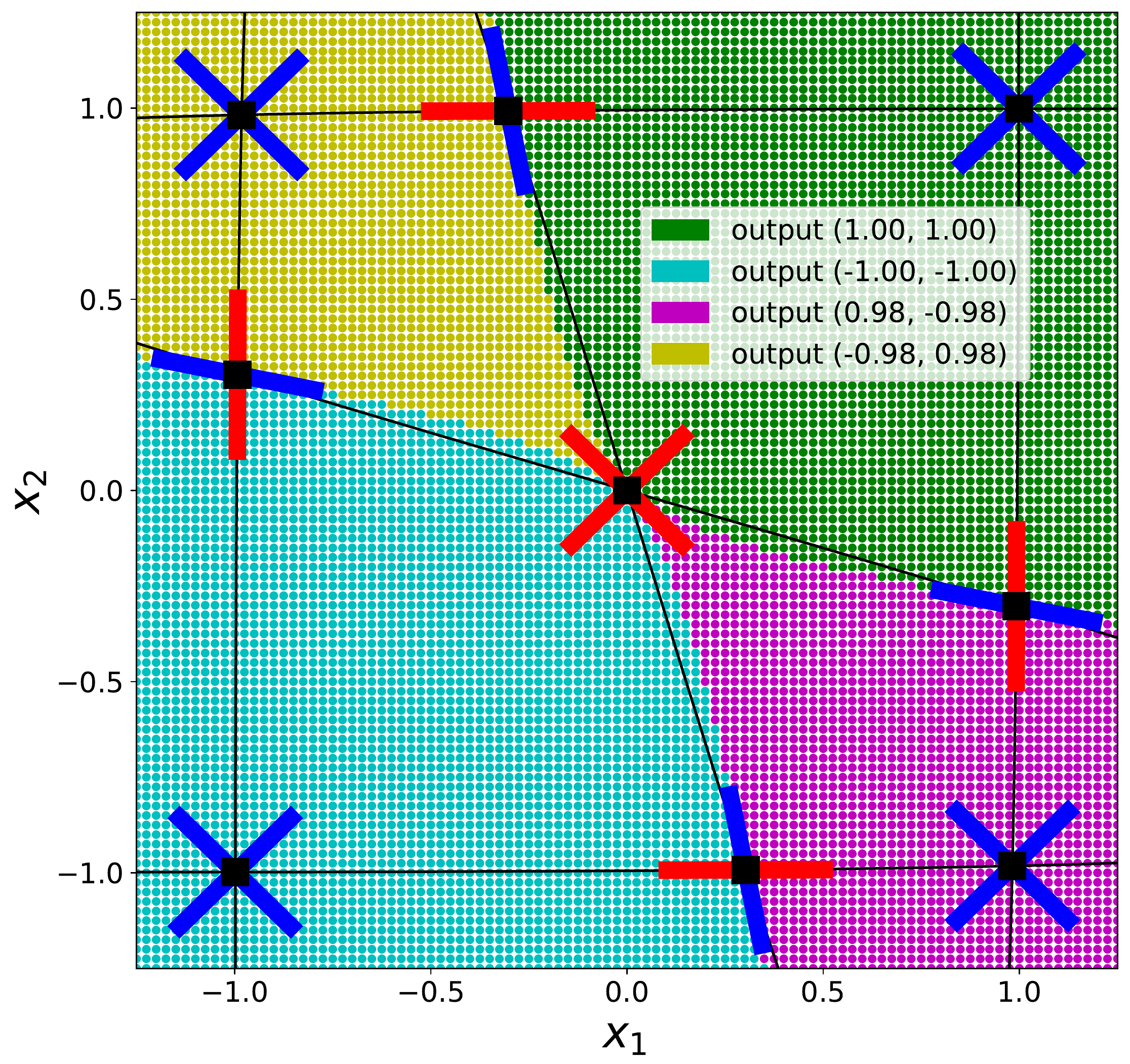}
\end{center}
\caption{Basins of attraction, nullclines, fixed points and corresponding eigenvectors of the linearized two-dimensional system  \eqref{eq:M_2dim}, with $b=0.2$.
Black curves represent the nullclines (see Appendix \ref{sec:bifurcations}) whose reciprocal intersections determine fixed points (black squares).
Red segments indicate eigenvectors of the linearized system for real eigenvalues larger than one. Blue line segments represent eigenvectors of real eigenvalues in $(0,1)$. Particularly important are blue lines of saddles, which represent local linear approximations of the boundary of the basins of attraction.
The whole phase space is divided in four basins of attraction associated to the four stable points and their boundaries. These boundaries between these basins coincide with the stable manifolds of the saddles.
The legend shows output values produced at every attractor, which, in this specific example, correspond with the internal state, i.e. the phase space coordinates of the attractors.
Points on the plane have been coloured according to the attractor to which they apparently converged to after 100 steps.}
\label{fig:2D_phasespace}
\end{figure}

In order to solve the $k$-bit flip-flop task, one needs to learn $2^k$ memory states (stable fixed points) and the related switching patterns dictated by control inputs. Here, we show how to design an ESN with two neurons in the recurrent layer, giving rise to an ENA able to solve the flip-flop task with $k=2$ bits.
According to the analysis of Appendix \ref{sec:bifurcations}, we know that it is possible to obtain, with $k$ neurons, up to $3^k$ fixed points, $2^k$ of which are stable, $3^k - 2^k -1$ are saddles, and $1$ (the origin) is a repeller.
Therefore, we set the trained reservoir weights \eqref{eq:trained_reservoir} according to condition \eqref{eq:9fixpoint_condit}.
In particular,
\begin{equation}
\label{eq:M_2dim}
\mathbf{M}(b) := \omega_{r}
\left[\begin{array}{cccc}
1 & b \\
b & 1 \\
\end{array}\right]
\end{equation}
with $\omega_{r}=3$ scaling the reservoir weights, and $b$ acting as tuning parameter.
As shown in Fig. \ref{fig:2D_phasespace}, for every $0 \leq b < 0.47 $, the autonomous system $\mathbf{x}[k] = \tanh\bigl(\mathbf{M}(b) \mathbf{x}[k-1] \bigl)$ has four stable attractors located near the vertices of the invariant square $[-1,1]^2$.

The input is injected into the ESN via the following weight matrix, $\mathbf{W}_{in}=\omega_{in}\mathbf{I}_2$, i.e. a scaled $2\times 2$ identity matrix, setting the scaling factor (called a hyperparameter in machine learning) to $\omega_{in}=6$.
Let us set the remaining parameters in \eqref{eq:activation_update_fb2} as $\boldsymbol{\epsilon}=0$ and $\alpha=1$, and let the ESN output \eqref{eq:esn_output} be the identity mapping.
Let us assume that, at time $k$, the system is in state $\mathbf{p}=(p_1,p_2)$, which is close to one of the four attractors, i.e., $\mathbf{p}_1 \approx (1,1), \ \mathbf{p}_2 \approx (-1,1), \ \mathbf{p}_3 \approx (-1,-1), \ \mathbf{p}_4 \approx (1,-1)$. 
The possible, non-null inputs at time-step $k$ are $\mathbf{u}[k]\in \{(1,0), (0,1), (-1,0), (0,-1)\}$. 
The action of the input pulse is encoded in a vector $\mathbf{\Delta}\in [-1, 1]^2$, representing the difference between the state before and after the occurrence of such a pulse, whose components are
\begin{equation}
\label{eq:kick_input}
    \Delta_j \approx \tanh\bigl( 3 p_j +  6 u_j  \bigl) - \tanh\bigl(  3  p_j \bigl) , \quad j=1,2,
\end{equation}
considering $\omega_r = 3$, $\omega_{in} = 6 $ and, for the sake of simplicity, $b=0$, which corresponds to the case of two independent neurons.
Hence, the switching mechanism between attractors is ruled by the signs of both $p_j$ and $u_j$; the former encoding the position and the latter the direction where to move.
Therefore, the state changes if and only if $p_j$ and $u_j$ assume different signs for some $j\in \{1,2\}$.
In fact, if they have the same sign, then \eqref{eq:kick_input} is null because $\tanh\bigl( \sgn(p_j) [ 3 + 6 ] \bigl) \approx \tanh\bigl( \sgn(p_j)  3 \bigl)$ due to saturating activation functions. While, if they have different sign, then \eqref{eq:kick_input} becomes close to $-1$ or $1$, according to $\sgn(p_j) $, because $ \tanh\bigl( \sgn(p_j)  [ 3 - 6 ]  \bigl) = -\tanh\bigl( \sgn(p_j) 3 \bigl)$. 
Clearly, it is not necessary that $|\omega_r-\omega_{in}|= \omega_r $ holds, as in our set up. 
Although we assumed $b=0$ for clarity of explanation, the conclusion is the same for each $b \in [0,0.47)$. However, when the parameter approaches the bifurcation value $b \approx 0.47$, the escaping dynamics from certain fixed points become significantly slower.

Given a set of stable fixed points, a $\delta>0$ gives rise to a specific ENA; see definition of ENA in \eqref{eq:ENA}.
For instance, a large $\delta$ will potentially activate all excitable connections between the attractors.
However, in order to properly solve the flip-flop task, some of the connections need not to be active, namely, the diagonal connections between $\mathbf{p}_1$ and $\mathbf{p}_3$, and between $\mathbf{p}_2$ and $\mathbf{p}_4$.
Due to symmetry, there are only three different excitability thresholds that an excitable connection can have, that is, $\delta_{th}(\mathbf{p}_2,\mathbf{p}_1) , \delta_{th}(\mathbf{p}_1,\mathbf{p}_2), \delta_{th}(\mathbf{p}_1,\mathbf{p}_3)$. In the particular configuration shown in Fig. \ref{fig:2D_phasespace}, it holds that $\delta_{th}(\mathbf{p}_2,\mathbf{p}_1) < \delta_{th}(\mathbf{p}_1,\mathbf{p}_2) < \delta_{th}(\mathbf{p}_1,\mathbf{p}_3)$.
Therefore, the underlying ENA supporting the nonautonomous ESN dynamics is defined as $X_{\text{exc}} (\{ \mathbf{p}_i \}_{i=1}^{4}, \delta)$, for every $\delta_{th}(\mathbf{p}_1,\mathbf{p}_2) < \delta < \delta_{th}(\mathbf{p}_1,\mathbf{p}_3)$.
We note that $\delta_{th}(\mathbf{p}_1,\mathbf{p}_3) \approx \sqrt{2}$, regardless of $b \in [0, \, 0.47) $.
We can reduce the size of the basin of attraction of $\mathbf{p}_2$ and $\mathbf{p}_4$ by increasing the value of $b$, which in turn reduces the excitability threshold of the corresponding connections, until at $b\approx 0.47$ a bifurcation occurs making them disappear\footnote{A bifurcation where both saddles collide with the corresponding stable points $\mathbf{p}_2$ and $\mathbf{p}_4$, simultaneously annihilating all six fixed points.}.
However, the basins of attraction of the other two stable points, $\mathbf{p}_1$ and $\mathbf{p}_3$, become bigger ($\delta_{th}(\mathbf{p}_1,\mathbf{p}_2) \approx 2 - \delta_{th}(\mathbf{p}_2,\mathbf{p}_1) $), so increasing their excitability thresholds. 
Therefore, when the ESN state is close to $\mathbf{p}_1$ (or $\mathbf{p}_3$), the input needs to be very precise to throw back the state to the narrow basins of $\mathbf{p}_2$ and $\mathbf{p}_4$. Moreover, it must have a large amplitude compared to what is needed to escape from such narrow basins.
Nevertheless, setting $\omega_{in}$ large enough and depending on $b$ (until a value of $\omega_{in}=6$, which is enough for every $0 < b < 0.47$), the system is able to reproduce the flip-flop dynamics without errors.

Unfortunately, it does not seem to be possible to design a two-dimensional ESN for the two-bit flip-flop where the excitability of each attractor can be tuned by changing the location of the nearby saddles. Nevertheless, as we will show in the next section, this becomes possible by including two additional dimensions in the model.

\subsection{A $2k$-dimensional model for $k$-bit flip-flop tasks}
\label{sec:4D-example}

In this section, we propose a $2k$-dimensional model able to solve $k$-bit flip-flop tasks.
For the sake of clarity, but without loss of generality, we show the $k=2$ bit case.
The key insight is to model the switching dynamics between two fixed points in a two-dimensional space and then build a model formed by two independent systems with two-dimensional switching dynamics.

\begin{figure}[ht!]
\centering
\includegraphics[keepaspectratio=true,scale=0.4]{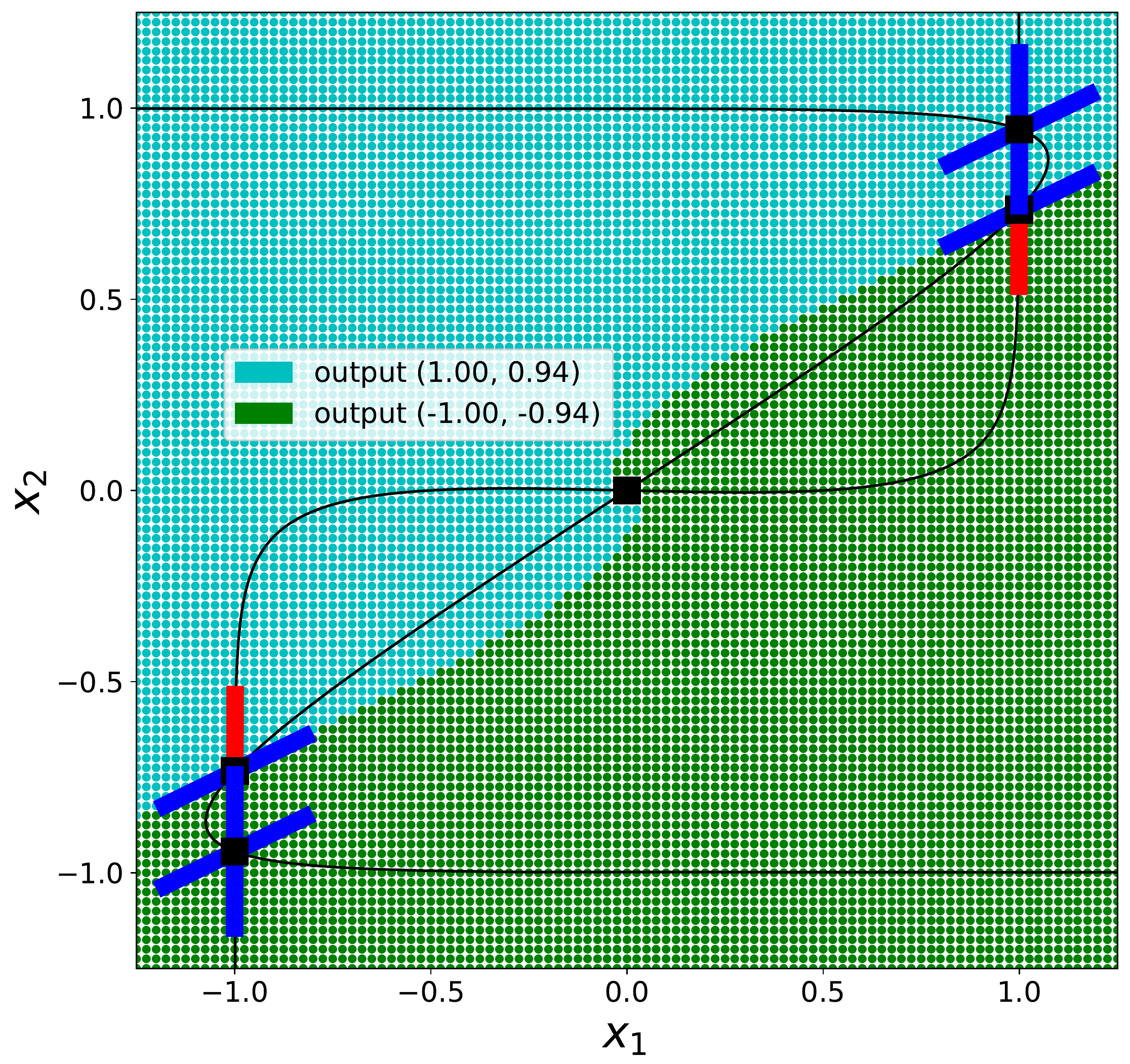}
\caption{Basins of attraction, nullclines, fixed points and eigenvectors of the linearized two-dimensional autonomous system having \eqref{eq:bloccoM_4dim} as reservoir matrix, with $s=2$.
Black curves represent the nullclines (see Appendix \ref{sec:bifurcations}) whose reciprocal intersections determine fixed points (black squares).
Red segments indicate real eigenvectors associated with real eigenvalues larger than one. 
Blue line segments represent real eigenvectors of real eigenvalues in $(0,1)$.
Points of the plane have been coloured according with the attractor on which they converged after 100 steps.}
\label{fig:4D_phasespace} 
\end{figure}

Unlike the previous example, here we want to tune the excitability of all connections. This can be obtained by imposing the condition in \eqref{eq:reverse5fixpoint_condit} and tuning the reservoir weights to change the position of the saddles.
Therefore, we define
\begin{equation}
\label{eq:bloccoM_4dim}
\mathbf{B} :=
\left[\begin{array}{cccc}
1.1 & 4 \\
-s & 4 \\
\end{array}\right],
\end{equation}
where $s$ is a real parameter. Hence, for $0\leq s<2.15$, the autonomous dynamical system defined by $\mathbf{x}[k] = \tanh\bigl(\mathbf{B} \mathbf{x}[k-1] \bigl)$ has one repeller (the origin), two attractors, namely $\mathbf{p}_1\approx(1,1), \, \mathbf{p}_2\approx (-1,-1)$, and two saddles.
By increasing $s$ within the $[0, 2.15)$ interval, we can make the saddles closer to the respective stable points, see Fig. \ref{fig:4D_phasespace}, hence decreasing the excitability thresholds of these attractors, until a saddle-node bifurcation occurs approximately at $s=2.15$, annihilating attractors with saddles.

Let us define the input matrix as follows:
\begin{equation*}
\label{eq:4D_inpmatrix}
\mathbf{W}_{in} :=\omega_{in}
\left[\begin{array}{cccc}
0 & 0 \\
1 & 0 \\
\end{array}\right],
\end{equation*}
where $\omega_{in}$ is a positive real parameter.
For instance, setting $s=2$ and $\omega_{in}=1$, the two-dimensional dynamical system defined by $\mathbf{x}[k] = \tanh\bigl(\mathbf{B} \mathbf{x}[k-1] + \mathbf{W}_{in}\mathbf{u}[k] \bigl)$ is able to accomplish the switching mechanism between $\mathbf{p}_1$ and $\mathbf{p}_2$ according to inputs $(1,0),\,(-1,0)$, and ignore other inputs, i.e., $(0,1),\,(0,-1)$.
Of course, replacing zeros in $\mathbf{W}_{in}$ with relatively small values (compared to $\omega_{in}$) does not change the results.

The complete system able to solve the two-bit flip-flop dynamics can be obtained by defining the reservoir as the following four-dimensional block diagonal matrix,
\begin{equation}
\label{eq:M_4dim}
   \mathbf{M} := \left[\begin{array}{cc} \mathbf{B} & \mathbf{0} \\ \mathbf{0} & \mathbf{B}  \end{array}\right],
\end{equation}
where $\mathbf{0}$ denotes a $2\times 2$ matrix with all zeros.
Starting from a given initial condition, for every $0\leq s<2.15$, the state of the four-dimensional autonomous dynamical system $\mathbf{x}[k] = \tanh\bigl(\mathbf{M} \mathbf{x}[k-1] \bigl)$ converges to one of these four attractors $(\mathbf{p}_1,\mathbf{p}_1),\,(\mathbf{p}_1,\mathbf{p}_2),\,(\mathbf{p}_2,\mathbf{p}_1),\,(\mathbf{p}_2,\mathbf{p}_2)$. 
In order to produce a two-dimensional output \eqref{eq:esn_output} suitable for the task at hand, we define $\mathbf{W}_{o} := \begin{pmatrix}1 & 0 & 0 & 0\\ 0 & 0 & 1 & 0 \end{pmatrix}$, which basically corresponds to a projection onto the first and third component, i.e., $ (x_1, x_2, x_3, x_4) \longmapsto (x_1, x_3)$.
Finally, we define the input matrix as
\begin{equation*}
\label{eq:W_4dim}
   \mathbf{W}_{in} := \omega_{in} \left[\begin{array}{cc} 0 & 0 \\ 1 & 0 \\ 0 & 0 \\ 0 & 1 \end{array}\right].
\end{equation*}
We considered the case of a 4-dimensional system composed by two independent, two-dimensional systems.
Nevertheless, the dynamics remains qualitatively the same even if the coupling between these two-dimensional systems is weak, i.e., if the zero matrices in \eqref{eq:M_4dim} are replaced by matrices with relatively small (absolute) values.
With those definitions in mind and, as before, $\boldsymbol{\epsilon}=0$ and $\alpha=1$, the ESN ruled by \eqref{eq:activation_update_fb2} and \eqref{eq:esn_output} correctly implements the flip-flop task with two bits.

As the dynamics of systems $(x_1, x_2)$ and $(x_3, x_4)$ are independent from each other, the set of fixed points of the overall dynamics turns out to be the Cartesian product of the set of fixed points of $(x_1, x_2)$ and the fixed points of $(x_3, x_4)$ system. This gives rise to a large number of fixed points: there are 4 stable points, 8 saddles with 1 unstable directions, 8 saddles with 2 unstable directions, 4 saddles with 3 unstable directions and 1 repeller.
However, the set of fixed points where the nonautonomous flip-flop dynamics take place is formed by 4 stable fixed points and 8 saddles with only one unstable direction; see Sec. \ref{sec:exp-manually-designed} for a detailed example.
Every stable fixed point is close to a pair of saddles and, due to symmetry, they are all at the same distance $\delta_{th}(\mathbf{p}_1,\mathbf{p}_2)$.
This quantity defines the excitability thresholds of the connections needed in the flip-flop task, and therefore implicitly defines the ENA ruling the behaviour of the four-dimensional ESN.

%%%%%%%%%%
%%%%%%%%%%
\section{Extracting ENAs from the ESN trajectory}
\label{sec:ena-extraction}

\begin{figure}[ht!]
\centering
\includegraphics[keepaspectratio=true,scale=0.55]{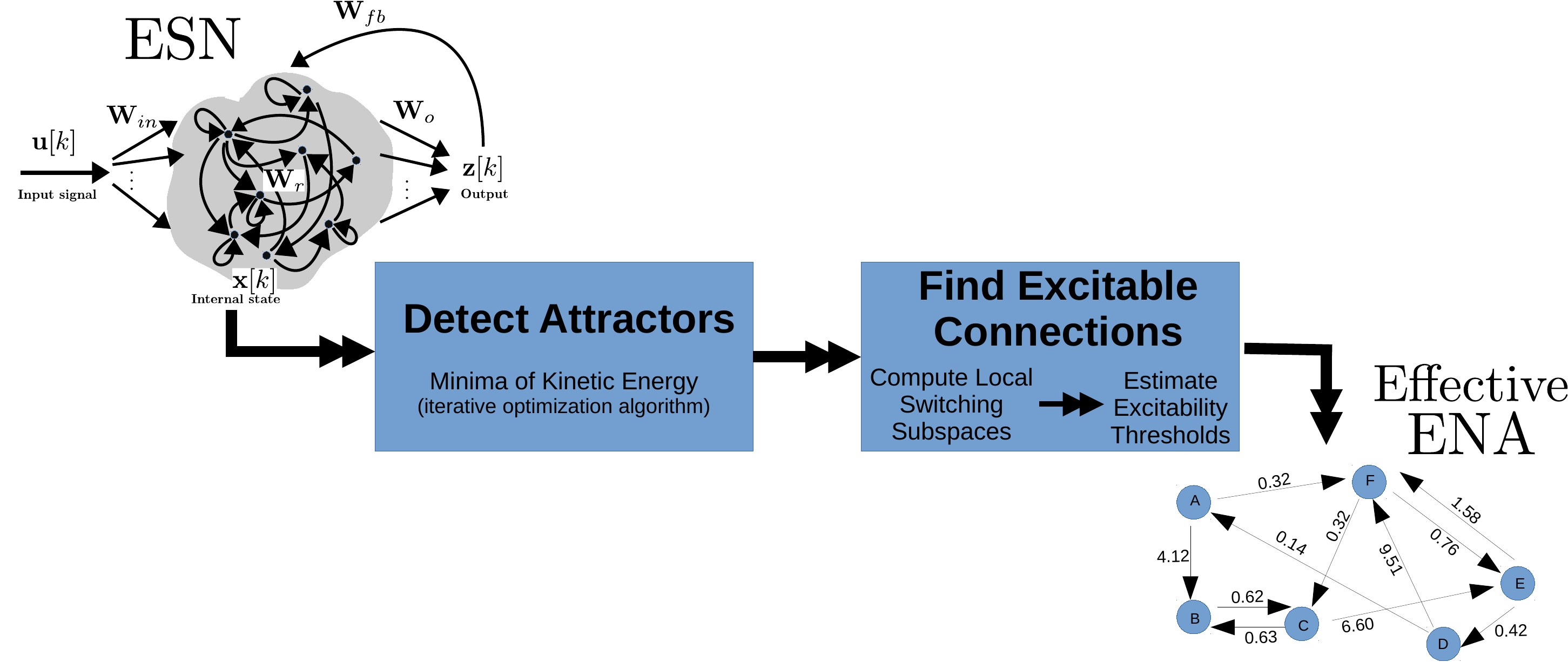}
\caption{Illustration of the proposed method to extract ENAs from trajectories.}
\label{fig:scheme_method} 
\end{figure}

In this section, we describe the proposed algorithm to extract an ENA from an ESN trajectory. The proposed algorithm, which is schematically illustrated in Fig. \ref{fig:scheme_method}, takes the trained reservoir matrix $\mathbf{M}$ \eqref{eq:trained_reservoir} and a trajectory of the nonautonomous ESN \eqref{eq:activation_update_fb2} as input and produces a weighted directed graph, representing the ENA describing the ESN behaviour, as output.
The algorithm is composed of two main steps.
In Sec. \ref{sec:optimization-fixed-points}, we describe the procedure to find fixed points of the ESN dynamics (corresponding to the vertices of the graph).
Successively, in Sec. \ref{sec:excitability-levels}, we show how to determine the excitable connections and related thresholds between stable fixed points (corresponding to the weighted directed edges).

\subsection{Finding fixed points of the dynamics}
\label{sec:optimization-fixed-points}

The optimisation algorithm we have used to find fixed points is based on \citet{sussillo2013opening}; see \cite{golub2018fixedpointfinder} for an open-source Tensorflow toolbox for finding fixed points in arbitrary RNN architectures and \cite{katz2018using} for an alternative method to identify fixed points.
The key idea is to define a scalar function whose minima correspond to fixed points of the ESN dynamics.

\begin{defn}
\label{def:def_veloc}
We call \emph{velocity field} of the autonomous system \eqref{eq:autonomous_system} the vector field $\mathbf{Q} : \mathbb{R}^{N_r} \longrightarrow \mathbb{R}^{N_r}$, defined as
    \begin{equation}
    \label{eq:def_veloc}
        \mathbf{Q}(\mathbf{x}):= \mathbf{F}(\mathbf{x}) - \mathbf{x},
    \end{equation}
with $\mathbf{F}$ being the map in \eqref{eq:autonomous_system}.
\end{defn}
Therefore, the velocity field takes the following form:
\begin{equation}
\label{eq:vel_field_express}
\mathbf{Q}(\mathbf{x})= \alpha \bigl[ \tanh(\mathbf{M}\cdot\mathbf{x}) - \mathbf{x} \bigl],
\end{equation}
where $\mathbf{Q}(\mathbf{x}[k])$ is the vector that needs to be added to the current state $\mathbf{x}[k]$ to obtain the next one.
In fact,
\begin{equation}
\begin{split}
\mathbf{x}[k+1]=\mathbf{F}(\mathbf{x}[k]) \Longleftrightarrow \mathbf{x}[k+1] - \mathbf{x}[k]=\mathbf{F}(\mathbf{x}[k]) - \mathbf{x}[k]  \Longleftrightarrow \mathbf{x}[k+1] = \mathbf{x}[k] + \mathbf{Q}(\mathbf{x}[k]).
\end{split}
\end{equation}

\begin{defn}
\label{def:kinetic_en}
We define \emph{kinetic energy} of the autonomous system \eqref{eq:autonomous_system} to be the following scalar function,
    \begin{equation}
    \label{eq:kinetic_en}
        q(\mathbf{x}) :=\dfrac{1}{2}\lVert\mathbf{Q}(\mathbf{x})\rVert^2.
    \end{equation}
\end{defn}
Note that fixed points $\mathbf{x}^* \in \mathbb{R}^{N_r}$ satisfy $\mathbf{Q}(\mathbf{x}^*)=\mathbf{0}$ if and only if $q(\mathbf{x}^*)=0$. Fixed points are hence identified as the global minima of \eqref{eq:kinetic_en}.
We use the quasi-Newton algorithm BFGS \cite{nocedal06Numericaloptimization} to minimise \eqref{eq:kinetic_en}.
In order to speed-up the optimisation by several orders of magnitude, we explicitly provided the gradient of \eqref{eq:kinetic_en} to BFGS, which reads:
\begin{equation}
\nabla q(\mathbf{x}_0) =  \mathbf{J}_{\mathbf{Q}}(\mathbf{x}_0)^{T} \mathbf{Q}(\mathbf{x}_0)  \,\, = \,\, \alpha \bigl( \mathbf{D}(\mathbf{x}_0) \mathbf{M} - \mathbf{I}_{N_r}  \bigl)^{T}  \mathbf{Q}(\mathbf{x}_0),
\end{equation}
where $\mathbf{I}_{N_r}$ is an $N_r\times N_r$ identity matrix, $ \mathbf{J}_{\mathbf{Q}}(\mathbf{x}_0) $ denotes the Jacobian matrix of the velocity field \eqref{eq:vel_field_express} and $ \mathbf{D}(\mathbf{x}_0) $ is a diagonal matrix defined in \eqref{eq:diagonal_squash} of Appendix \ref{sec:linear-stability}, both evaluated on $ \mathbf{x}_0 $.

The initial conditions for minimising \eqref{eq:kinetic_en} are determined by randomly sampling states from a trajectory of the nonautonomous ESN \eqref{eq:activation_update_fb2}. The convergence of the BFGS algorithm depends on a tolerance.
In fact, the algorithm may converge to similar solutions that, depending on chosen tolerance, are numerically different.
As these solutions represent fixed points of the dynamics, we aggregate them in a meaningful way and return a non-redundant set of fixed points. For this purpose, as post-processing step, we run a clustering algorithm on the set of all solutions and return only cluster representatives; details provided in Appendix \ref{sec:fixed_point_aggregation}.

\subsection{Determining excitable connections between attractors}
\label{sec:excitability-levels}

Once stable fixed points have been identified, we determine the excitable connections between them.
As discussed before (and in more detail in Appendix \ref{sec:bifurcations}) , the ESN training \eqref{eq:trained_reservoir} induces bifurcations that generate new fixed points; and possibly also other attracting regions in phase space that, however, are not explicitly modelled in this work.
The ESN is driven by control inputs that allow us to correctly perform switches between stable states.
As a consequence, we first need to understand how such inputs affect the dynamics from a geometric point of view.
This is done in Sec. \ref{sec:switch_subspaces} by introducing the notion of local switching subspace (LSS), which is strictly related to the notion of local input response set, \eqref{eq:loc_inp_resp_set}, introduced in Sec. \ref{sec:na}.
Then, in Sec. \ref{sec:grid} we describe how we determine all excitable connections that are relevant for the task under consideration and compute their excitability thresholds.
The method we propose is based on a grid of points lying in the LSS, accounting for the input action on the dynamics. By simulating the autonomous dynamics with initial conditions taken from such a grid, we are able to approximate input-driven excitability thresholds \eqref{eq:inp_driven_threshold} and also to quantify how likely it is that the RNN uses such connections while solving the task.

\subsubsection{Local switching subspaces}
\label{sec:switch_subspaces}

\begin{defn}
\label{def:pdvs}
Let us consider an ESN trajectory \eqref{eq:activation_update_fb2}, $\mathbf{x}[0], \mathbf{x}[1], \mathbf{x}[2], \ldots, \mathbf{x}[k], \ldots$, obtained with inputs $\mathbf{u}[1], \mathbf{u}[2], \ldots, \mathbf{u}[k+1], \ldots$.
Moreover, let us denote with $\mathcal{K} :=\{k_i\}_{i \in \mathbb{N}}$ the set of indices for which $\mathbf{u}[k_i] \neq \mathbf{0}$.
We define \emph{pulse difference vector} (PDV) a vector containing the difference between pre- and post-input states, namely  
    \begin{equation}
    \label{eq:pdvs}
        \mathbf{x}[k_i] - \mathbf{x}[k_i-1], \quad k_i \in \mathcal{K}.
    \end{equation}
\end{defn}
The PDVs \eqref{eq:pdvs} convey relevant information about the action of inputs on ESN state and we exploit such information to determine the excitable connections and related thresholds.
\begin{remark}
We remind the reader that the number of non-zero inputs is controlled by the parameter $p$ of the exponential distribution (see \eqref{eq:task} and related discussion in the text), which in turn determines the (average) number of PDVs available for the following analysis.
\end{remark}

\begin{figure}[ht!]
\centering
\includegraphics[keepaspectratio=true,scale=0.56]{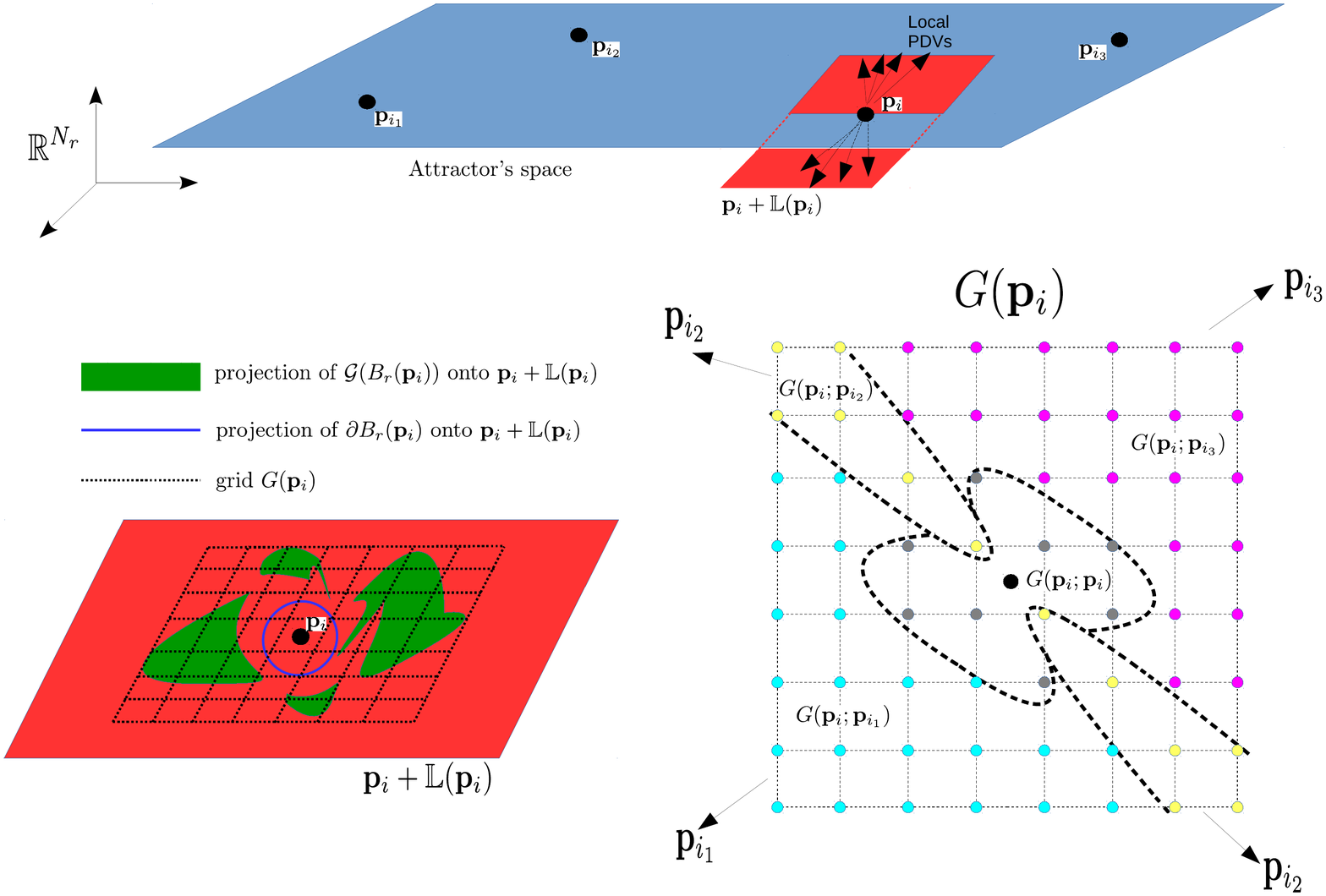}
\caption{\textbf{Top:} in blue, the low-dimensional space containing the attractors; black arrows represent local PDVs \eqref{eq:pdvs} originating from the attractor $\mathbf{p}_i$ which in turn define the LSS of $\mathbf{p}_i$, sketched in red. 
\textbf{Bottom left:} representation of the local input response set $\mathcal{G}( B_{r}( \mathbf{p}_i  ) )$, in green, and the grid $G( \mathbf{p}_i  )$, dashed line, in the LSS of $\mathbf{p}_i$, in red.
\textbf{Bottom right:} illustration of the partition of a two-dimensional grid $G( \mathbf{p}_i )$. Every colour represents a subset \eqref{eq:grid_subset_points}. Dashed black lines track the boundary of the basins corresponding to those attractors reachable from $\mathbf{p}_i$ through excitable connections, which can be enabled by inputs allowing exploration of the hypercube centred on $\mathbf{p}_i$.}
\label{fig:switchingsubspace} 
\end{figure}

In order to compute only those connections that are actually used by the ESN while solving the task, we need to focus on the action of inputs in the neighbourhood of each attractor.
To this end, we consider an Euclidean ball $B_{r}( \mathbf{p} )$ of radius $r\geq0$ centred on an attractor $\mathbf{p}$, and call \emph{local} PDVs of $\mathbf{p}$ the finite set of PDVs that originate inside $B_{r}(\mathbf{p})$.
Therefore, referring to \eqref{eq:loc_inp_resp_set}, the local PDVs of $\mathbf{p}$ is a collection of vectors originating in $B_{r}(\mathbf{p})$ and ending in $\mathcal{G}( B_{r}( \mathbf{p} )) $.
\begin{defn}
\label{def:LSS} 
Let $\mathbb{L}(\mathbf{p})$ be the vector space obtained by means of principal components analysis of the local PDVs \eqref{eq:pdvs} of the attractor $\mathbf{p}$, retaining only $l\ll N_r$ principal components.
We define the \emph{local switching subspace} (LSS) of $\mathbf{p}$, and we denote it as $\mathbf{p} + \mathbb{L}(\mathbf{p})$, the affine space composed by attaching the vector space $\mathbb{L}(\mathbf{p})$ over $\mathbf{p}$; see Fig. \ref{fig:switchingsubspace}, top panel.
\end{defn}

\subsubsection{Estimation of excitability thresholds}
\label{sec:grid}

The idea is to sample the LSS of a stable point $\mathbf{p}_i$ and describe it as a grid of points $G(\mathbf{p}_{i})$. Then, we consider those points on the grid as initial conditions for the autonomous system \eqref{eq:autonomous_system}, which is then iterated for a sufficiently large number of steps to ensure convergence to nearby attractors. Finally, tracing the evolution of these initial conditions allows us to estimate excitability thresholds and other relevant quantities.

The proposed algorithm for finding excitability thresholds is graphically illustrated in Fig. \ref{fig:switchingsubspace}.
The algorithm is based on a hypercube $H(\mathbf{p}_{i})$ centred on attractor $\mathbf{p}_{i}$ that is contained within the LSS $\mathbf{p}_{i} + \mathbb{L}(\mathbf{p}_{i})$.
The length of the hypercube is such that $H(\mathbf{p}_{i})$ contains the projection of $\mathcal{G}( B_r(\mathbf{p}_i) )$ on $\mathbf{p}_{i} + \mathbb{L}(\mathbf{p}_{i})$.
In order to estimate excitability thresholds, we consider a mesh with a pre-defined density of points on the hypercube $H(\mathbf{p}_{i})$, thus obtaining a grid of points, $G(\mathbf{p}_{i}) = \{ \mathbf{g}_{j}^{i} \}$. To simplify the notation, we use a single index $j$ to enumerate points of the grid.
Through the $\omega$-limit set \eqref{eq:omega_limitset} of the grid $G(\mathbf{p}_{i})$, that is:
$$
\omega_{\mathbf{F}}(G(\mathbf{p}_{i})) = \bigcup_{j} \omega_{\mathbf{F}}(\mathbf{g}_j^{i}),
$$
we can compute the following nonnegative integer $c(\mathbf{p}_{i}):= | \omega_{\mathbf{F}}(G(\mathbf{p}_{i})) | - 1 $, which counts the number of excitable connections originating from $\mathbf{p}_{i}$ which are allowed to be activated through input.
Indeed, $\omega_{\mathbf{F}}(G(\mathbf{p}_{i}))$ is composed of a set of stable fixed points, $\{ \mathbf{p}_{i_0}, \mathbf{p}_{i_1}, \mathbf{p}_{i_2}, \ldots, \mathbf{p}_{i_{c(\mathbf{p}_i)}} \}$, determining the endpoints of the $c(\mathbf{p}_i)$ different excitable connections.
\begin{remark}
Note that if the grid is sufficiently dense then the attractor itself is always included in such a set of fixed points. However we do not count this as an excitable connection. In what follows, we assume that the attractor is indexed by $i_0$, i.e., $\mathbf{p}_{i_0} = \mathbf{p}_{i}$.
\end{remark}

With these definitions in mind, we are ready to compute thresholds of excitable connections.
Let us denote with
\begin{equation*}
\label{eq:sigma_partition}
    \sigma_i: \{1,\ldots, |G(\mathbf{p}_{i})| \} \longrightarrow \{i_0, i_1,\ldots, i_{c(\mathbf{p}_{i})} \},
\end{equation*}
an indexing function such that $\mathbf{p}_{\sigma_i(j)} = \omega_{\mathbf{F}}( \mathbf{g}_j^i )$. Through the preimage $\sigma_i^{-1}(\cdot)$, we obtain a partition of points of the grid into the following subsets:
\begin{equation}
\label{eq:grid_subset_points}
   G(\mathbf{p}_{i}; \mathbf{p}_{i_t} ) :=  \{ \mathbf{g}_j^i \in G(\mathbf{p}_{i}) \,\,\, | \,\,\, j \in  \sigma_i^{-1}(i_t)  \}, \ t = 0, 1, \ldots, c(\mathbf{p}_{i}).
\end{equation}
The points of the grid belonging to the subset defined by \eqref{eq:grid_subset_points} are all destined to converge to $\mathbf{p}_{i_t}$. Therefore, we estimate the input-driven excitability threshold \eqref{eq:inp_driven_threshold} of the connection from $\mathbf{p}_i$ to $\mathbf{p}_{i_t}$ as follows:
\begin{equation}
\label{eq:estimate_excitab}
    \widetilde{\delta}_{\text{inp}}(\mathbf{p}_i, \mathbf{p}_{i_t}) = \min \left\{ \lVert\mathbf{g}_j^i - \mathbf{p}_i \rVert\,\,\, : \,\,\, \mathbf{g}_j^i  \in G(\mathbf{p}_{i}; \mathbf{p}_{i_t} )  \right\}, \quad t=1, \ldots,  c(\mathbf{p}_{i}).
\end{equation}

The excitability thresholds \eqref{eq:estimate_excitab} represent geometric properties of the attractors and related basins in phase space learned through training.
In order to determine the effective excitability (accounting for inputs) of each connection outgoing from $\mathbf{p}_i$, we exploit the local topology of the LSS of $\mathbf{p}_{i}$ by means of the grid partition \eqref{eq:grid_subset_points} induced by $\sigma_i(\cdot)$.
To this end, we define the ratio of initial conditions taken from the grid that converged to attractor $\mathbf{p}_{i_t}$ as follows:
\begin{equation}
\label{eq:escap_vol_ratio}
    \nu_{i,i_t} := \dfrac{|G(\mathbf{p}_{i}; \mathbf{p}_{i_t} )|}{|G(\mathbf{p}_{i} )| - |G(\mathbf{p}_{i}; \mathbf{p}_i)|} \,\,\, \in [0,1].
\end{equation}

In the limit of infinite number of points in the grid, the quantity \eqref{eq:escap_vol_ratio} converges to the ratio of volumes between the portion of the hypercube belonging to the basin of $\mathbf{p}_{i_t}$ and the portion of the hypercube that does not belong to the basin of $\mathbf{p}_{i}$.
Therefore, dense grids give ratios \eqref{eq:escap_vol_ratio} providing accurate information about how the LSS is distributed between basins of attraction of stable fixed points.
Finally, merging both \eqref{eq:estimate_excitab} and \eqref{eq:escap_vol_ratio} into a single expression, we define \emph{effective excitability} of the connection from $\mathbf{p}_i$ to $\mathbf{p}_{i_t}$ as follows:
\begin{equation}
\label{eq:beta}
    \beta_{i,i_t} :=  \dfrac{\nu_{i,i_t} }{ \widetilde{\delta}_{\text{inp}}(\mathbf{p}_i, \mathbf{p}_{i_t})}.
\end{equation}
Note that this quantity takes into account both the distance between the attractor $\mathbf{p}_{i}$ and the basin's boundary, and the volume of the basin itself.
A low value for $\beta_{i,i_t}$ indicates that it is difficult to activate the connection from $\mathbf{p}_i$ to $\mathbf{p}_{i_t}$ during the task. This may be due (i) to the small volume occupied by the basin of the attractor $\mathbf{p}_{i_t}$ in the LSS of $\mathbf{p}_{i}$ or (ii) to a very high excitability threshold associated with such a connection.
On the other hand, $\beta_{i,i_t} \gg 1$ necessarily implies that such a connection has a low excitability threshold, since $\nu_{i,i_t} \in [0,1]$.
As a consequence, the distance between the basin of attraction of $\mathbf{p}_{i_t}$ and $\mathbf{p}_{i}$ is small, thus the connection can be easily activated during the task.

\paragraph{Remarks on computational complexity.}
There are three parameters controlling the complexity and, accordingly, the accuracy of the search in the grid: the dimension $\zeta_1$ of the hypercube, the length $\zeta_2$ of the hypercube's edge, and the number of points $\zeta_3$ on each edge determining the density of the grid.
$\zeta_3$ and $\zeta_2$ have a linear and polynomial impact on the computational complexity of the algorithm, respectively. However, $\zeta_1$ is more critical as it increases exponentially the number of points in the grid and, accordingly, the overall complexity.
In the simulations we always set $\zeta_1=l$, that is, the dimension of the hypercube is equal to the dimension of the LSS which is low in our case. More generally, the dimension of LSSs depends on the complexity of the inputs and their impact on the dynamics, and this will need to be assessed on a case-by-case basis.

%%%%%%%%%%%%%
\section{Simulations}
\label{sec:experiments}

In this section we discuss the results of our simulations and relate this to our theoretical framework.
In Sec. \ref{sec:exp-manually-designed}, in order to evaluate the correctness of the algorithm proposed in Sec. \ref{sec:ena-extraction}, we apply it to the manually-designed, low-dimensional ESN maps discussed in Sec. \ref{sec:2D-example} and Sec. \ref{sec:4D-example}.
Sec. \ref{sec:high_dim_method} applies our method to high-dimensional trained ESNs.
We show that, even though the ESN is high-dimensional, the dynamics generated by the trained reservoir \eqref{eq:trained_reservoir} is effectively low-dimensional. We also show the usefulness of ENAs for giving a mechanistic interpretation of prediction errors occurring during task execution. Finally, in Sec. \ref{sec:noise}, we show how ENA models can be used to assess the robustness to noise of trained ESNs. For all simulations, we use $p=0.1$ as parameter of the exponential distribution governing the occurrence of input pulses and set the tolerance of the kinetic energy \eqref{eq:kinetic_en} for detecting minima to $10^{-6}$.

\subsection{Evaluation on manually designed low-dimensional ESNs}
\label{sec:exp-manually-designed}

For the grid search algorithm, we used $\zeta_1=2, \zeta_2=4, $ and $\zeta_3=223$.

\begin{figure}[ht!]
\begin{minipage}{\textwidth}
\centering
\includegraphics[width=\textwidth]{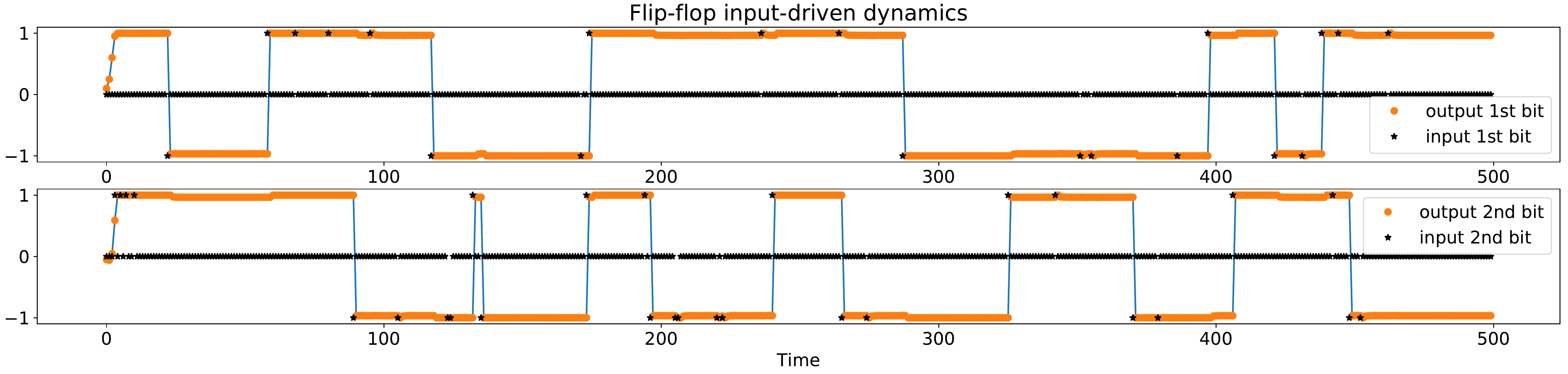}
\end{minipage}\\
\begin{minipage}{\textwidth}
\centering
    \includegraphics[keepaspectratio=true,scale=0.27]{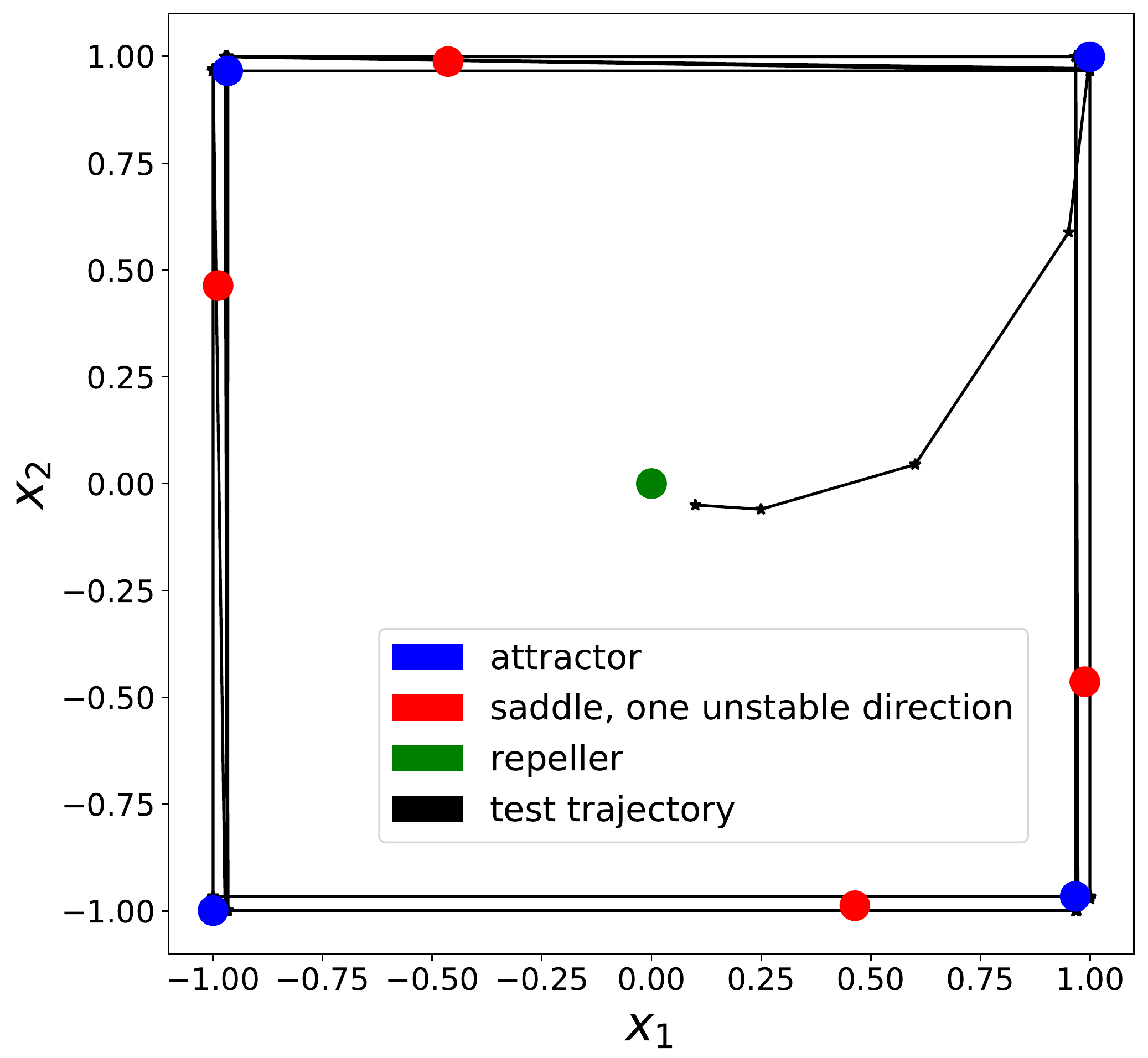}~\includegraphics[keepaspectratio=true,scale=0.25]{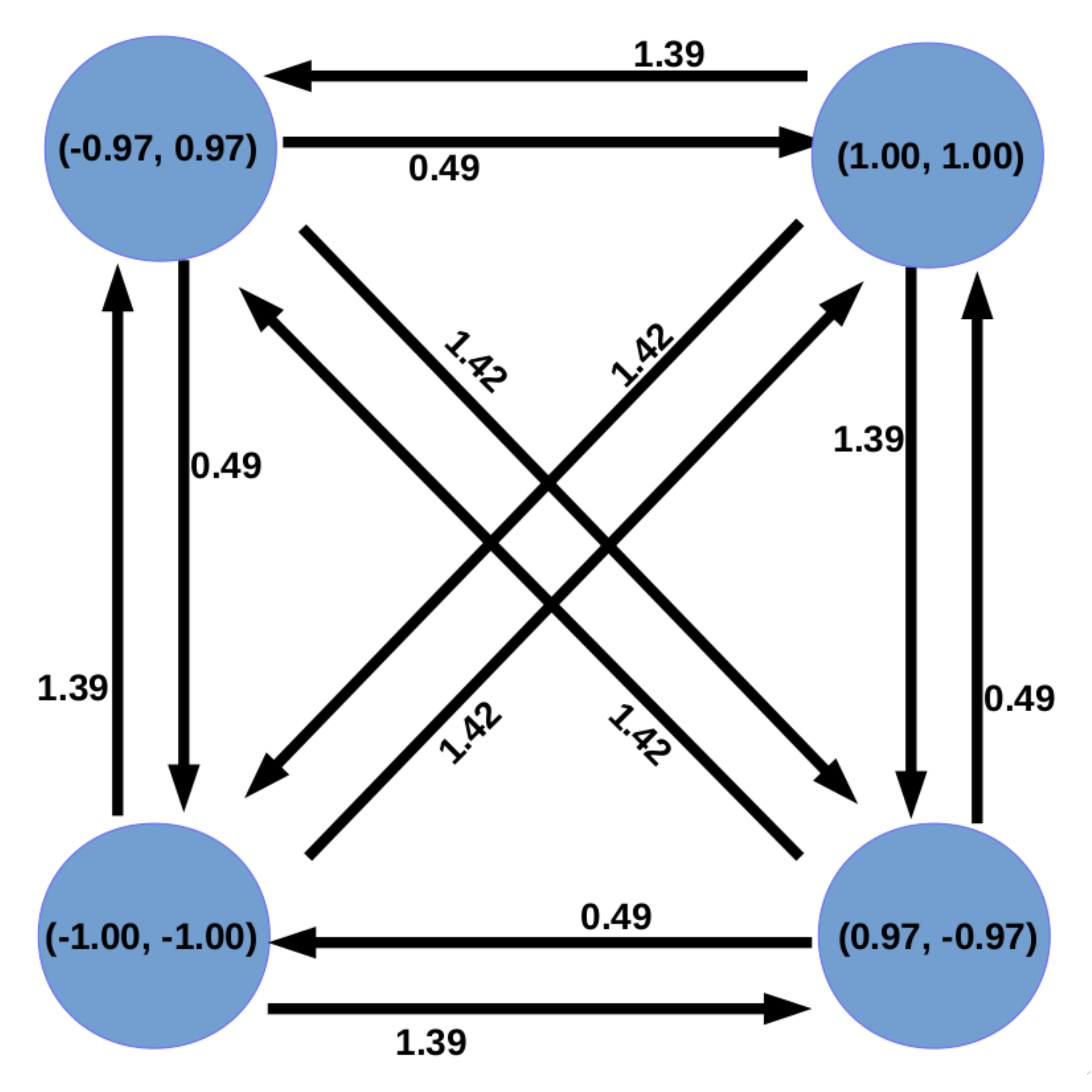}~\includegraphics[keepaspectratio=true,scale=0.25]{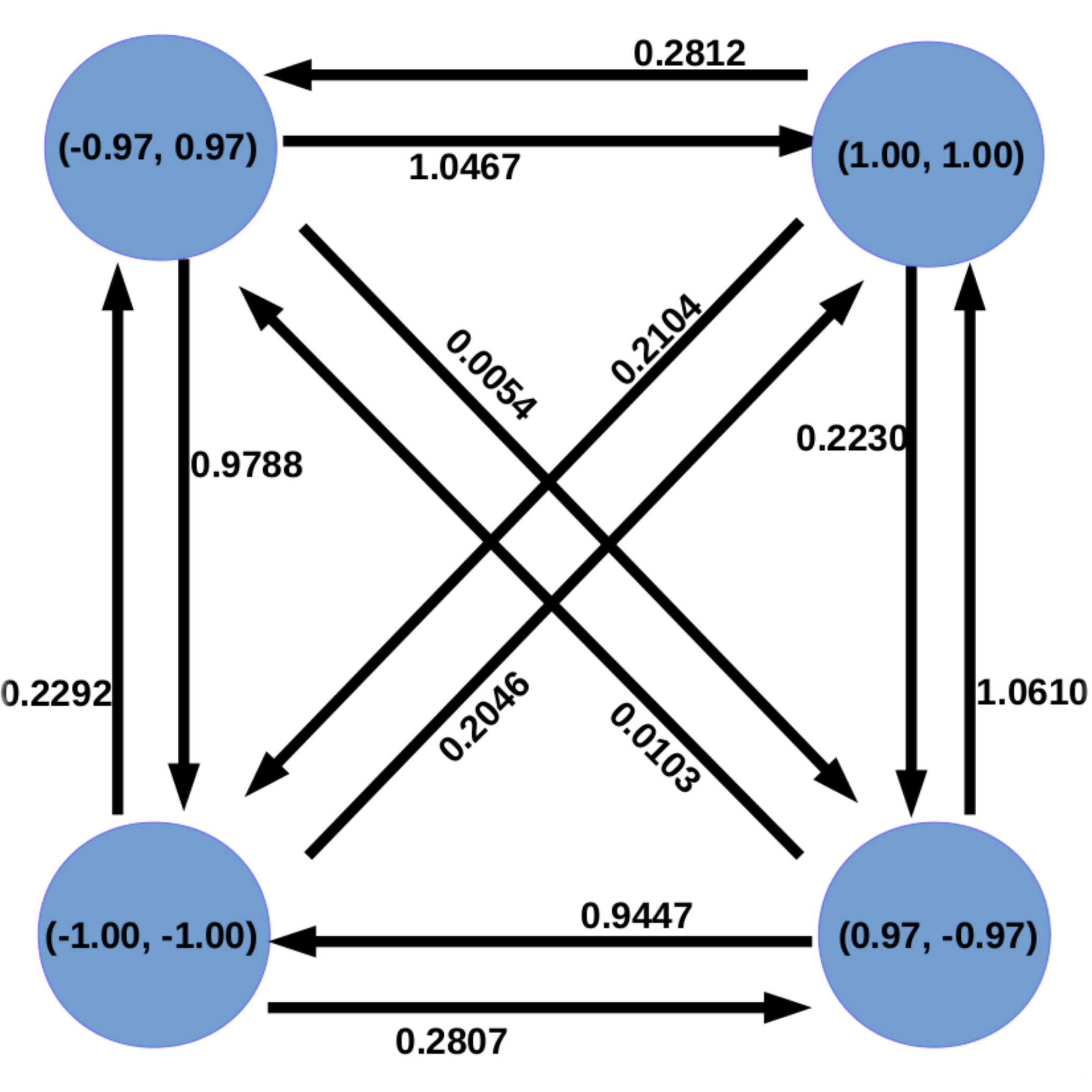}
\end{minipage}
\caption{\textbf{Top:} output produced by the two-dimensional ESN defined in Sec. \ref{sec:2D-example}. \textbf{Bottom left:} fixed points found by the optimisation algorithm with 100 initial conditions. Length of the test trajectory was $1000$ steps. \textbf{Bottom centre:} extracted ENA with edges weighted by excitability thresholds \eqref{eq:estimate_excitab}. \textbf{Bottom right:} extracted ENA with edges weighted according to \eqref{eq:beta}. Node labels represent output values \eqref{eq:esn_output} produced on the attractors.}
\label{fig:2D_ENA}
\end{figure}

\paragraph{Minimal-dimension model}
The LSS determined as described in Sec. \ref{sec:switch_subspaces} corresponds to the whole 2D phase space.
As a consequence, excitability thresholds \eqref{eq:threshold} match the corresponding input-dependent counterparts \eqref{eq:inp_driven_threshold}.
For each attractor, the Euclidean distances from the two closest saddles are consistent with the estimation of excitability thresholds provided by our method.
As discussed in Sec. \ref{sec:2D-example} and graphically represented in Fig. \ref{fig:2D_ENA} bottom-centre panel, we find three excitability thresholds in the computed ENA: $0.49, 1.39$, and $1.42$. In Fig. \ref{fig:2D_ENA}, bottom-right panel, we show that undesired connections (i.e., connections corresponding to state transitions that are not defined in the flip-flop task) between fixed points $(0.97,-0.97)$ and $(-0.97, 0.97)$ have very low effective excitability values \eqref{eq:beta}, indicating that it is unlikely to use such connections during the task execution.
The low effective excitability is due to the small volume ratio \eqref{eq:escap_vol_ratio} of the basins associated with the two attractors. On the other hand, larger thresholds characterise the undesired connections from $(-1,-1)$ to $(1,1)$, reflecting a lack of symmetry between the basin volumes of $(1,1), (-1,-1)$ and $(0.97, -0.97), (-0.97, 0.97)$.

\begin{figure}[ht!]
\begin{minipage}{\textwidth}
\centering
    \includegraphics[width=0.9\textwidth]{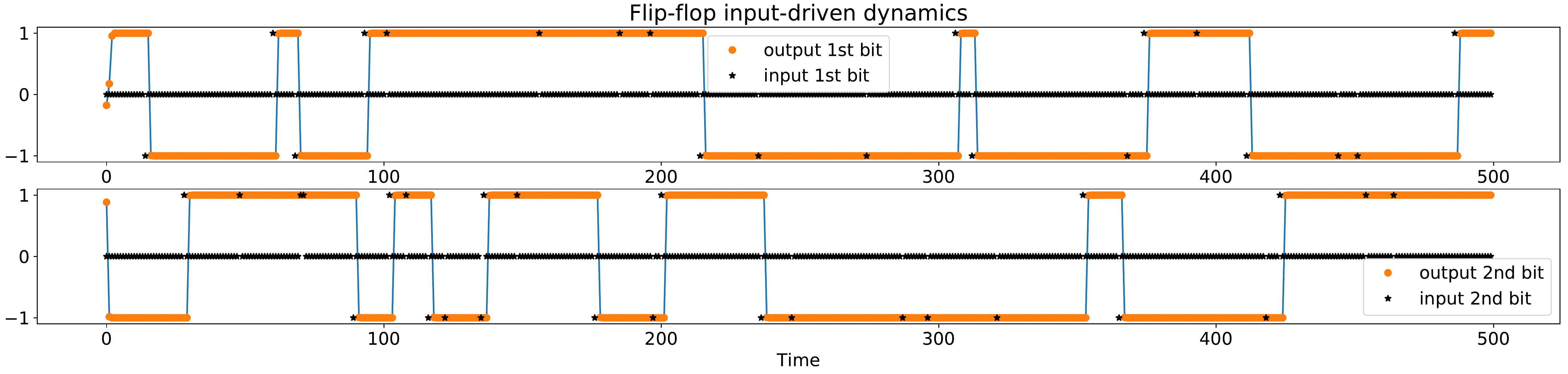}
\end{minipage}\\
\begin{minipage}{\textwidth}
\centering
    %\begin{flushright}
        \includegraphics[width=0.46\textwidth]{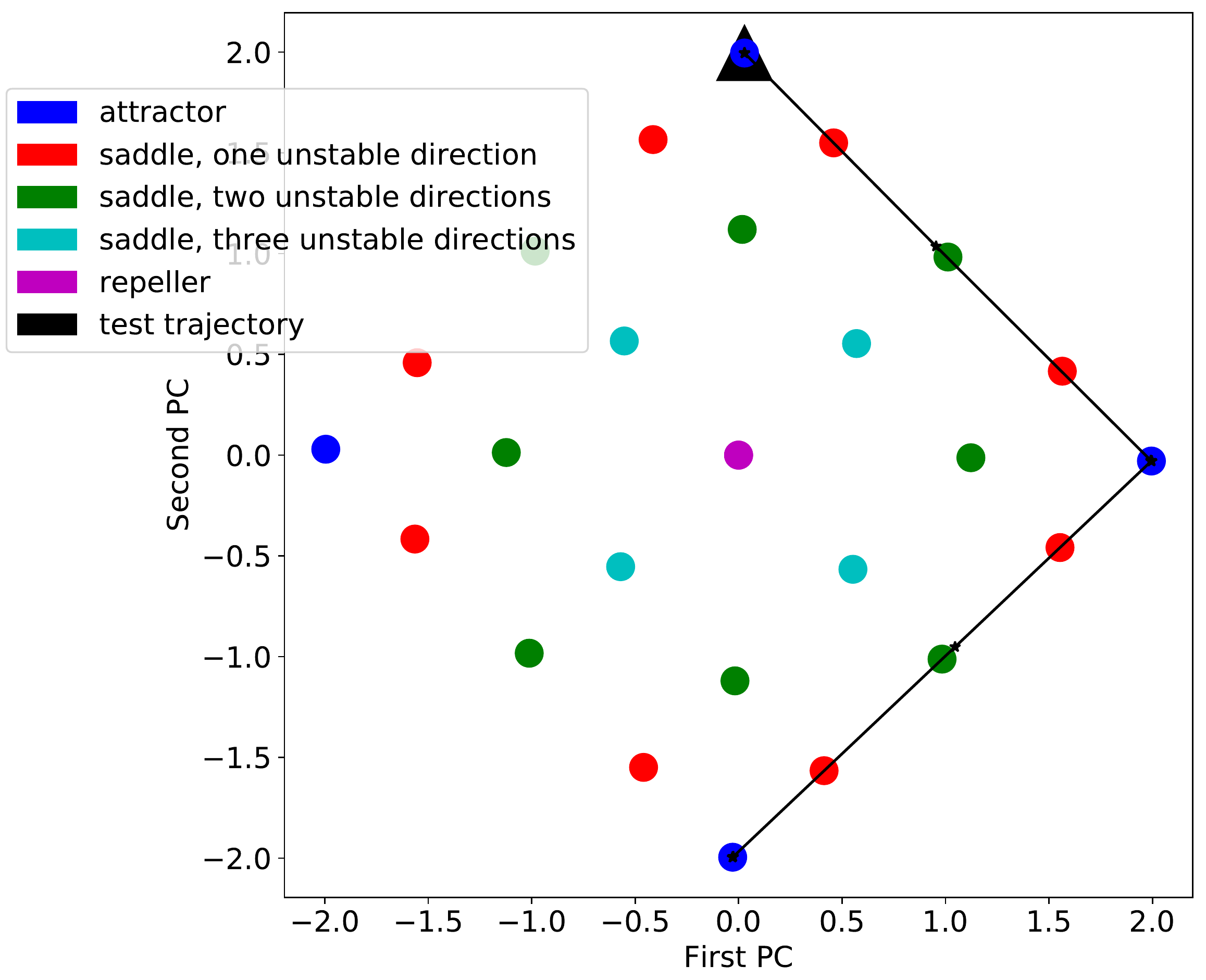}  \includegraphics[width=0.45\textwidth]{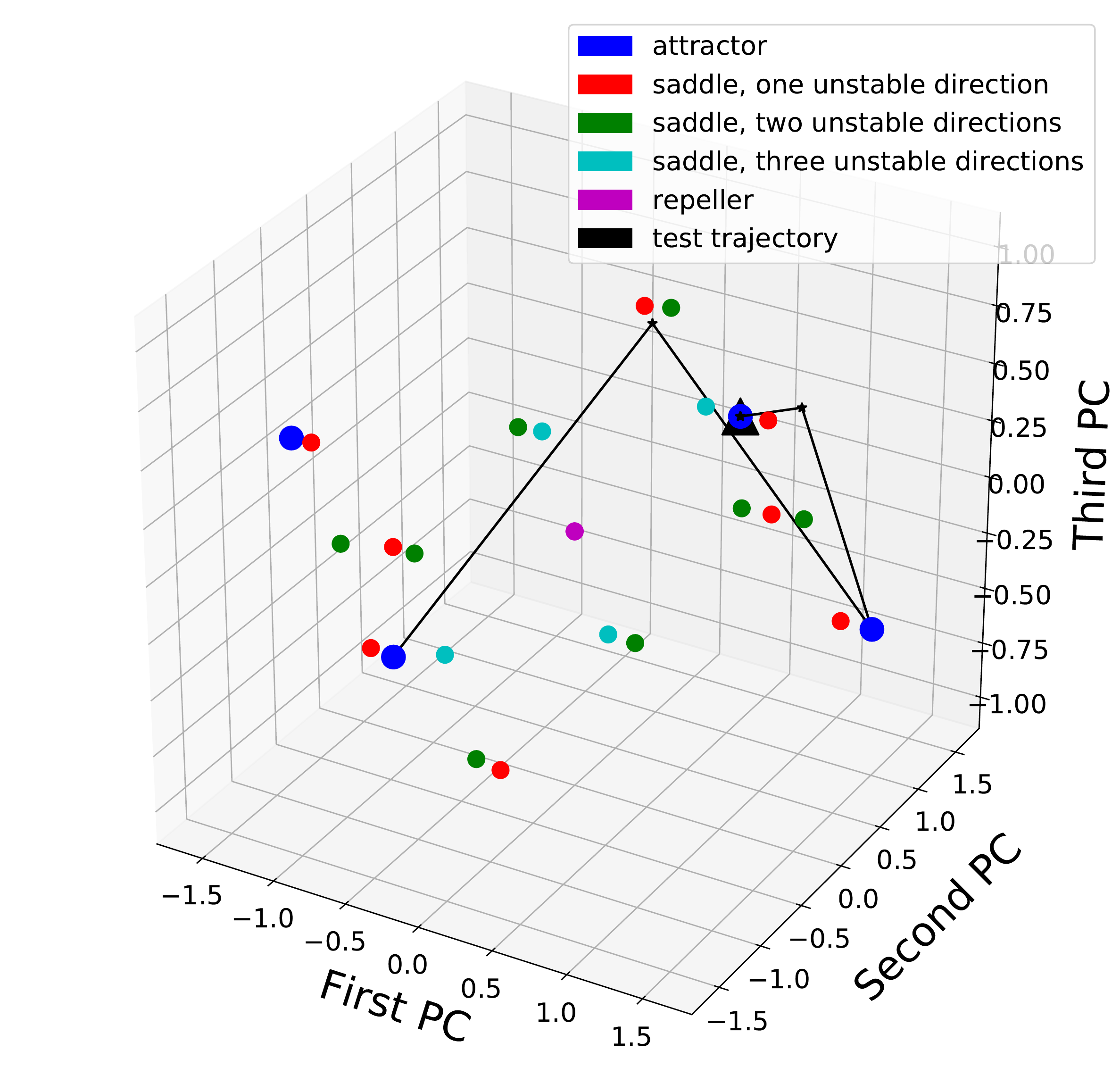}
    %\end{flushright}
\end{minipage}\\
\begin{minipage}{\textwidth}
\centering
    \includegraphics[keepaspectratio=true,scale=0.33]{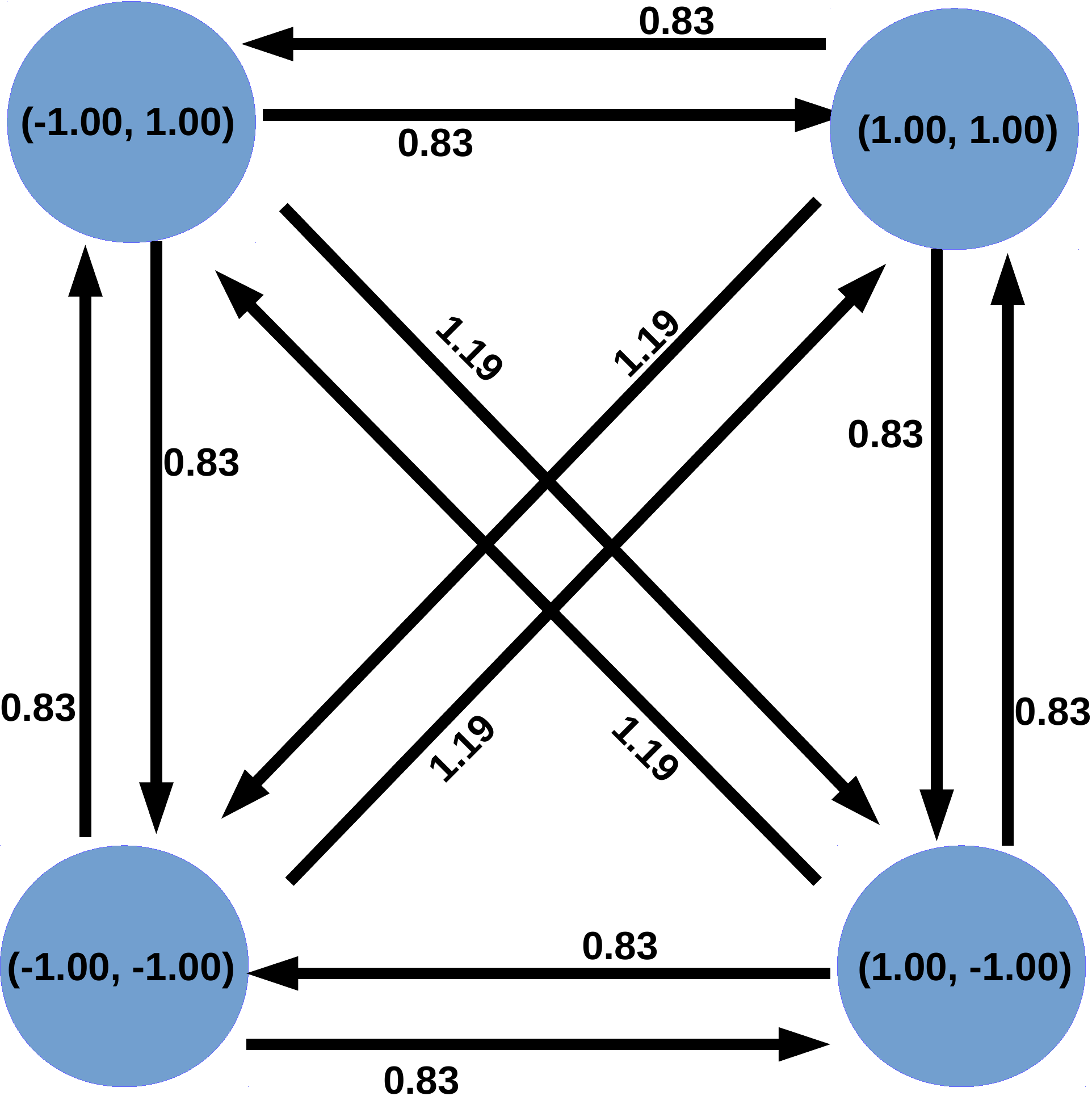}  \includegraphics[keepaspectratio=true,scale=0.33]{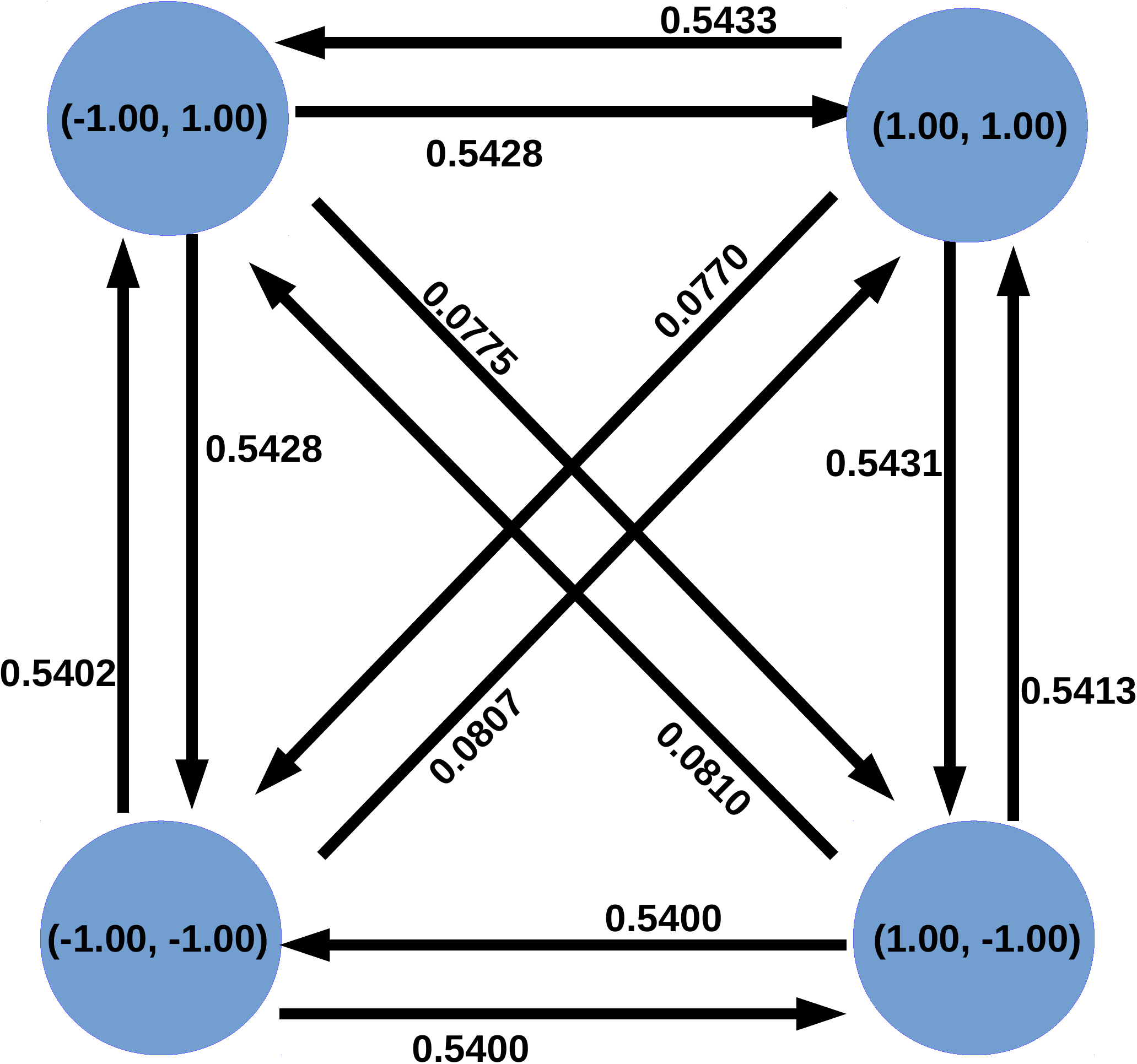}
\end{minipage}
\caption{\textbf{Top:} output of the four-dimensional ESN defined in Sec. \ref{sec:2D-example}. \textbf{Centre left:} fixed points found by the optimisation algorithm with 200 initial conditions, depicted on the space spanned by the first two principal components. In black, it is shown a trajectory starting on the attractor (marked by a black triangle); to improve readability, the trajectory is limited to two switches only. \textbf{Centre right:} fixed points and trajectory depicted in the centre-left picture are embedded in the space spanned by the three principal components. \textbf{Bottom left:} ENA with edge weights representing excitability thresholds \eqref{eq:estimate_excitab}. \textbf{Bottom right:} ENA with edge weights representing the effective excitability of connections \eqref{eq:beta}. Node labels represent output values produced by the ESN on the attractors.}
\label{fig:4D_ENA}
\end{figure}

\paragraph{Four-dimensional model}
The first two principal components obtained via principal component analysis are related to all identified fixed points and produce a cumulative variance ratio larger than $0.96$; Fig. \ref{fig:4D_ENA} shows a trajectory of the map described in Sec. \ref{sec:4D-example} and related fixed points projected on the plane spanned by the two principal components.
For every attractor, the computed LSS is a two-dimensional plane; this is expected, since it is actually the plane depicted in Fig. \ref{fig:4D_phasespace}. Furthermore, we note that none of these planes is aligned with the one where the attractors lie, stressing the importance of defining reference frameworks local to each attractor that take the action of inputs into account.
Fig. \ref{fig:4D_ENA}, middle-right panel, shows how the input moves the states out of the plane, using additional dimensions for the switches.
The computed ENA, weighted with excitability thresholds \eqref{eq:estimate_excitab} and effective excitability \eqref{eq:beta}, is shown in Fig. \ref{fig:4D_ENA}, bottom-left and bottom-right panels, respectively. Symmetries of the dynamical system are clearly present in the resulting directed graphs.
All desired connections are characterised by an excitability threshold of $\simeq 0.83$, while undesired ones have higher thresholds equal to $ \simeq  1.19$. 
We note that the presence of undesired connections does not imply that the ESN actually uses such connections during the task execution.
To this end, the effective excitability thresholds \eqref{eq:beta}, shown in the bottom-right panel of Fig. \ref{fig:4D_ENA}, provide us with a more realistic picture of the behaviour under the action of inputs.
Note that the effective excitability thresholds are very low for the undesired connections, implying that the LSS used by inputs is mostly occupied by basins corresponding to attractors adjacent to the end-point of desired connections.

\subsection{Application of the proposed method to high-dimensional trained ESNs}
\label{sec:high_dim_method}

We now consider implementation using ESNs \eqref{eq:activation_update_fb2} with a reservoir composed of 500 neurons. 
Experimenting with ESNs with different reservoir sizes confirm that one can get ESNs can be successfully trained independent of the precise number of neurons, as long as there are enough. We choose $500$ neurons for hardware and computing time considerations.
For the grid search algorithm, we use $\zeta_1=3, \zeta_2=18, $and $\zeta_3=12$.
The state-update \eqref{eq:activation_update_fb2} is configured without leakage, $\alpha = 1$ and standard deviation of noise $\boldsymbol{\epsilon}$ is set to $10^{-4}$ during training.
The entries of matrices $\mathbf{W}_{in}$, $\mathbf{W}_{fb}$, and $\mathbf{W}_{r}$ are i.i.d. drawn from a uniform distribution in $[-1,1]$; the sparseness of $\mathbf{W}_{r}$ is $95\%$. Moreover, the reservoir matrix was rescaled to obtain a spectral radius equal to $0.9$. Finally, the training set length is always $50000$ time-steps.

\begin{figure}[ht!]
\centering
\begin{tabular}{cc}
\multicolumn{2}{c}{\includegraphics[keepaspectratio=true,scale=0.31]{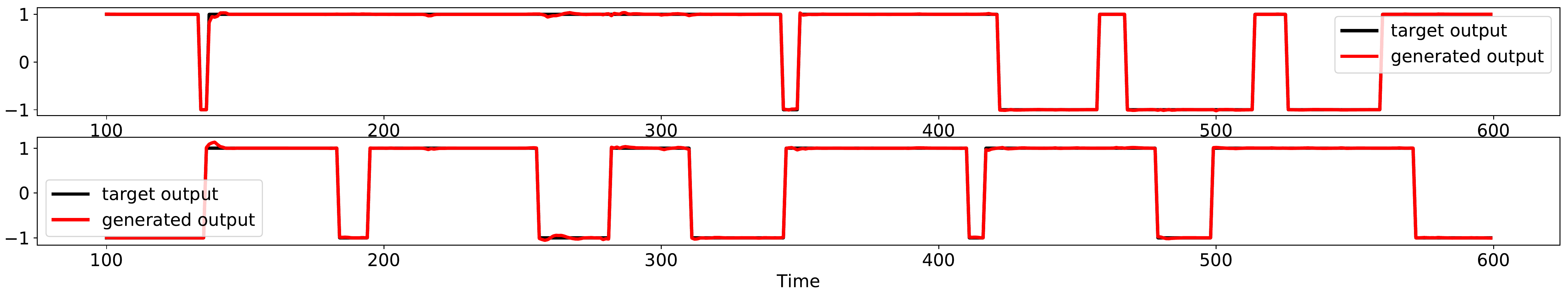}} \\
\includegraphics[keepaspectratio=true,scale=0.38]{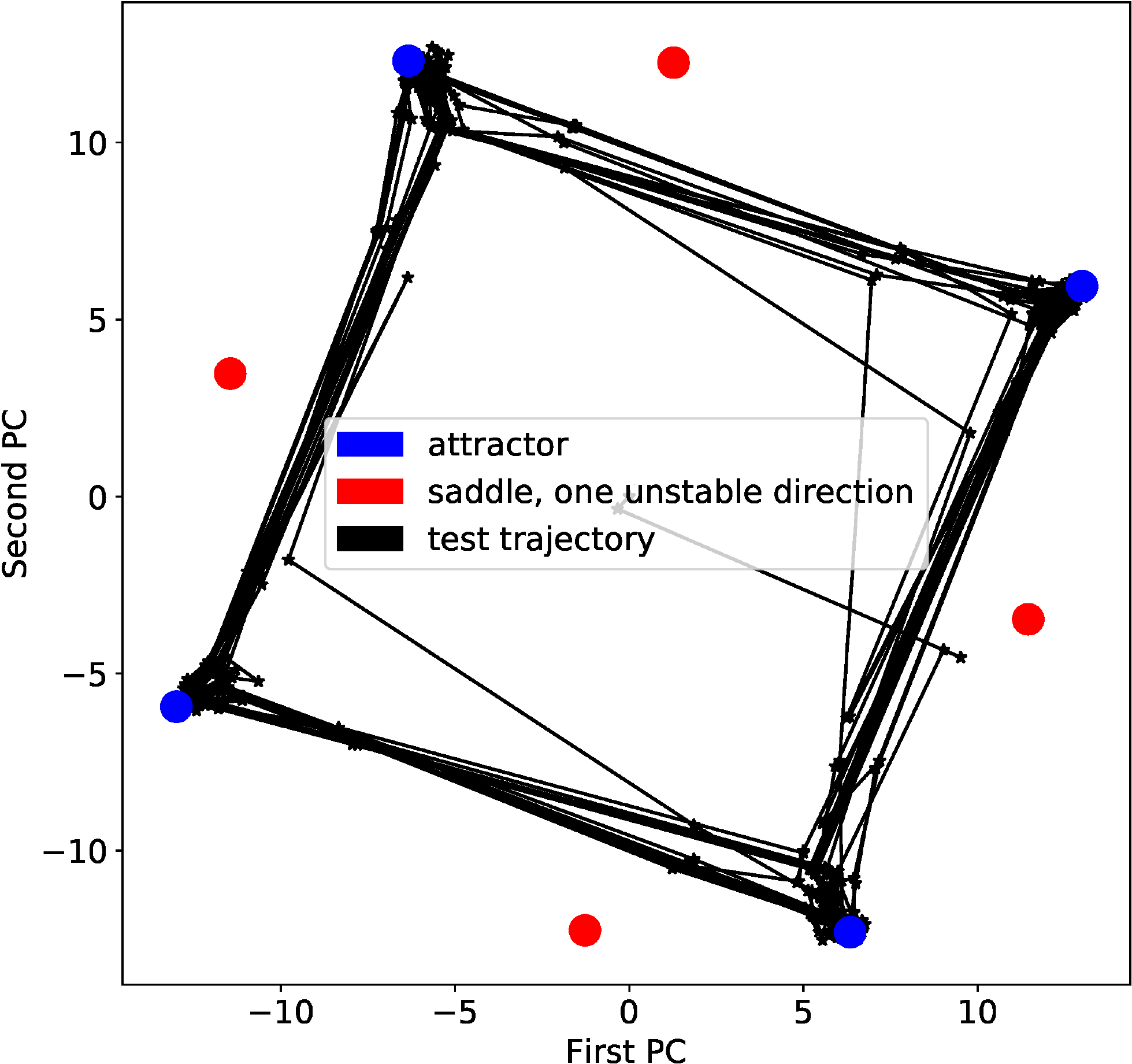} & \includegraphics[keepaspectratio=true,scale=0.36]{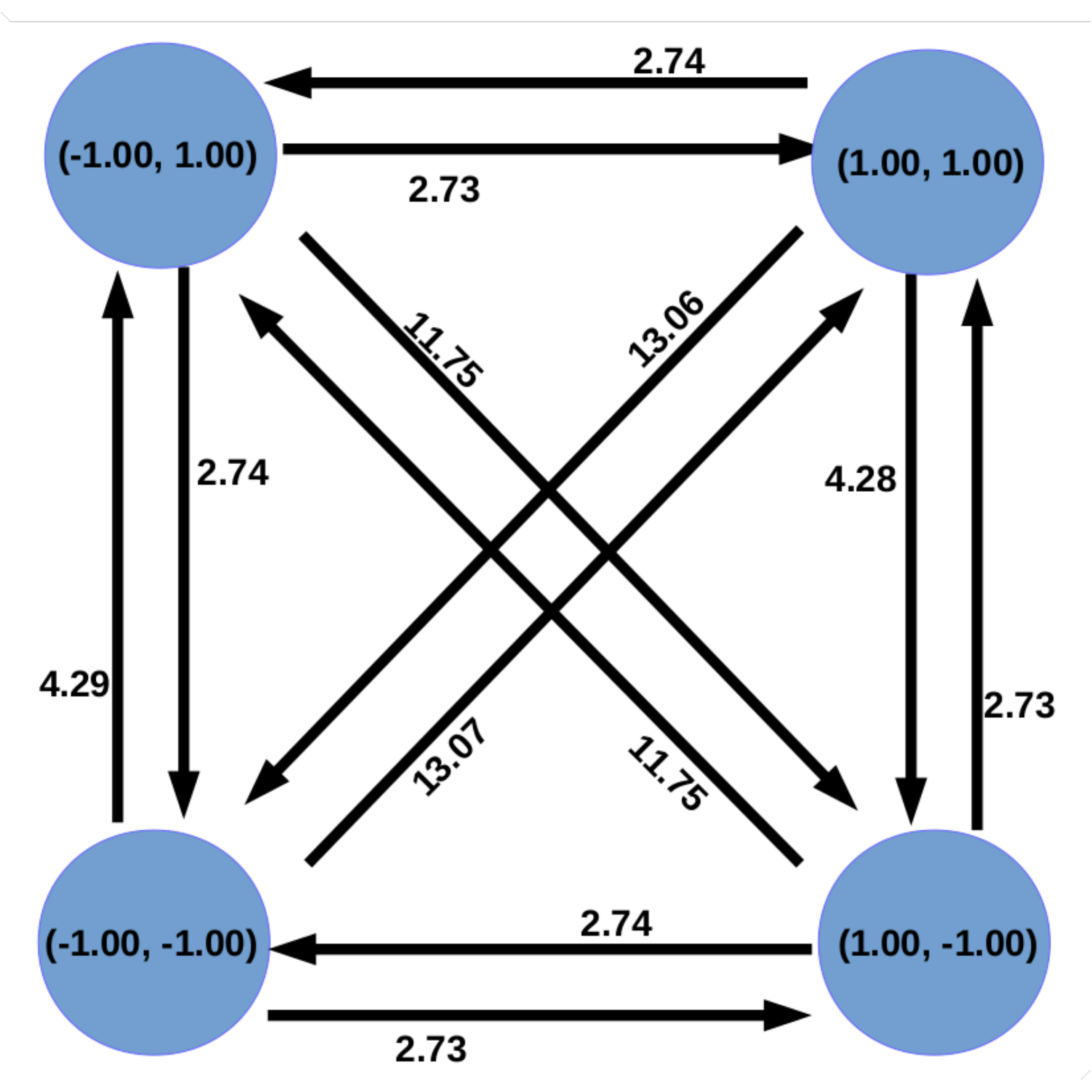}
\end{tabular}
\caption{\textbf{Top:} output produced by a trained ESN and target output. \textbf{Bottom left:} fixed point found by means of the optimisation algorithm with 500 initial conditions. The plane in the picture represents the space spanned by the first two principal components determined over all 500 solutions returned by the optimisation algorithm. \textbf{Bottom right:} extracted ENA with edge weights representing excitability thresholds \eqref{eq:estimate_excitab}. Node labels represent output values produced by the ESN \eqref{eq:esn_output} on the attractors.}
\label{fig:ESN_high_dim}
\end{figure}

\paragraph{Low-dimensional dynamics}
Fig. \ref{fig:ESN_high_dim}, top panel, shows the output produced by an ESN achieving high prediction performance.
The extracted ENA, shown in the bottom-right panel, reveals that undesired connections have excitability thresholds \eqref{eq:inp_driven_threshold} significantly higher than those of desired connections. This means that, in the LSS of every stable point, basins corresponding to attractors adjacent to the end-point of undesired connections stand relatively far away from the stable point compared to basins of attractors adjacent to the end-point of desired connections.
The fixed points of the dynamics lie in a two-dimensional subspace of the 500-dimensional phase space (the cumulative variance ratio is close 1).
It is observed that trajectory spends most of the time close to such a plane. Hence, based on a principal component analysis of the trajectory, we can claim that the dynamics is two-dimensional. 
Nevertheless, during the task execution, the trajectory is occasionally driven away from such a 2D plane by the inputs in order to achieve the switches between attractors, and this feature is crucial to understand how the trained neural network behaves while solving the task. 
The cumulative variance ratio for the identification of the LSS of every attractor, as described in Sec. \ref{sec:switch_subspaces}, revealed that the switching between stable fixed points takes place in a three-dimensional subspace of the phase space, highlighting that the overall dynamics of ESNs is effectively low-dimensional after training.
It is worth stressing that such LSSs are usually not aligned with the standard coordinate system of the original phase space and the subspaces where attractors lie, suggesting that inputs operate in phase space regions that are disjoint with respect to the low-dimensional linear subspace of the attractors; this is consistent with results reported in \cite{sussillo2013opening}.

\paragraph{Computation accuracy and spurious attractors}
Various measures of accuracy exist for quantifying the performance on prediction tasks. For instance, the mean-squared-error (MSE) is typically adopted in tasks involving continuous targets. The MSE is a real-valued scalar that informs us about how close the computed output is to the target one.
However, it is not possible to infer additional insights by looking only at the MSE; for instance, it is not possible to provide a mechanistic description of errors in the computation, i.e., why and how they occur.
Here, we show how the effective ENA extracted from the ESN trajectory can be used to describe how the computation takes place in phase space and, in particular, how to diagnose the nature of prediction errors.

\begin{figure}[ht!]
\begin{minipage}{\textwidth}
\centering
\includegraphics[width=0.9\textwidth]{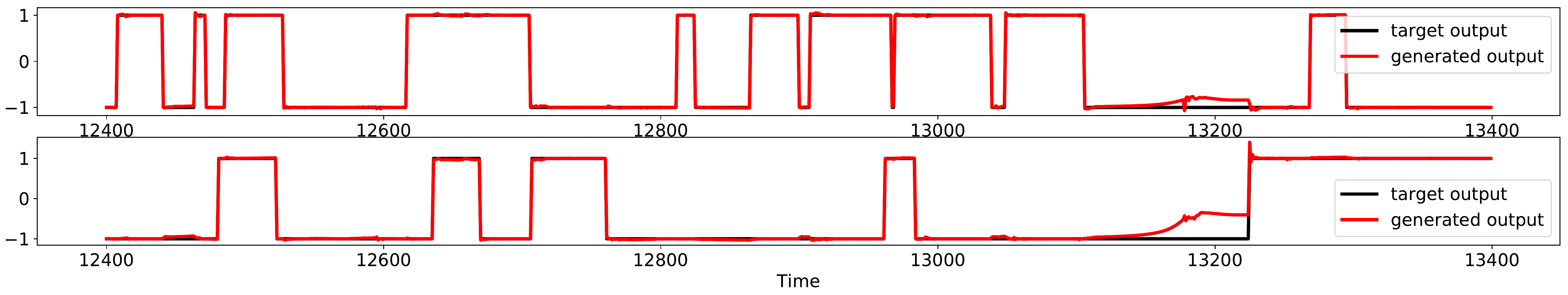}
\end{minipage}\\
\begin{minipage}{\textwidth}
\centering
    \includegraphics[width=0.45\textwidth]{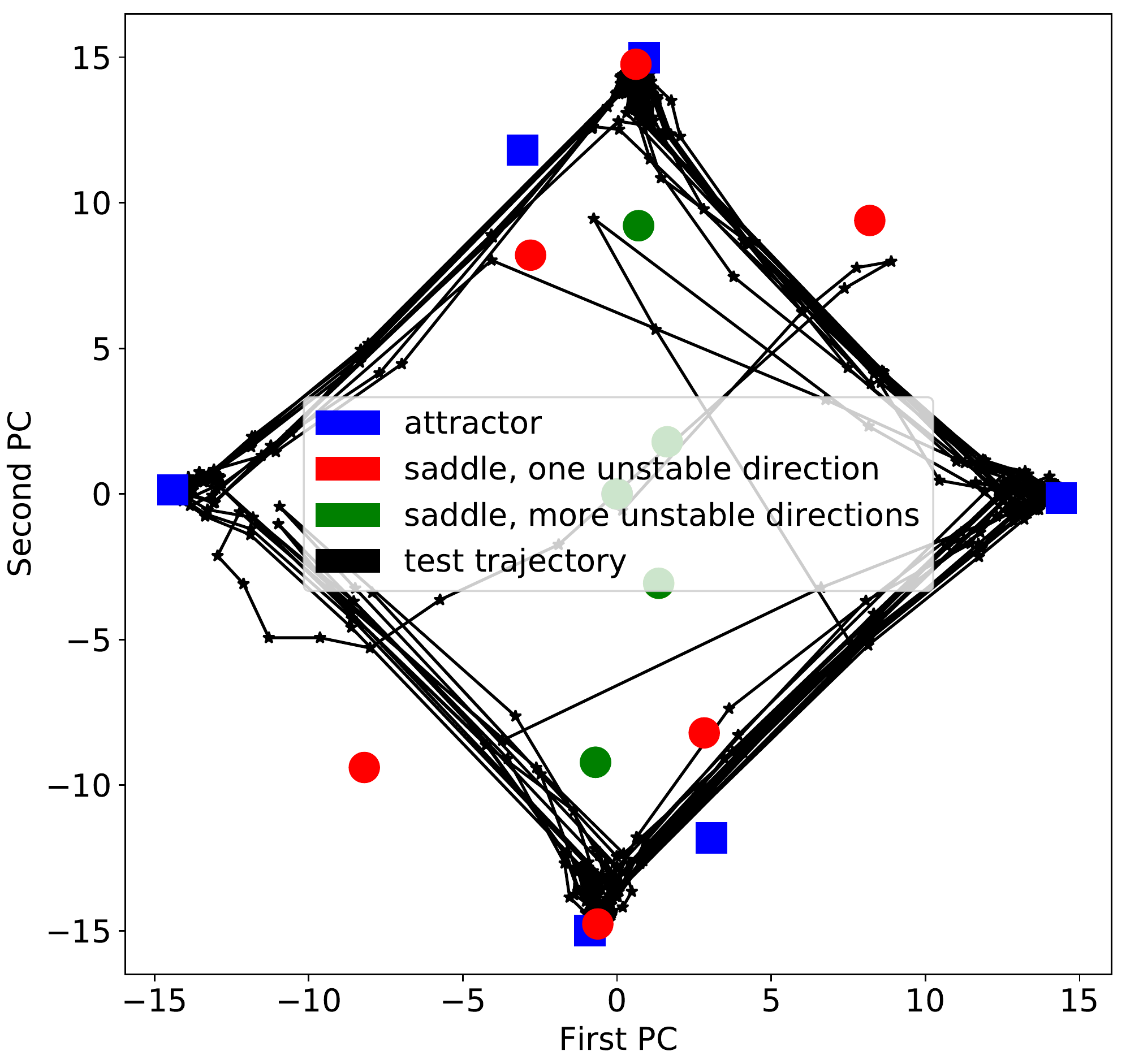}~\includegraphics[width=0.45\textwidth]{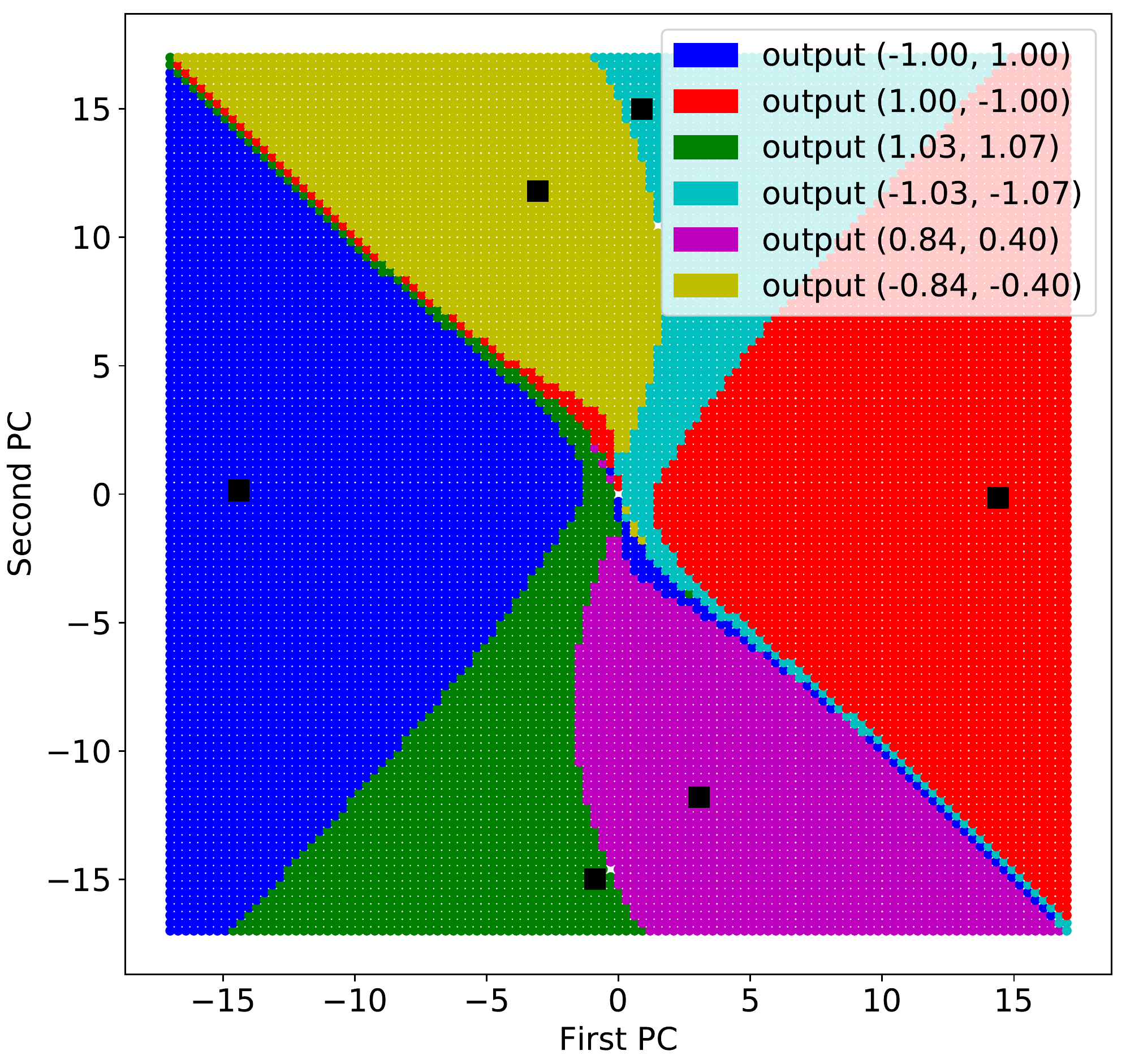}
\end{minipage}\\
\begin{minipage}{\textwidth}
\centering
    \includegraphics[width=0.4\textwidth]{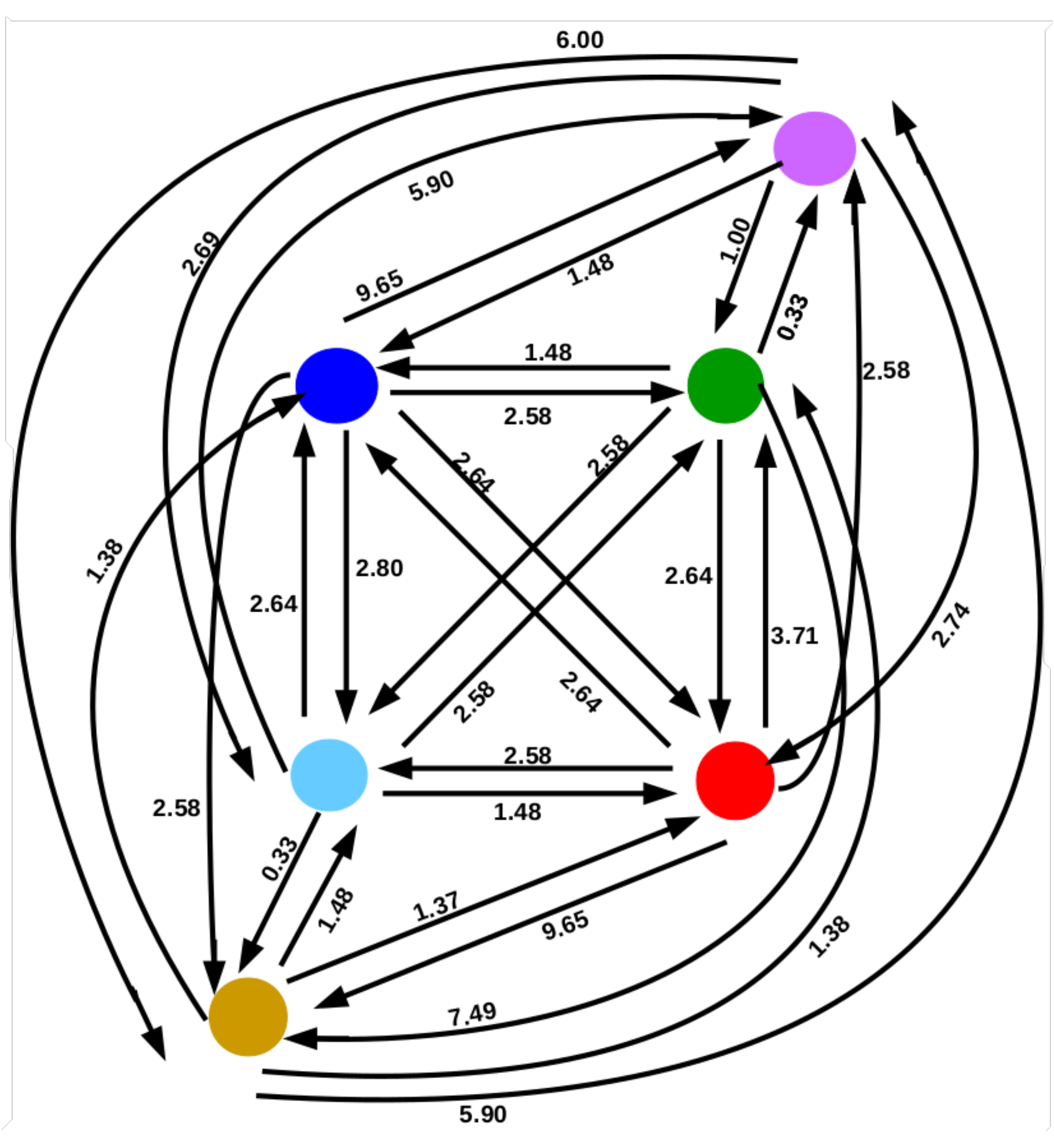}~\includegraphics[width=0.4\textwidth]{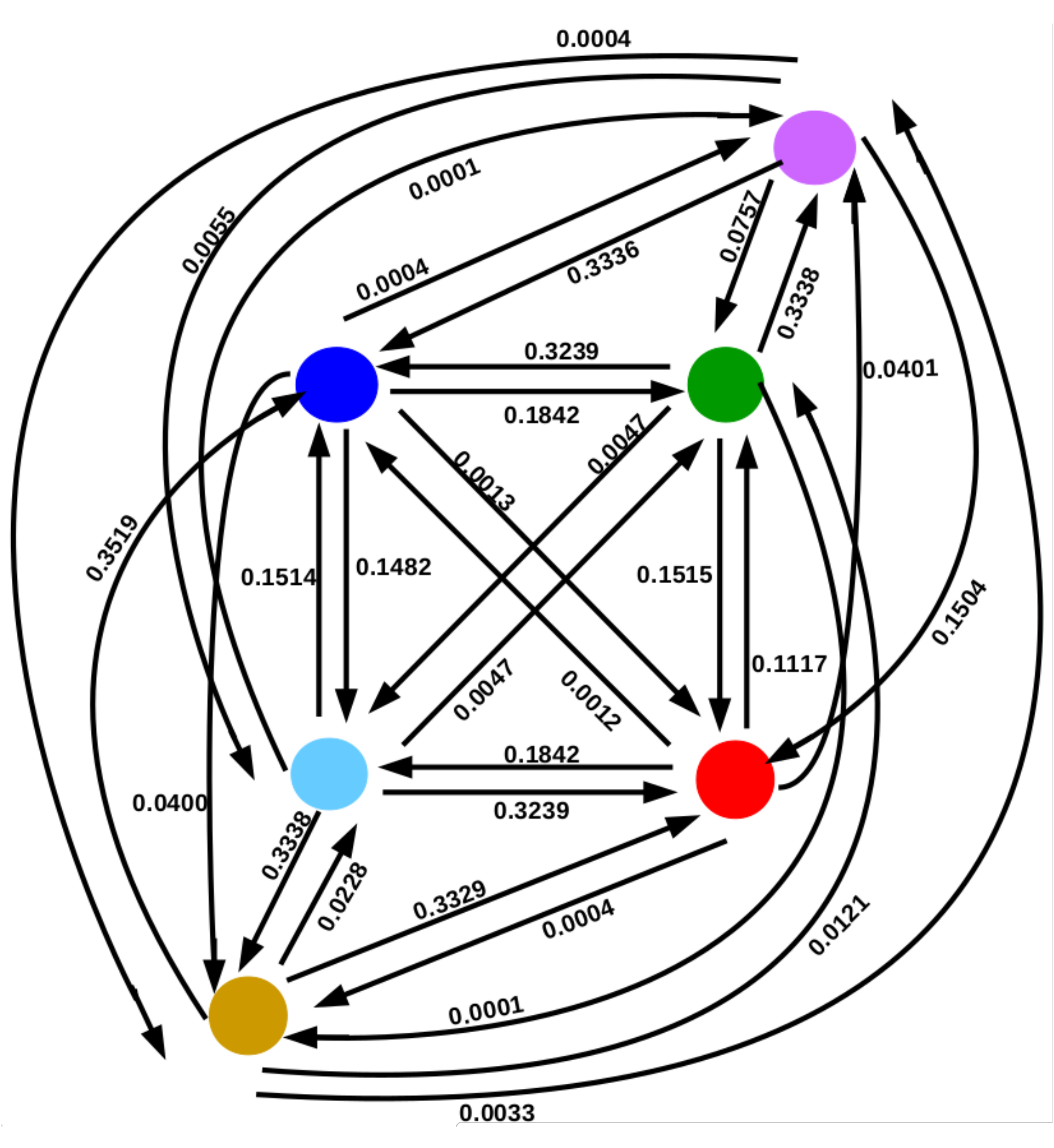}
\end{minipage}
\caption{\textbf{Top:} output of the trained ESN showing an incorrect switch around step $ 13200 $. \textbf{Center left:} fixed points found by the optimisation algorithm plotted in their two-dimensional subspace together with the trajectory. \textbf{Center right:} attractor plane divided by basins of attraction of every stable fixed point. The figure is drawn assuming a transient of $300$ time-steps. Every basin is coloured depending on the attractor where the ESN converges to without applying any input. \textbf{Bottom left:} extracted ENA weighted with estimation of input-driven excitability thresholds \eqref{eq:estimate_excitab}. \textbf{Bottom right:} extracted ENA weighted with effective excitability value \eqref{eq:beta}. Nodes are coloured according to the colours used in the center-right figure.}
\label{fig:6basins} 
\end{figure}

The top panel in Fig. \ref{fig:6basins} shows the output produced by an ESN achieving low MSE of the order of $10^{-3}$.
The small errors observable around time-step $13200$ can be explained by looking at the ENA model depicted in the bottom panels of Fig. \ref{fig:6basins}, which is represented with excitability thresholds \eqref{eq:estimate_excitab}, left panel, and with effective excitability values \eqref{eq:beta}, right panel.
The directed graphs (whose topology is clearly identical) reveal the presence of two extra stable fixed points in the ESN phase space, which are generated during training.
Nodes of these graphs are coloured according to output values produced by the ESN.
The activation of the related excitable connections brings the ESN to operate in the proximity of such superfluous states, producing inaccurate output values and hence explaining the origin of such errors. Notably, the directed edge -- in the graph on the right-hand side -- connecting the cyan with the yellow node has a relatively high value of effective excitability.
This means that, in the LSS of the related attractor (cyan), whose output is $(-1.03, -1.07)$, it is relatively easy to transition to the basin of the other attractor (yellow).
This, in turn, produces an output value equal to $(-0.84, -0.40)$, which is significantly different from the target output, i.e., $(-1, -1)$. The prediction error, visible to the naked eye in the top panel around time-step $13200$, is indeed due to the activation of that excitable connection.

Both of these spurious attractors act as a sort of surrogates for the correct ones (i.e., those producing lower prediction errors). In fact, once the internal state switches to a spurious attractor, the ESN still behaves consistently.
More precisely, spurious attractors are associated with higher effective excitability values on connections ending up in the correct attractors -- see the bottom-right panel of Fig. \ref{fig:6basins}. The existence of these spurious attractors and the unravelling of their roles in the ESN computation could not easily be inferred by looking only at the MSE or plotting the outputs produced by the ESN. This demonstrates the importance of ENA models for describing the ESN behaviour.

%\clearpage
\subsection{Noise tolerance and effective excitability of ENAs}
\label{sec:noise}

Here, we aim to provide a further example of the importance of ENAs for characterising the computation of ESNs.
In particular, we ask the following question: given a set of ESNs trained on the same task and achieving the same performance during training, which one will be more robust to noise during test? Typical performance measures, such as MSE, cannot be used to answer such a question and provide useful insights.
We show that an ENA model of the computation, weighted with effective excitability values \eqref{eq:beta}, allows us to assess robustness to noise of ESNs.

As a perturbation, we consider the usual white Gaussian noise term $\boldsymbol{\epsilon}$ in the state-update \eqref{eq:activation_update_fb2}, corresponding to perturbations directly applied to all neuron pre-activation values. We stress that such perturbations are applied only during the test phase; training is implemented as described in the previous section.
By increasing the noise standard deviation, the ESN trajectory gets increasingly perturbed neglecting the possibility to reach the proximity of attractors, hence affecting the accuracy of the resulting output values.

We note that: (i) ESNs yielding ENAs with higher effective excitability values are less robust to noise perturbations than ESNs giving rise to ENAs with low effective excitability values (see \cite{ashwin2018sensitive}), and (ii) ESNs producing ENAs with balanced edge weights on desired outgoing connections are more robust to noise perturbations than ENAs with unbalanced outgoing weights.
Regarding (i), high excitability values imply the existence of connections with low excitability thresholds.
That is, the basin of the attractor corresponding to the end-point of such a connection is very close to the attractor associated with the origin of the connection, resulting in unnecessary switches that induce errors.
Concerning (ii), for a given attractor, unbalanced outgoing connection weights could be a symptom of unbalanced distribution of space between basins of attraction in the LSS or the existence of some basins significantly closer to the attractor than other basins.
In the latter case, a reasoning similar to (i) applies. On the other hand, if there is indeed an asymmetric distribution of volumes between basins then basins of attractors corresponding to connections possessing large volume ratios will fill most of the LSS, leaving little space for basins of those attractors related to connections with small volume ratios.
Therefore, if the ESN state is close to an attractor with unbalanced outgoing connection weights, but with similar excitability thresholds, then under noise perturbations it is more likely to switch towards an attractor reachable through a connection with high effective excitability even when such a connection should not be activated.
For these reasons, an ENA, where each node is characterised by balanced weights on outgoing connections and very low effective excitability values on undesired connections, provides a prototypical example of ESNs whose behaviour is robust to noise.

\begin{figure}[ht!]
\begin{minipage}{0.5\textwidth}
\centering
\includegraphics[width=.75\textwidth]{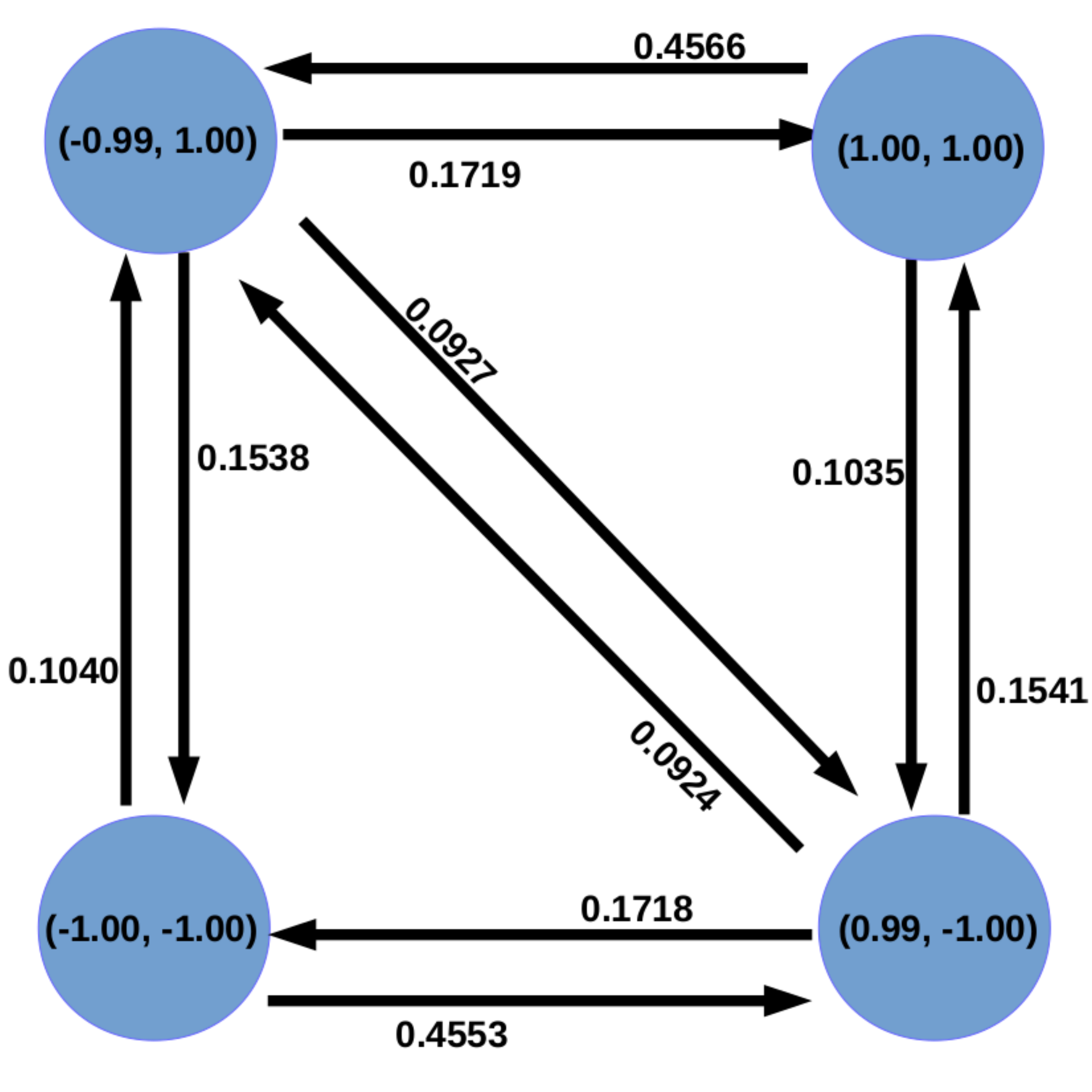}\\
\includegraphics[width=.95\textwidth]{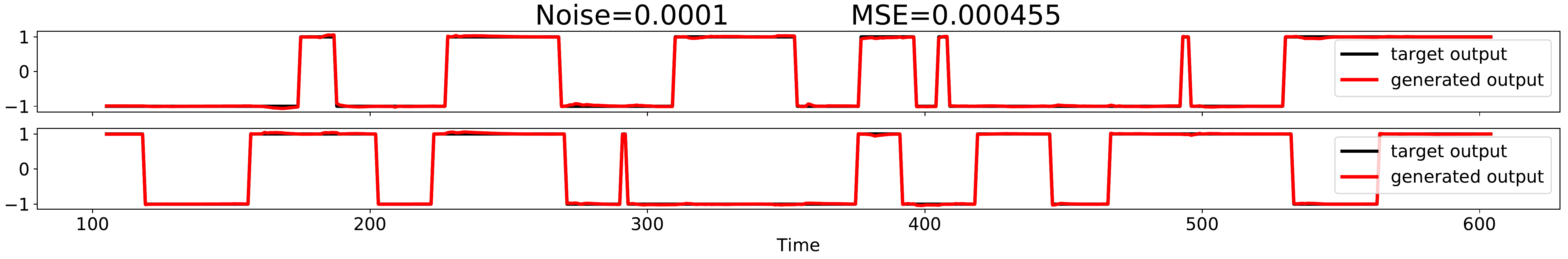}\\
\includegraphics[width=.95\textwidth]{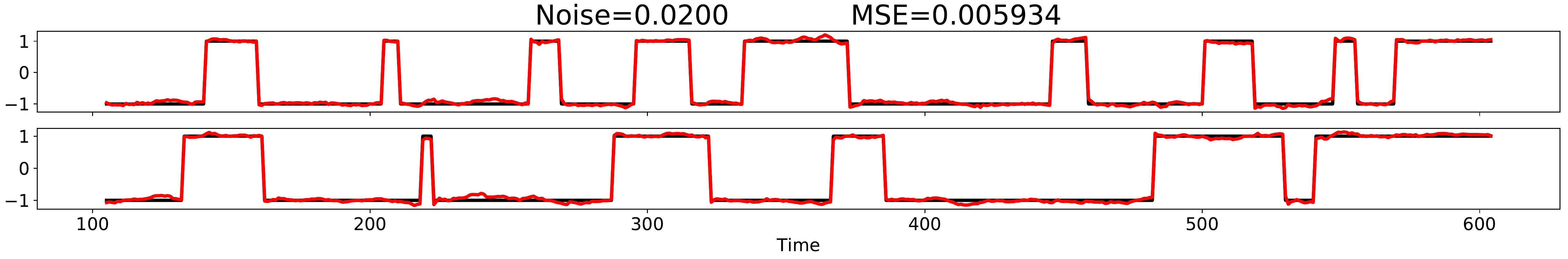}\\
\includegraphics[width=.95\textwidth]{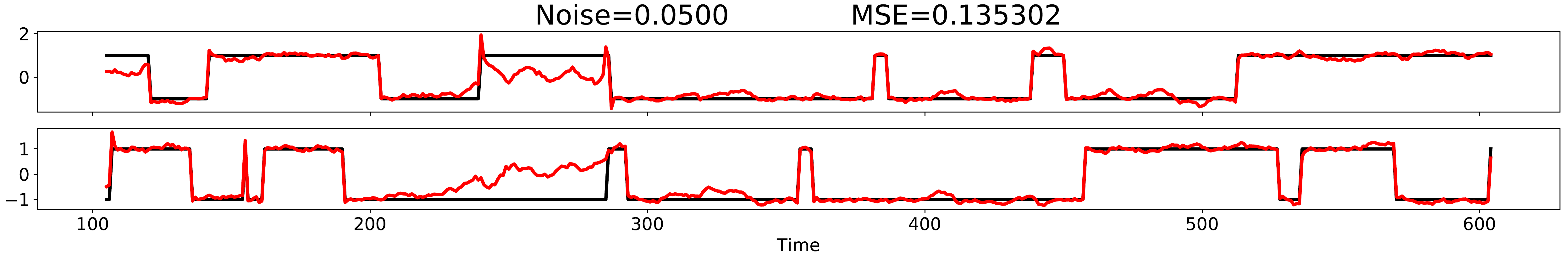}\\
\includegraphics[width=.95\textwidth]{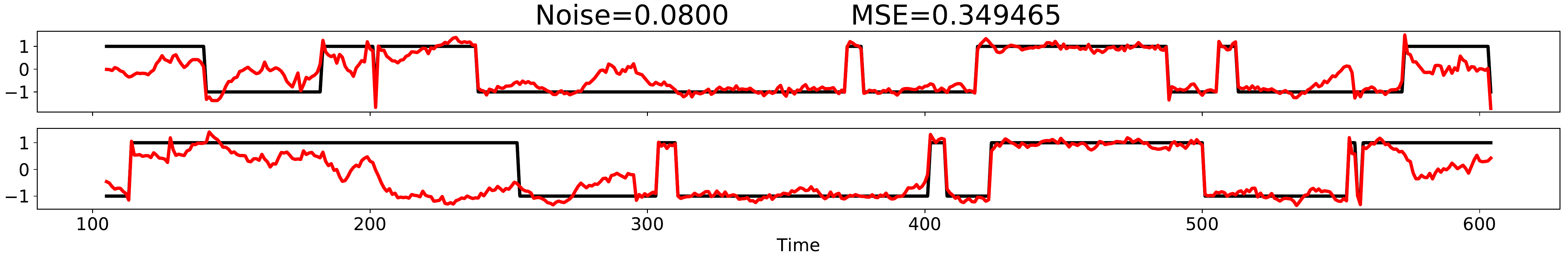}\\
\includegraphics[width=.95\textwidth]{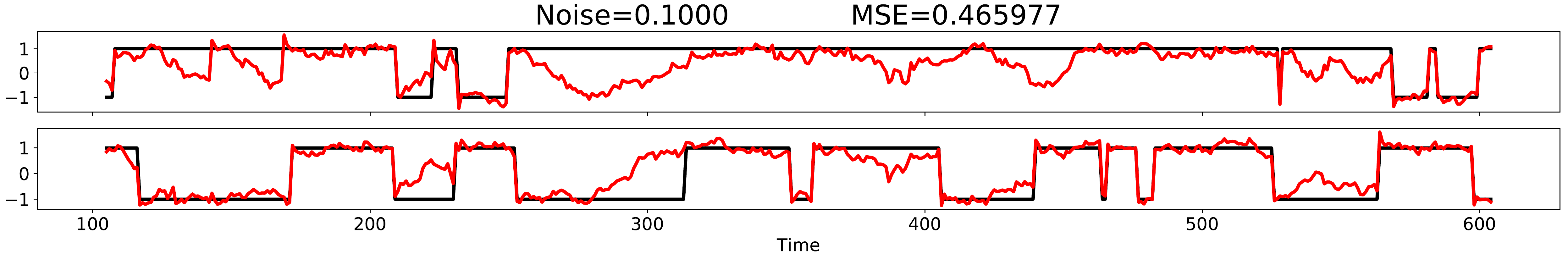}\\
\includegraphics[width=.95\textwidth]{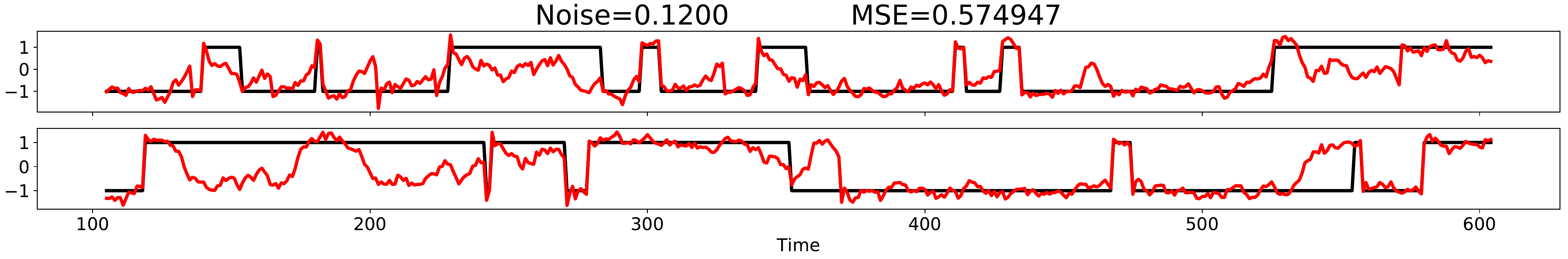}\\
\includegraphics[width=.95\textwidth]{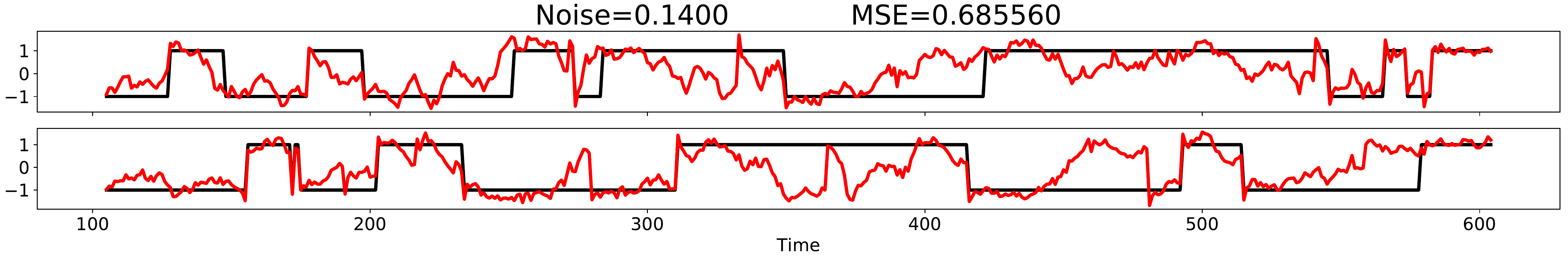}\\
\includegraphics[width=.95\textwidth]{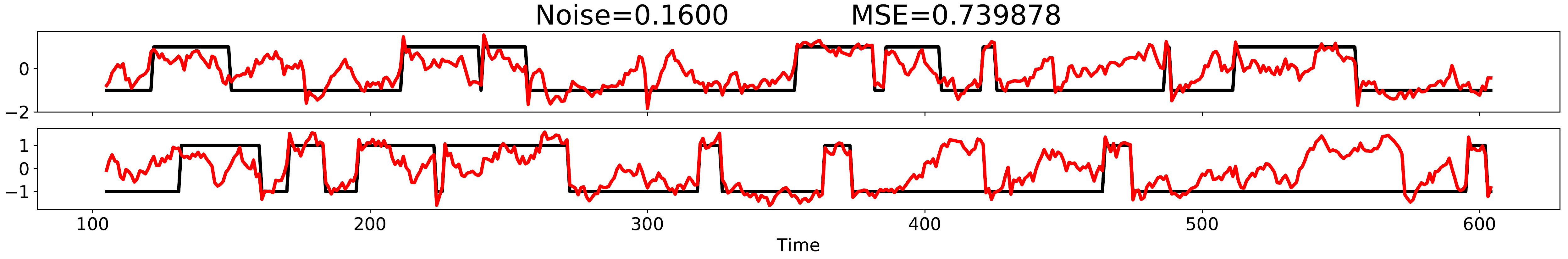}
\end{minipage}~\begin{minipage}{0.5\textwidth}
\centering
\includegraphics[width=.75\textwidth]{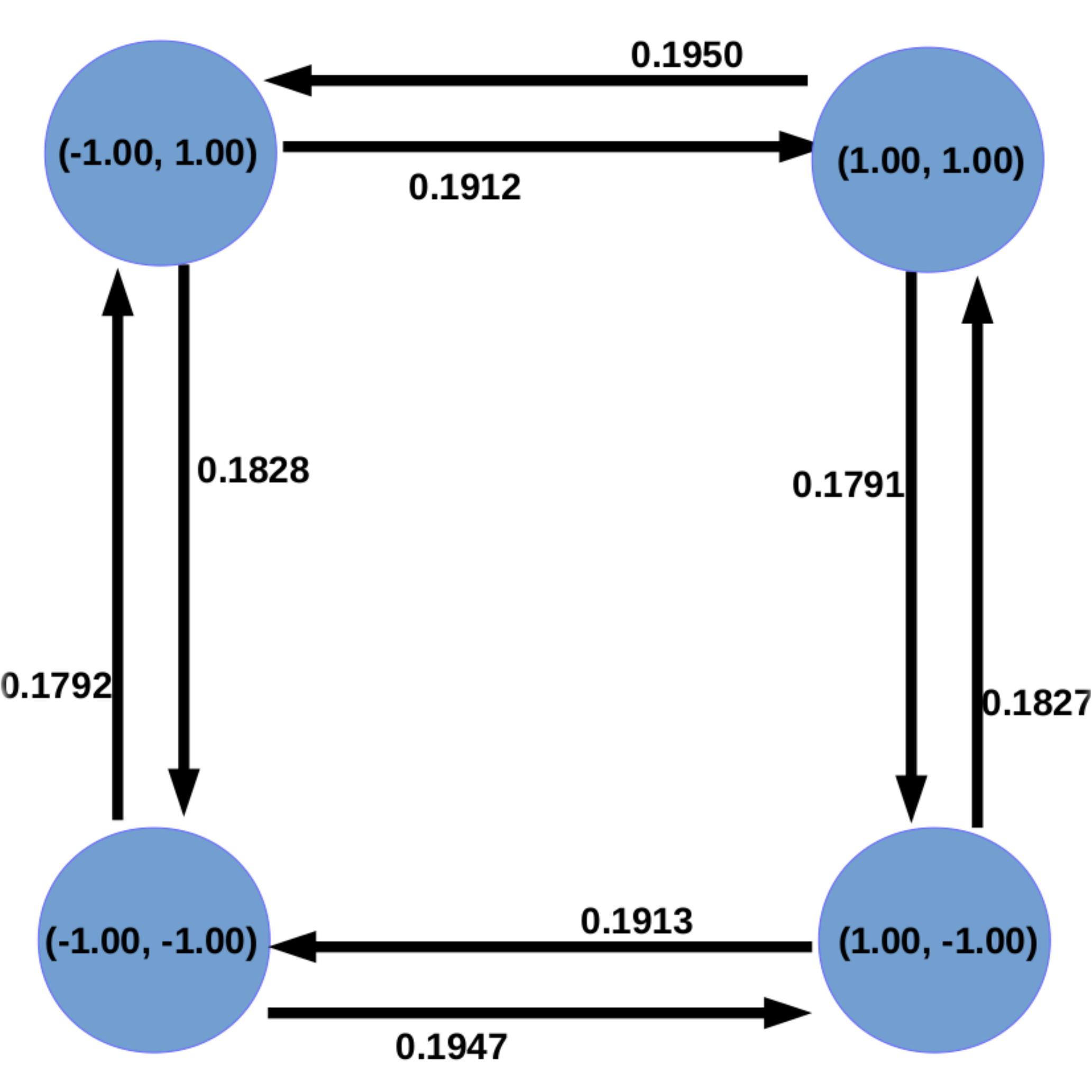}\\
\includegraphics[width=.95\textwidth]{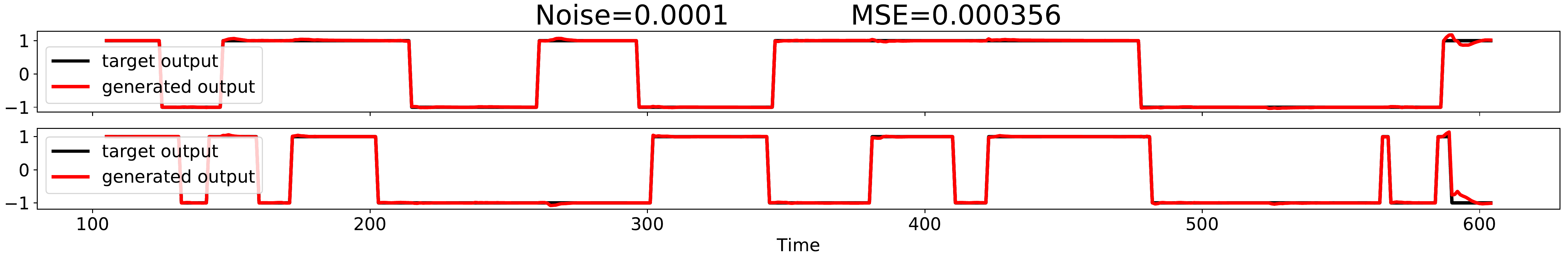}\\
\includegraphics[width=.95\textwidth]{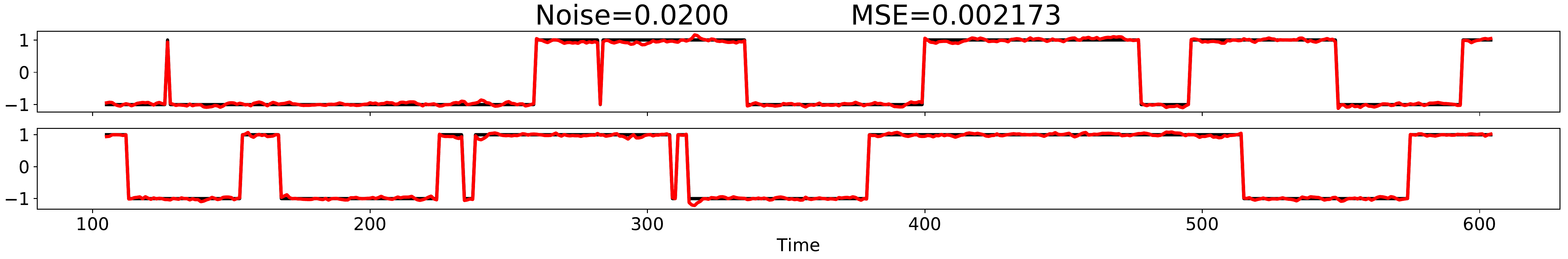}\\
\includegraphics[width=.95\textwidth]{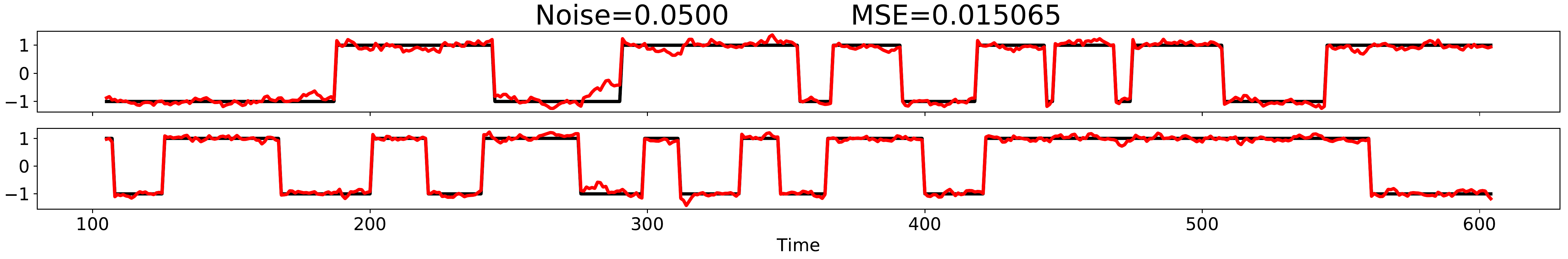}\\
\includegraphics[width=.95\textwidth]{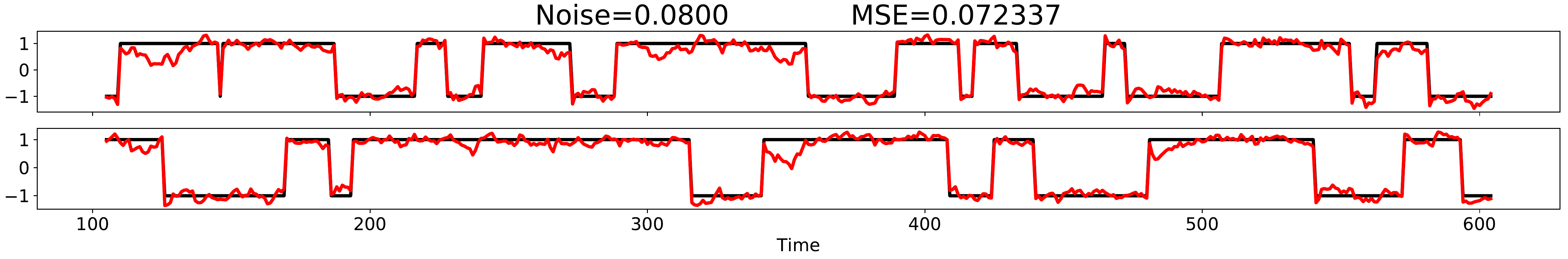}\\
\includegraphics[width=.95\textwidth]{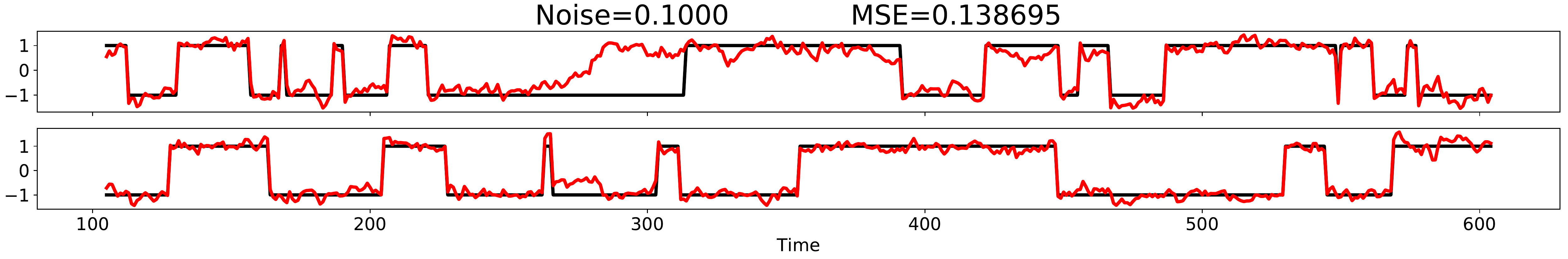}\\
\includegraphics[width=.95\textwidth]{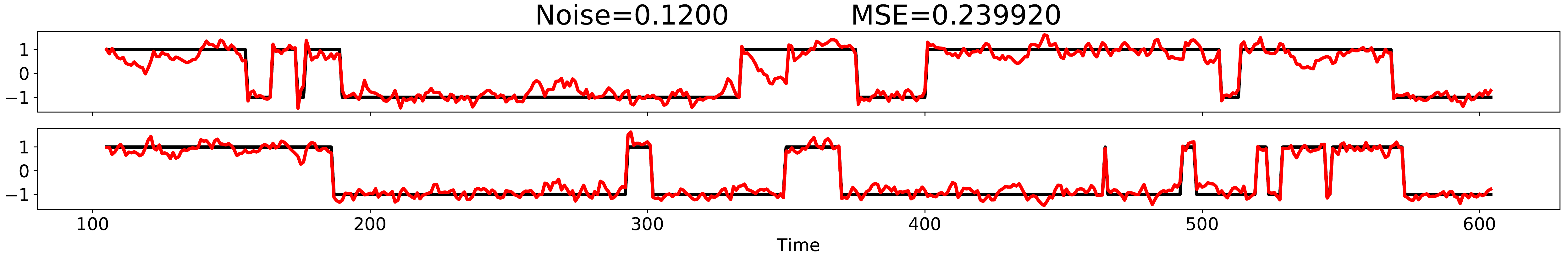}\\
\includegraphics[width=.95\textwidth]{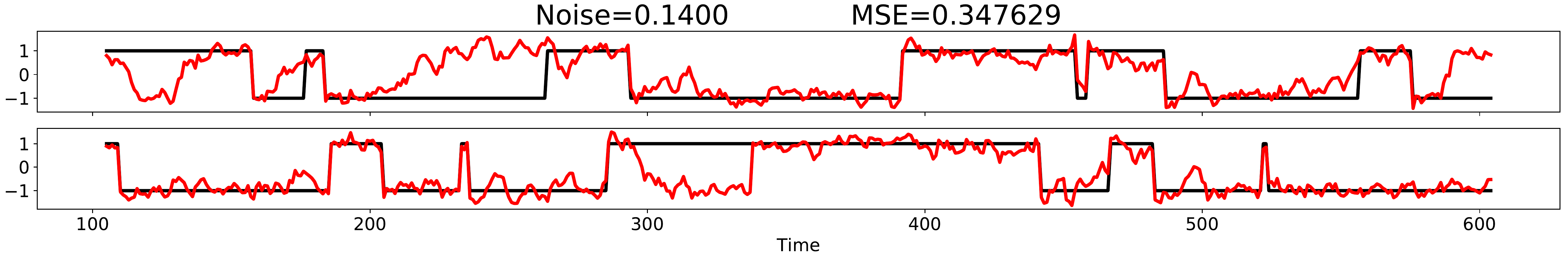}\\
\includegraphics[width=.95\textwidth]{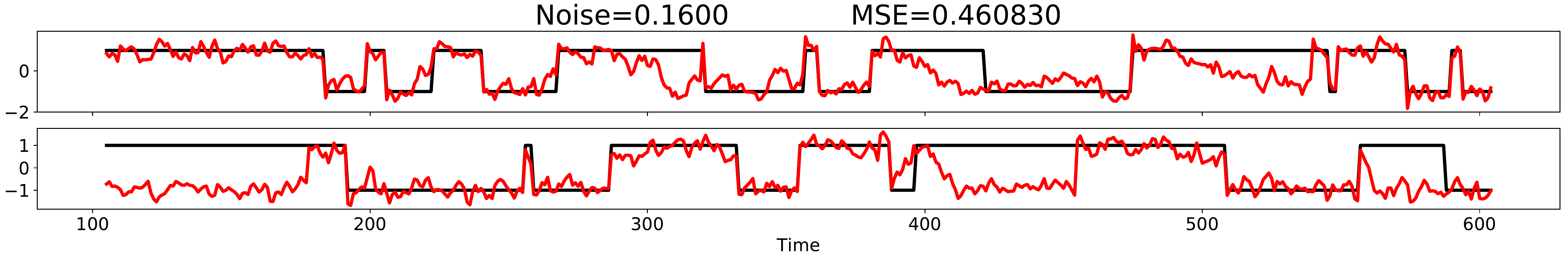}
\end{minipage}
\caption{\textbf{Left:} results corresponding to a trained ESN giving rise to an ENA with undesired connections and unbalanced outgoing weights \eqref{eq:beta}. \textbf{Right:} results obtained with a trained ESN giving rise to an ENA with balanced outgoing weights for all excitable connections. From top to bottom, in both columns, ESN outputs and related MSEs obtained with increasing noise standard deviation: $10^{-4}$, $2\cdot10^{-2}$, $5\cdot10^{-2}$, $8\cdot10^{-2}$, $10^{-1}$, $1.2\cdot10^{-1}$, $1.4\cdot10^{-1}$, and $1.6\cdot10^{-1}$.}
\label{fig:noise_robustness}
\end{figure}

To quantify the robustness of this behaviour to noise, we selected two ESNs that solve the two-bit flip-flop task with very high and comparable accuracy -- MSE during training is $\simeq10^{-4}$. We test them on the same input series composed of $10^5$ time-steps and recorded their MSEs by considering different noise instances with increasing standard deviation. Directed graphs representing the extracted ENAs are shown in Fig. \ref{fig:noise_robustness}.
Results for increasing noise standard deviation are divided in two columns: on the left column, we report results obtained by the ESN that is least tolerant to noise.
The directed graph on the left-hand side of Fig. \ref{fig:noise_robustness} shows the presence of two undesired connections. The edge weights of desired connections are not balanced, especially those outgoing from attractors with output values $(1,1)$ and $(-1,-1)$. Conversely, the directed graph on the right-hand side does not contain undesired connections and possess balanced outgoing weights.
The absence of undesired connections indicates that, in the LSS of every attractor, the basins of the attractors corresponding to undesired connections do not exist or, alternatively, they are very small and hence not detectable with the grid density used in our simulations; therefore, they are not relevant for describing the ESN behaviour according with the numerical precision we considered in our simulations.
Finally, we highlight that a performance breakdown for the ESN on the left-hand side panel is observed starting from noise standard deviation of $8\times10^{-2}$. On the other hand, the ESN on the right-hand side is significantly more robust to noise, denoting a performance breakdown for noise standard deviation of $1.4\times10^{-1}$.

%\clearpage
\section{Conclusions}
\label{sec:conclusions}

In this paper, we present a novel methodology for modeling and interpreting the behaviour of RNNs driven by inputs.
In order to obtain a mechanistic model describing how RNNs solve tasks, we exploit the theoretical framework offered by excitable network attractors \cite{ashwin2016designing}, which are defined as networks of stable fixed points connected by excitable connections.
We introduce a procedure to extract excitable network attractors directly from a trajectory generated by a trained RNN.
Such a procedure is composed of two main steps: first, fixed points are computed by solving a non-linear optimisation problem \cite{sussillo2013opening} and successively excitable connections, with related thresholds, are determined by simulating the dynamics of the autonomous system.

We validate our theoretical developments by considering ESNs trained on the flip-flop task, a simple yet relevant benchmark that consists of learning a prescribed number of stable states and related switching patterns guided by control inputs.
We cannot see any particular theoretical limitations in the application of our framework to more complex RNN architectures, although for more complicated dynamical tasks a detailed computation of bifurcation behaviour may become unfeasible.
Simulation results provide several interesting insights on how RNNs solve tasks and highlight the usefulness of excitable network attractors in describing how RNN undertake computations.
We train echo state networks by means of ridge regression. Our results (not shown here) suggest that the regularisation parameter has a direct impact on the number of attracting regions in phase space generated through training: using too low values produces under-regularised models with a large number of attractors.
An interesting future perspective consists of studying the impact of different training mechanisms (e.g., via FORCE learning or similar online approaches) on the resulting network attractor.

We believe that the proposed modelling framework based on excitable network attractors will be suitable to describe the RNN behaviour for many if not all tasks requiring the learning of a number of attractors (which need not be stable fixed points) and related switching patterns. To this end, in future we will also examine classification tasks. A related next step includes the possibility to handle inputs that are not instantaneous pulses, opening the way to more interesting case studies of practical relevance.
As mentioned in the introductory sections network attractors can in principle be constructed between any type of invariant sets, including limit cycles and strange attractors. Our focus on fixed points was mainly dictated by the fact that computing fixed points from a trajectory is significantly easier than computing limit cycles, for instance. 

Clearly, fixed points are not powerful enough to accurately model all possible behaviours in phase space.
Therefore, an interesting future perspective consists in extending our modeling framework to handle networks composed of heterogeneous attractors, such as a network in phase space connecting fixed points and limit cycles. In turn, this will allow modeling more complex RNN behaviour.
Finally, future directions include embedding the directed graph representing the excitable network attractor extracted from the trajectory in a new phase space, thus producing a set of ordinary differential equations \cite{ashwin2013designing} describing the RNN behaviour for the task under consideration.

%%%%%%%%
\clearpage
\appendices

\section{Linear stability analysis}
\label{sec:linear-stability}

By considering the case where the neural network is autonomous, we follow \eqref{eq:autonomous_system} and write the related map as $\mathbf{x}[k]=\mathbf{F}(\mathbf{x}[k-1])=\mathbf{G}(\mathbf{x}[k-1], \mathbf{0})$.
The $i$th component of the ESN state, $x_i, i=1,\ldots, N_r$, evolves according to the scalar map $F_i(\mathbf{x})=(1-\alpha)x_i + \alpha\tanh(\mathbf{M}_{(i)}\cdot\mathbf{x})$.
By noting that 
$$
\dfrac{\partial F_i}{\partial x_j}(\mathbf{x}_0)= (1-\alpha)\delta_{ij} + \alpha\bigl( 1- \tanh^2(\mathbf{M}_{(i)}\cdot\mathbf{x}_0) \bigl)m_{ij},
$$
where $ \delta_{ij} $ is the Kronecker delta, it is possible to write the Jacobian matrix evaluated onto $\mathbf{x}_0$ as $\mathbf{J}_{\mathbf{F}}(\mathbf{x}_0)= (1-\alpha)\mathbf{I}_{N_r} + \alpha\mathbf{D}(\mathbf{x}_0) \mathbf{M}$, where
%$\mathbf{D}(\mathbf{x}_0)=\mathrm{diag}(1 - \tanh^2 ( \mathbf{M}_{(1)}\cdot\mathbf{x}_0), \ldots, 1 - \tanh^2 ( \mathbf{M}_{(N_r)}\cdot\mathbf{x}_0))$ 
\begin{equation}
\label{eq:diagonal_squash}
\mathbf{D}(\mathbf{x}_0)=
\left[\begin{array}{cccc}
1 - \tanh^2 ( \mathbf{M}_{(1)}\cdot\mathbf{x}_0 ) & 0          & \ldots                 	                                    & 0 \\
0                                                 & 1 - \tanh^2 ( \mathbf{M}_{(2)}\cdot\mathbf{x}_0 )                                                 & \ldots                	                                    & 0 \\
\vdots                                            & \vdots     & \ddots  
	                                    & \vdots \\
0                                                 & 0          & \ldots 	                                    	                                    & 1 - \tanh^2 ( \mathbf{M}_{(N_r)}\cdot\mathbf{x}_0 ) \\
\end{array}\right]
\end{equation}
is an $N_r \times N_r $ diagonal matrix representing the squashing action of $\tanh(\cdot)$ along the saturating components of $\mathbf{x}_0$.
By linearizing the network state-update \eqref{eq:activation_update_fb2} around a given fixed point $\mathbf{x}^*$, we obtain the linear system $\delta\mathbf{x}[k+1] = \mathbf{J}_{\mathbf{F}}(\mathbf{x}^*) \delta\mathbf{x}[k]$, where $\delta\mathbf{x}[k] = \mathbf{x}[k] - \mathbf{x}^*$.
It is known \cite{strogatz2014nonlinear} that if a fixed point $\mathbf{x}^*$ is hyperbolic (i.e., $\mathbf{J}_{\mathbf{F}}(\mathbf{x}^*)$ has no eigenvalues on the unit circle in the complex plane), then the linear approximation provides a bona-fide characterisation of the non-linear behaviour around that fixed point.
Therefore, the (linear) stability of $\mathbf{x}^*$ is completely determined by the spectral radius of $\mathbf{J}_{\mathbf{F}}(\mathbf{x}^*)$. If all eigenvalues of $\mathbf{J}_{\mathbf{F}}(\mathbf{x}^*)$ are inside the unit circle, then $\mathbf{x}^*$ is a stable fixed point; on the other hand, if even just one eigenvalues has norm larger than one, then the linearized map is expanding along the corresponding eigenvectors and $\mathbf{x}^*$ is called a saddle\footnote{If every eigenvalue has norm greater than one, then the fixed point is a repeller.}.
We conclude observing that holds $ x_i^* =  \tanh ( \mathbf{M}_{(i)}\cdot\mathbf{x}^* ) \Longleftrightarrow \alpha x_i^*   = \alpha \tanh ( \mathbf{M}_{(i)}\cdot\mathbf{x}^* )  \Longleftrightarrow      x_i^* + \alpha x_i^* =  x_i^* + \alpha \tanh ( \mathbf{M}_{(i)}\cdot\mathbf{x}^* )	 \Longleftrightarrow    x_i^*= (1-\alpha)x_i^* + \alpha \tanh ( \mathbf{M}_{(i)}\cdot\mathbf{x}^* )$, hence the number of fixed points and their positions in phase space do not change on varying $\alpha \in (0,1]$.
However, their linear stability properties are directly affected by $\alpha$.

\section{Bifurcation of fixed points in ESNs}
\label{sec:bifurcations}

In what follows, we perform a bifurcation analysis of one-dimensional ESN map and provide some sufficient conditions to design two-dimensional ESNs with a desired number of fixed points. Particularly, we focus on \emph{fold bifurcations}\footnote{We refer to \citet{kuznetsov2013elements} for the terminology; the fold bifurcation is also known as \emph{saddle-node}, \emph{limit point} or \emph{turning point} bifurcation.} of low-dimensional ESN maps, which constitute the only codimension-1 bifurcation that generate new fixed points; as discussed by \citet{tivno2001attractive}, it is the usual mechanism for creating new attractive fixed point in RNNs.
Moreover, as shown by \citet{beer2006parameter}, some fold bifurcations are responsible for reducing the dimensions where the actual dynamics takes place, dividing the RNN parameter space in different regions of effective dimensionality.

\subsection{ESN maps in one dimension}
Here, we consider the following one-dimensional map,
\begin{equation}
\label{eq:1D_ESN}
x[k] = \tanh(m x[k-1] + w),
\end{equation}
where $(m, w) \in \mathbb{R}^2$ are the bifurcation parameters; we simplify the state-update in \eqref{eq:activation_update_fb2} and set leak rate $\alpha=1$, $\boldsymbol{\epsilon}=0$, and remove explicit reference to inputs.
Nevertheless, studying the map \eqref{eq:1D_ESN} is still useful to obtain insights about high-dimensional input-driven ESN dynamics.
In fact, in the high-dimensional case \eqref{eq:activation_update_fb2} the activation function of the $j$th neuron is determined by 
\begin{equation}
x_j[k] = \tanh \Bigl(m_{jj}x_j[k-1] \,\,\, + \,\,\, \sum_{s\neq j} m_{js}x_s[k-1] + (\mathbf{W}_{in})_{(j)} \cdot \mathbf{u}[k] +  \varepsilon_j   \Bigl).
\end{equation}
Hence, the parameter $w$ in \eqref{eq:1D_ESN} can be interpreted as the weighted sum of all incoming neurons, plus input and noise terms.
Fixed points of the map $F(x) = \tanh(m x + w)$ are the solutions $x^*$ of the equation $Q(x^*) = 0$, where 
\begin{equation}
\label{eq:1D-velocity}
Q(x) := F(x) - x \,\, = \,\,  \tanh( m x + w) -x.
\end{equation}
It is not possible to find a closed-form expression for the fixed points.
However, it is possible to state that: (i) for every $(m, w) \in \mathbb{R}^2$, there exists at least one fixed point; (ii) there can be one, two, or three fixed points.
The proof of these statements follow straightforwardly from the fact that $\lim_{x\rightarrow\pm\infty}Q(x)=\mp \infty$ and because, if $m<1$ then $Q$ is monotonic. Otherwise, there exist two critical points, $x_{l} < x_{r}$, namely
\begin{equation}
\label{eq:fold_points}
    x_{l,r} = \dfrac{1}{m} \Biggl[  \pm \tanh^{-1}\Biggl(\sqrt{\dfrac{m-1}{m}}\Biggl) - w  \Biggl],
\end{equation}
such that the function $Q(x)$ folds.
Moreover, since $Q(x^*) = 0 \Longleftrightarrow  x^* =  \tanh( m x^* + w) \in (-1,1) $ for every $( m, w) \in \mathbb{R}^2$, all fixed points lie in the open interval $(-1,1)$.
Critical points in \eqref{eq:fold_points} are solutions to $Q'(x)=0$, or equivalently to $F'(x) = 1$.

\begin{figure}[ht!]
\begin{minipage}{0.33\textwidth}
\centering
\includegraphics[keepaspectratio=true,scale=0.2]{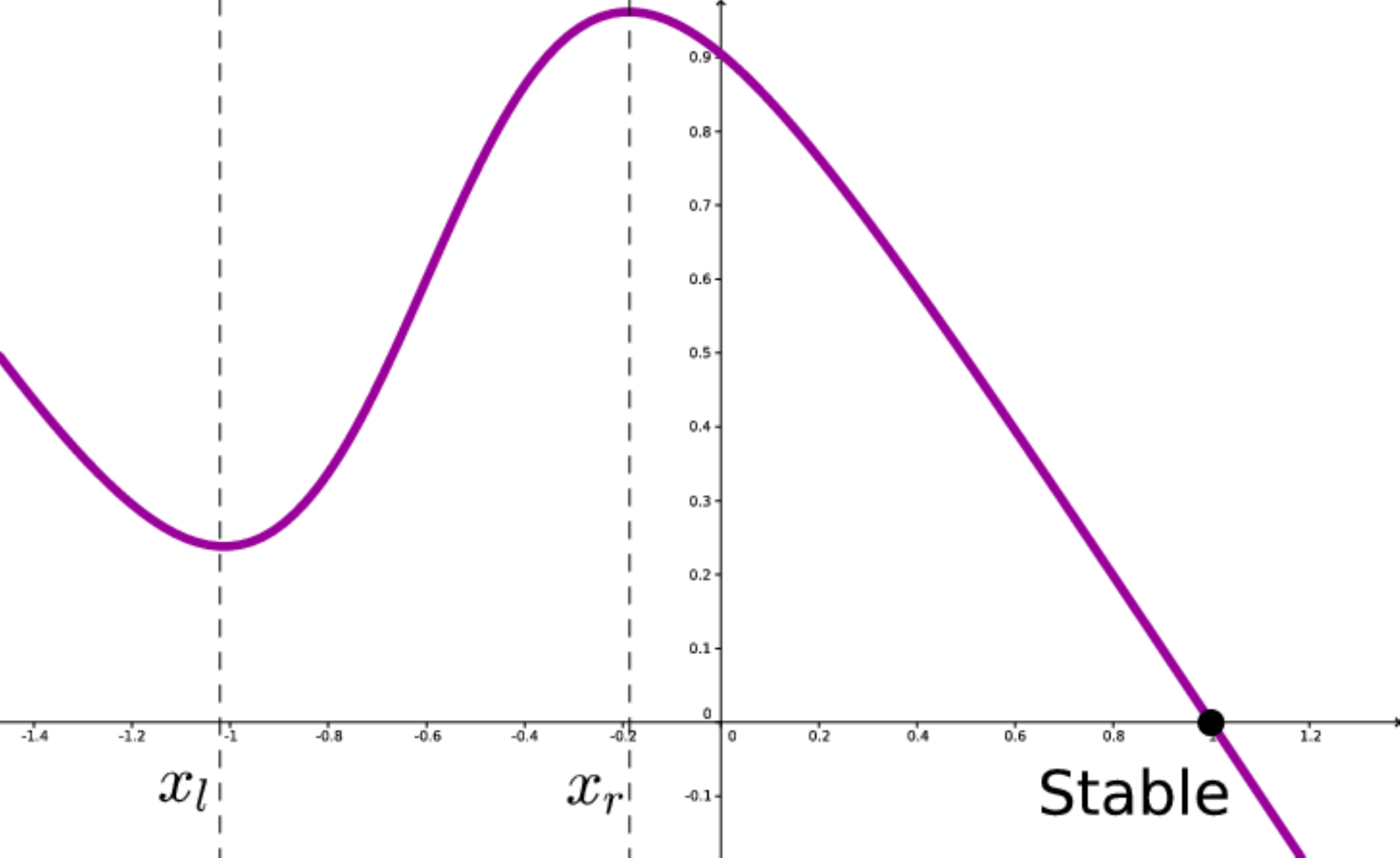}\\
\includegraphics[keepaspectratio=true,scale=0.2]{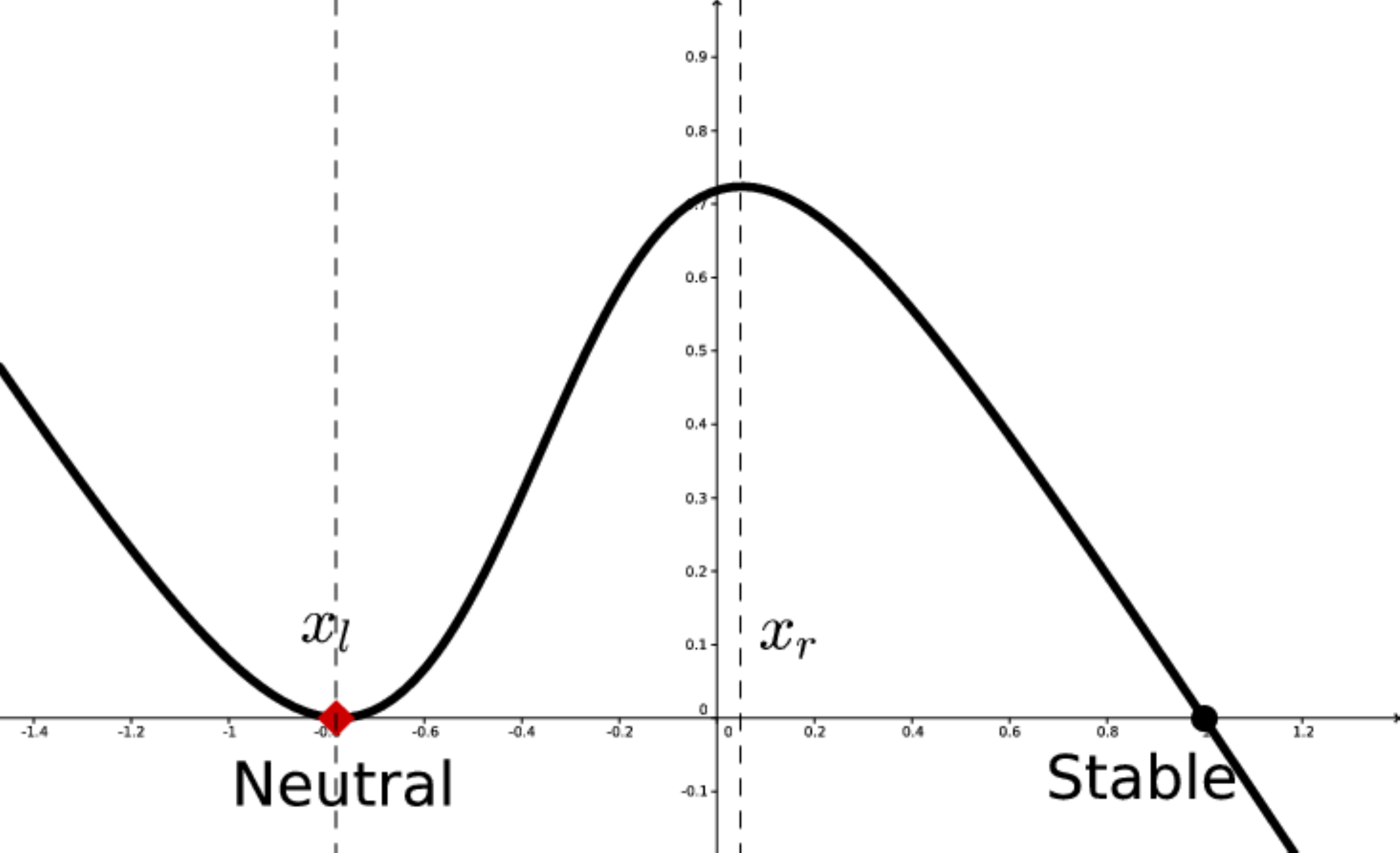}\\
\includegraphics[keepaspectratio=true,scale=0.2]{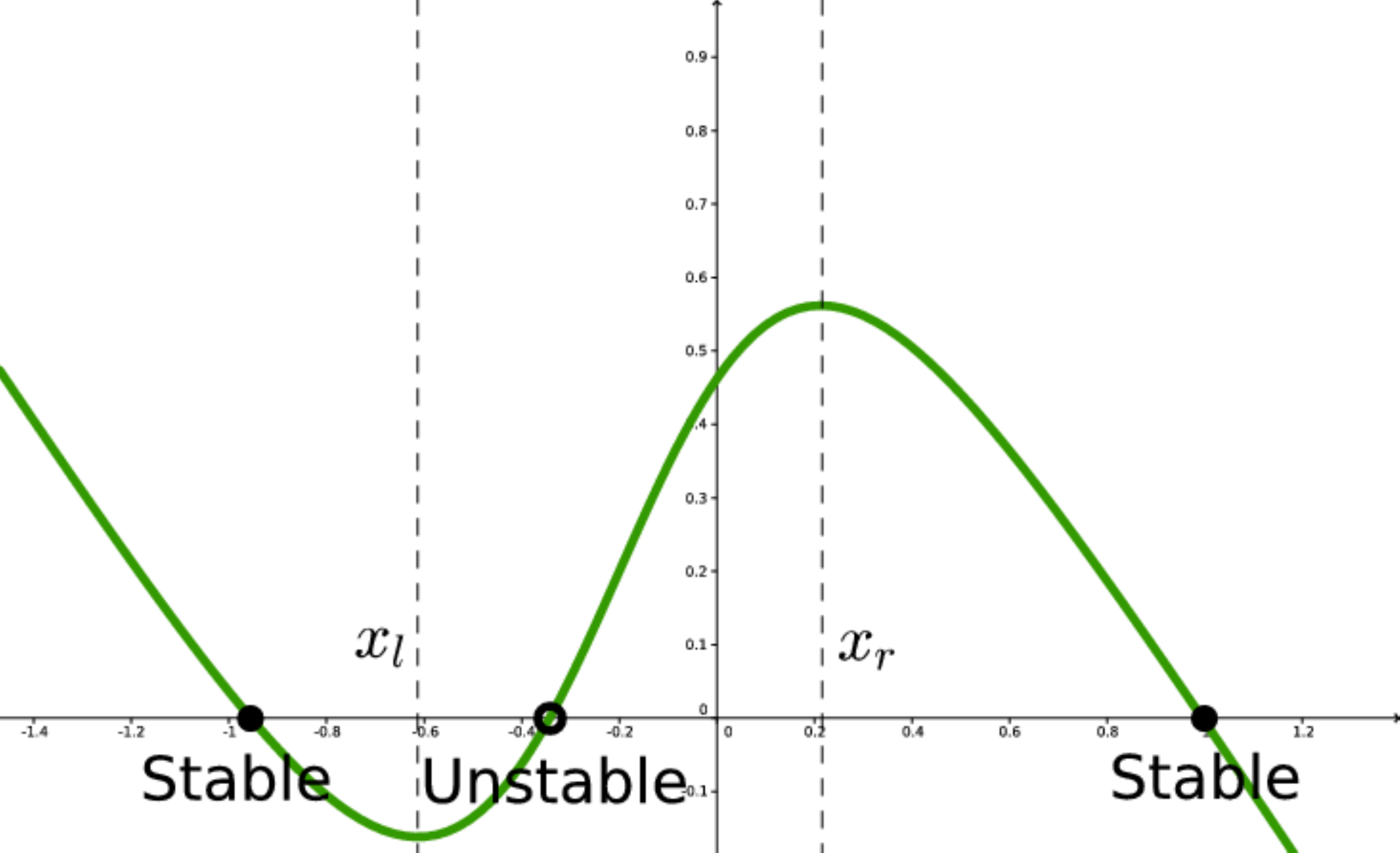}
\end{minipage}~\begin{minipage}{0.62\textwidth}
\centering
\includegraphics[keepaspectratio=true,scale=0.42]{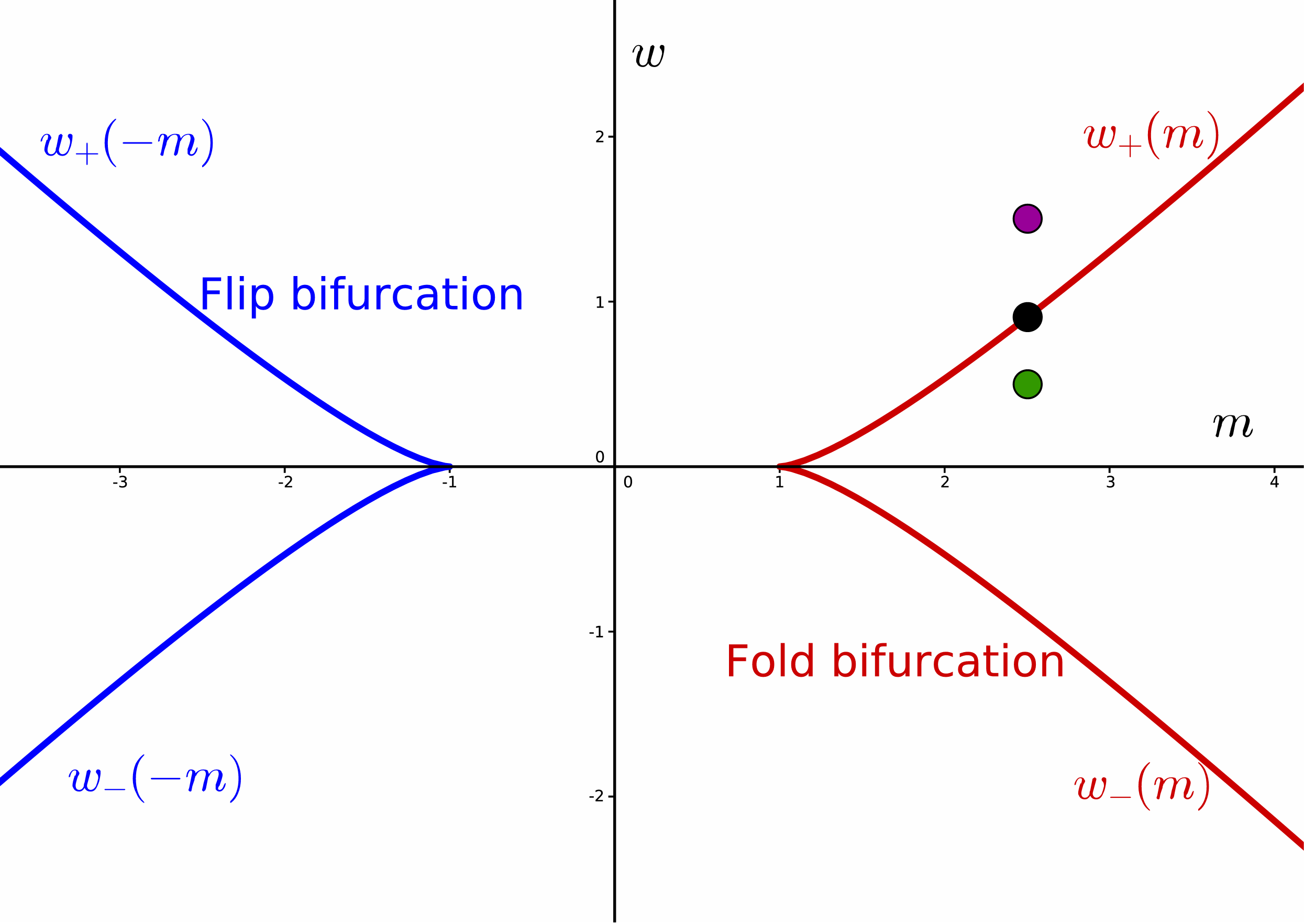}
\end{minipage}
\caption{ \textbf{Left:} three different folding configuration of the function $Q(x)$ in \eqref{eq:1D-velocity} with $m=2.5$; purple $w=1.5$, black $w\approx 0.9$, green $w=0.5$. \textbf{Right:} bifurcation diagram of fixed points in one-dimensional ESN.}
\label{fig:bif_1dim}
\end{figure}

Let us assume $m>0$. We observe a fold bifurcation whenever a critical point assumes the zero value. The condition $Q(x_{l,r}) = 0$ gives rise to the following parametrization of the fold bifurcation curve,
\begin{equation}
\label{eq:fold_curve}
    w_{\pm}(m) := \pm \Biggl[  m \sqrt{\dfrac{m-1}{m}}  \,\, - \,\,  \tanh^{-1}\Biggl( \sqrt{\dfrac{m-1}{m} }  \Biggl)   \Biggl], \qquad m \in [1, +\infty),
\end{equation}
which possesses two symmetric branches ending on a cusp; see Fig. \ref{fig:bif_1dim} for an illustration.
Crossing that curve in parameter space towards the region containing the semi-axis of $m>1$, a new fixed point is formed ($x_l$ or $x_r$, depending on the branch) and splits in a pair of fixed points, one stable and one unstable\footnote{The particular case of crossing that curve through the cusp gives rise to a pitchfork bifurcation of the origin.}.
This curve delimits the boundary between a dynamic regime where two stable points coexist with an unstable one, and a regime where only one fixed point exists.
Due to symmetry, in the $m<0$ case, there are two bifurcation branches identifying a \emph{flip} or \emph{period doubling bifurcation}, which is analytically described by the curve $w_{\pm}(-m)$ with $m\leq -1$. Crossing that curve towards the region containing the semi-axis of $m<-1$, a stable fixed point loses stability and gives rise to a 2-periodic attracting trajectory surrounding it.
We note that the flip bifurcation is detrimental for the flip-flop task considered here, as it gives period rather than fixed point attractors.

\subsection{ESN maps in two dimensions}
Here, we consider a two-neuron reservoir. We denote the activation functions of these neurons as $x$ and $y$, so that the ESN state evolution defines a trajectory $(x[k],y[k]) \in [-1, 1]^2, k=1, 2, ...,$ ruled by:
\begin{equation}
\label{eq:2D_ESN}
\begin{array}{l}
    x[k]=\tanh(a x[k-1] + b y[k-1] )  \, , \\
    y[k]=\tanh(c x[k-1] + d y[k-1] )  \, . \\
\end{array}
\end{equation}

It is know that a fixed point can undergo only fold or flip bifurcations in one-dimensional, discrete-time dynamical systems \cite{kuznetsov2013elements}. Nevertheless, considering two or more dimensions, also \emph{Neimark-Sacker} bifurcations could occur, where a stable point loses stability and an invariant curve surrounding it is created. These are the only possible codimension-1 bifurcations of a fixed point for a discrete-time dynamical system.
Among them, only the fold bifurcation can generate new fixed points.
The origin is always a fixed point of \eqref{eq:autonomous_system}. Considering \eqref{eq:2D_ESN}, the Jacobian matrix evaluated on the origin reads
\begin{equation*}
\mathbf{W}_{r}=
    \begin{pmatrix}
        a & b \\
        c & d
    \end{pmatrix}.
\end{equation*}
Therefore, training ESNs via \eqref{eq:trained_reservoir} implies a qualitative change of the dynamics around the origin if some eigenvalue crosses the unit circle, i.e., fixed point $(0,0)$ bifurcates.
However, adding a low-rank matrix to the reservoir can induce global bifurcations far away from the origin as well; hence, we cannot rely on the local bifurcation of the origin to deduce the global attractor structure after training.
As a counterexample, suppose $\lambda_1, \lambda_2 >1$. The diagonal matrix $\mathrm{diag}(\lambda_1, \lambda_2)$ and the upper triangular matrix $\begin{pmatrix} \lambda_1 & \gamma \\0 & \lambda_2 \end{pmatrix}$ share the same spectrum $\{\lambda_1, \lambda_2\}$.
Nevertheless, if the coupling is strong enough, that is, $|\gamma|>\dfrac{w_{+}(\lambda_1)}{ y^* }$, where $w_{+}(\lambda_1)$ is the function \eqref{eq:fold_curve} evaluated on $\lambda_1$ and $ y^*$ is the positive stable solution of $y[k]=\tanh(\lambda_2 y[k-1])$, then the fold bifurcation curve is crossed, making the dynamics for $x$ variable trivial. As a consequence of this, the upper triangular matrix induces dynamics with just two attractors (plus two saddles and the origin is a repeller), while the diagonal matrix induces four attractors (plus four saddles and the repeller).

\begin{figure}[ht!]
    \begin{minipage}{\textwidth}
    \centering
    \includegraphics[keepaspectratio=true,scale=0.3]{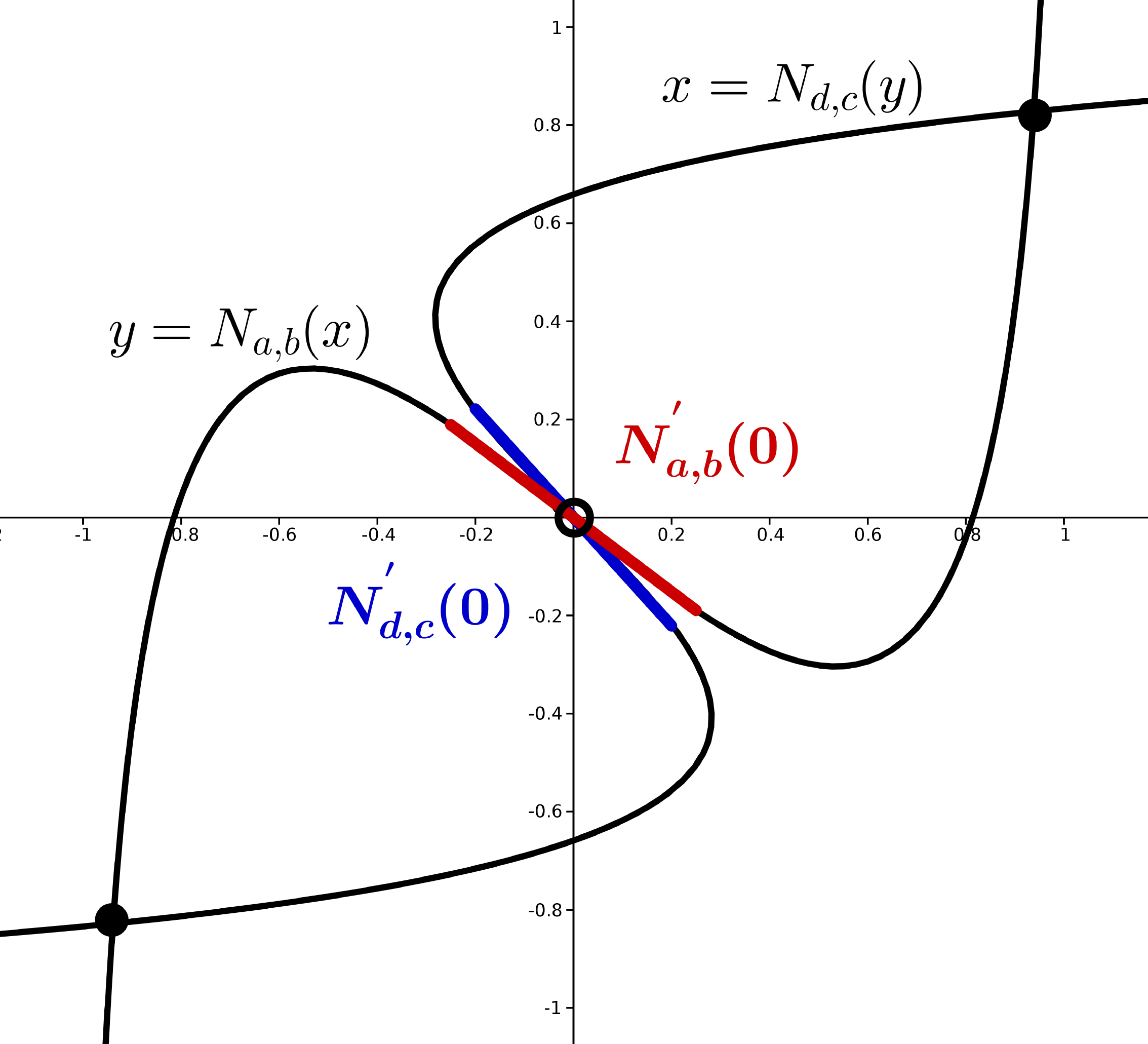}~\includegraphics[keepaspectratio=true,scale=0.3]{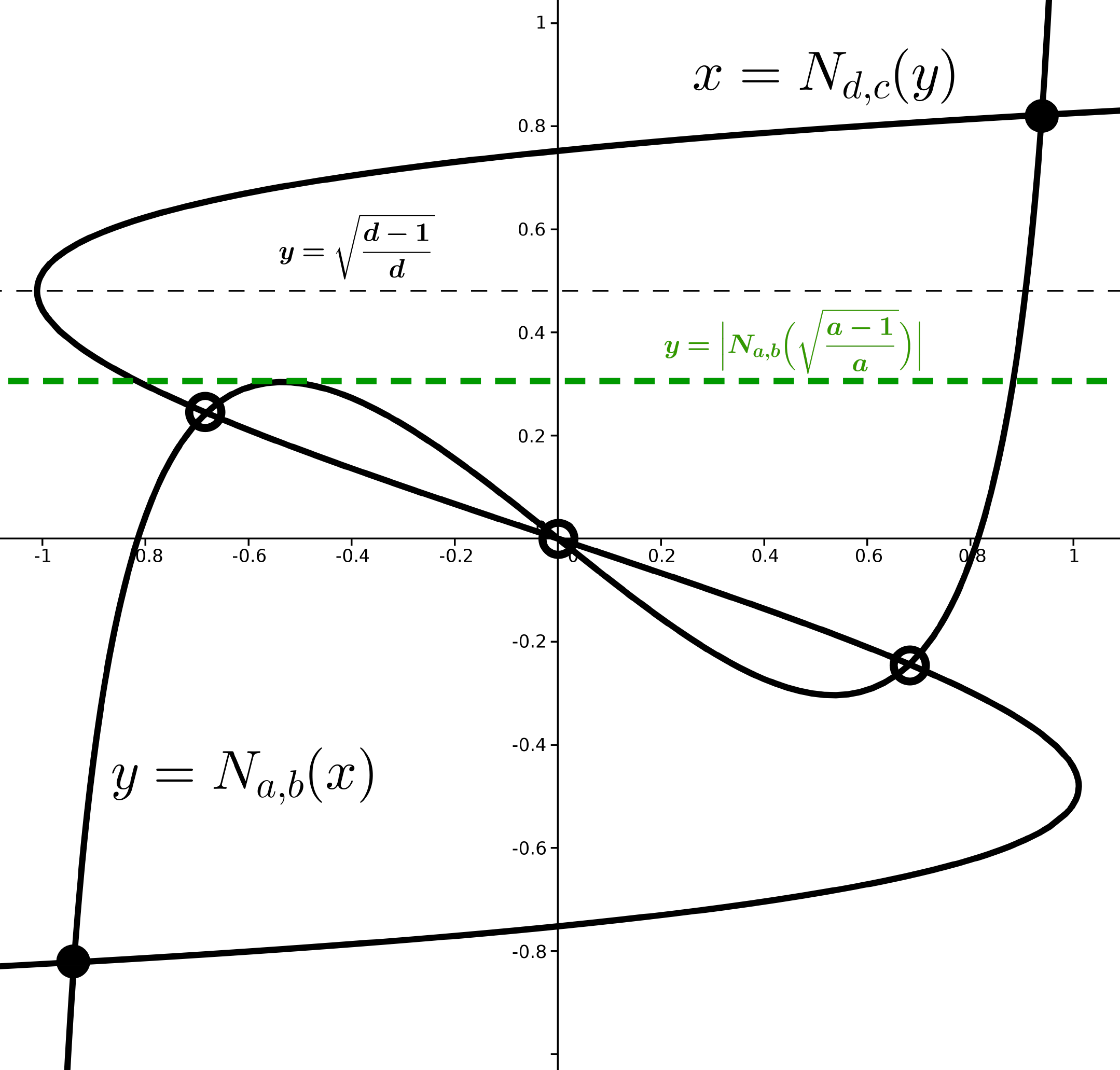}
    \end{minipage}
    \vspace{0.2cm}
    \begin{minipage}{\textwidth}
    \centering
    \includegraphics[keepaspectratio=true,scale=0.25]{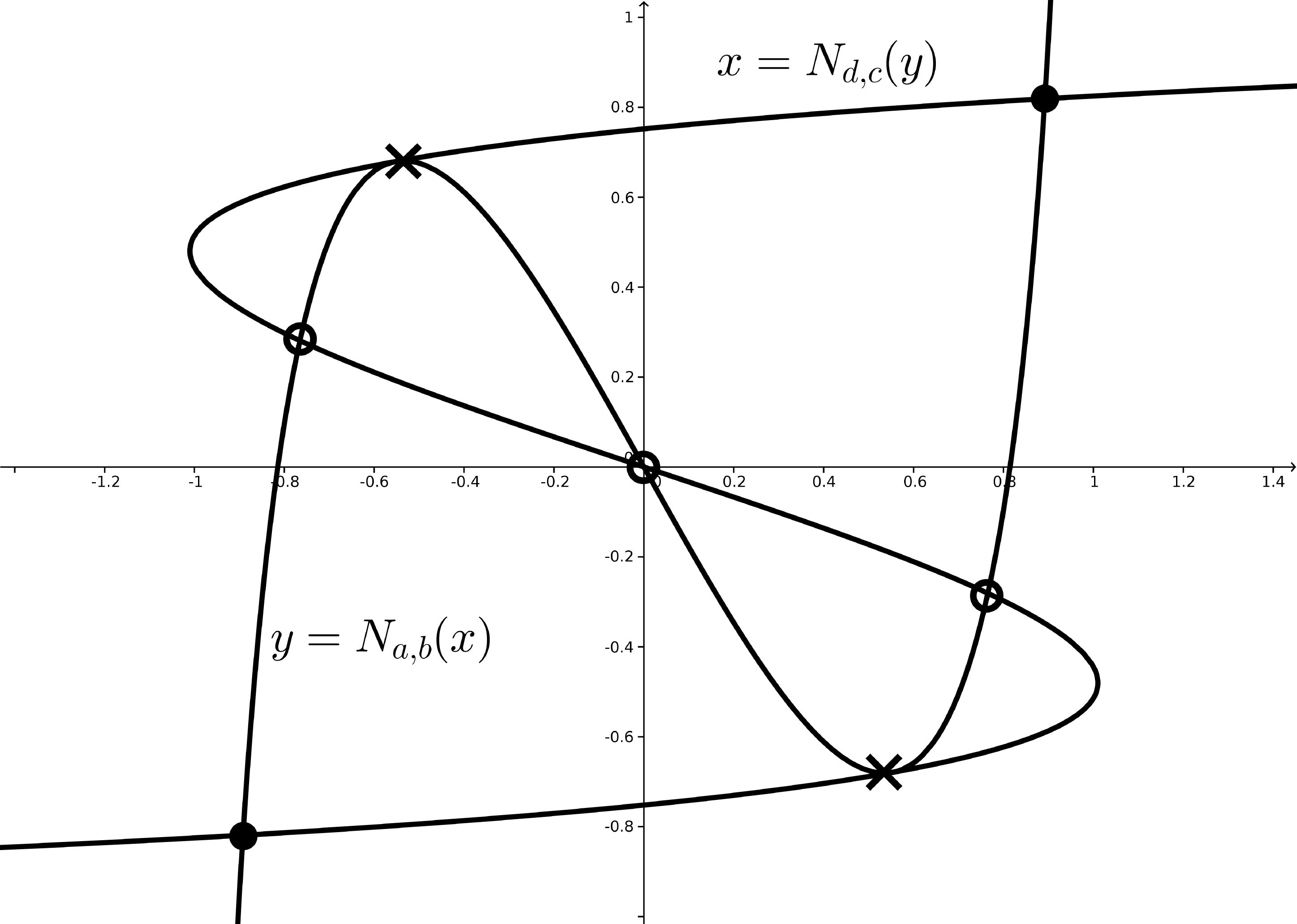}~\includegraphics[keepaspectratio=true,scale=0.25]{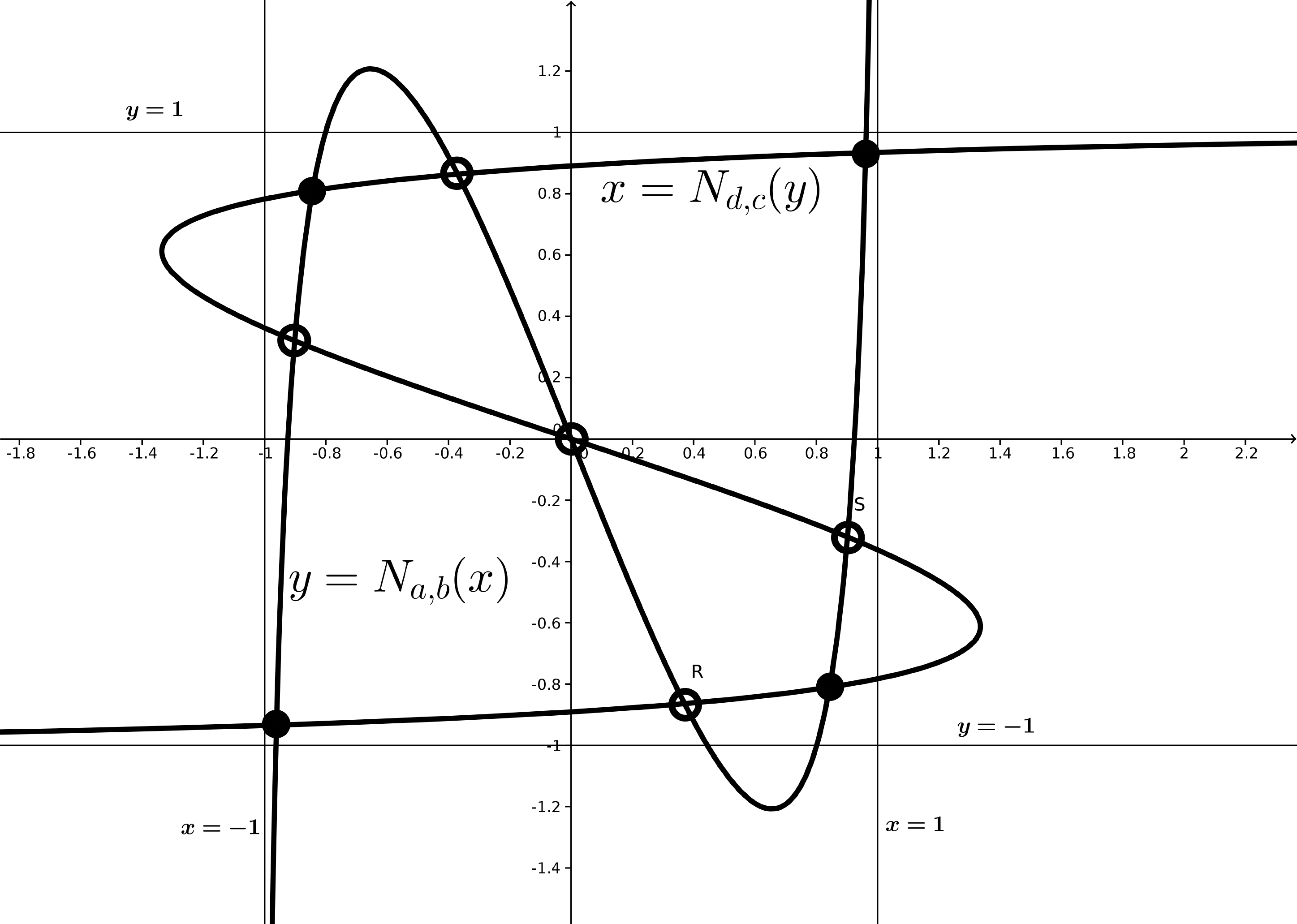}
    \end{minipage}
    \vspace{0.2cm}
    \begin{minipage}{\textwidth}
    \centering
    \includegraphics[keepaspectratio=true,scale=0.25]{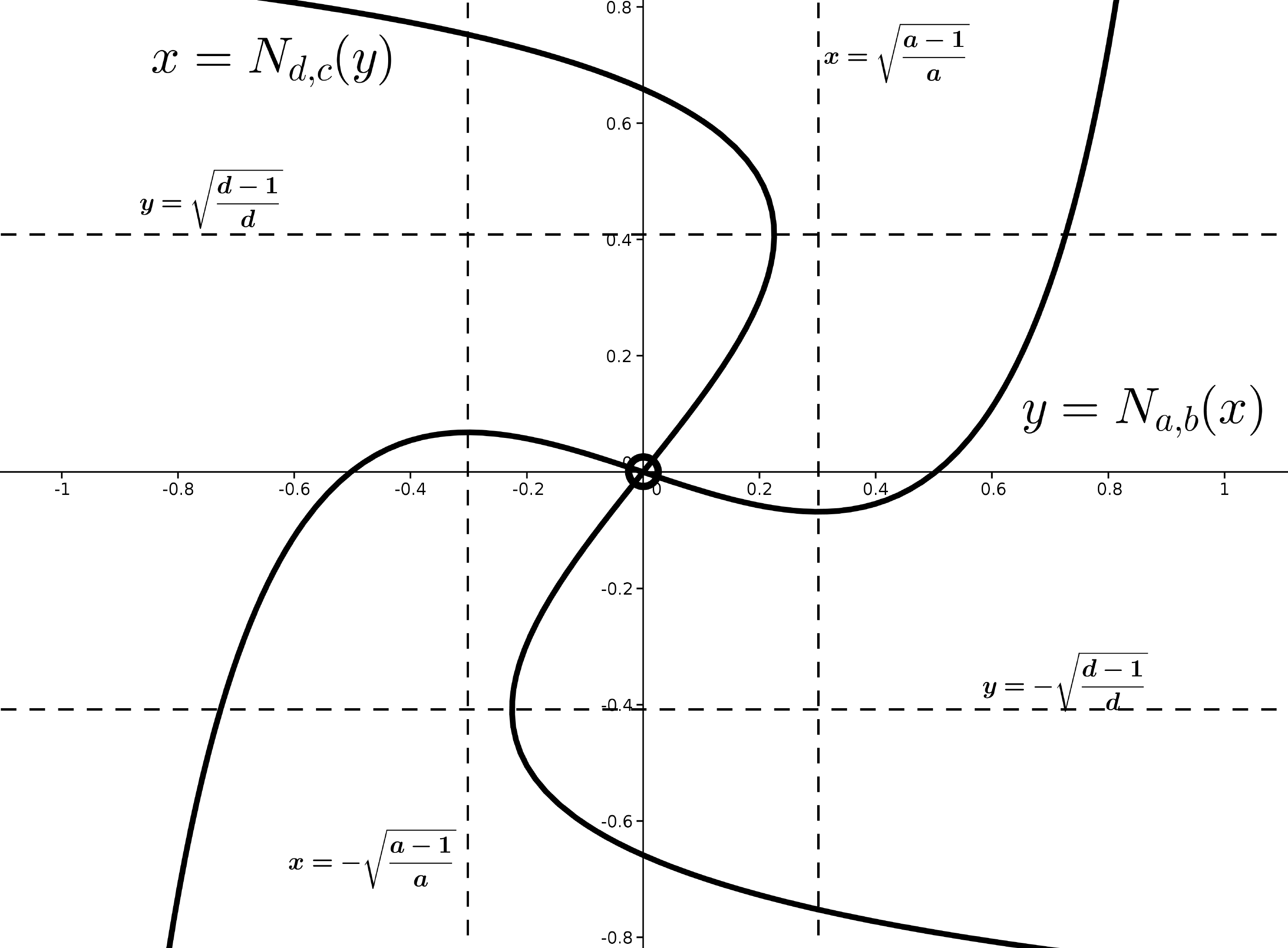}~\includegraphics[keepaspectratio=true,scale=0.25]{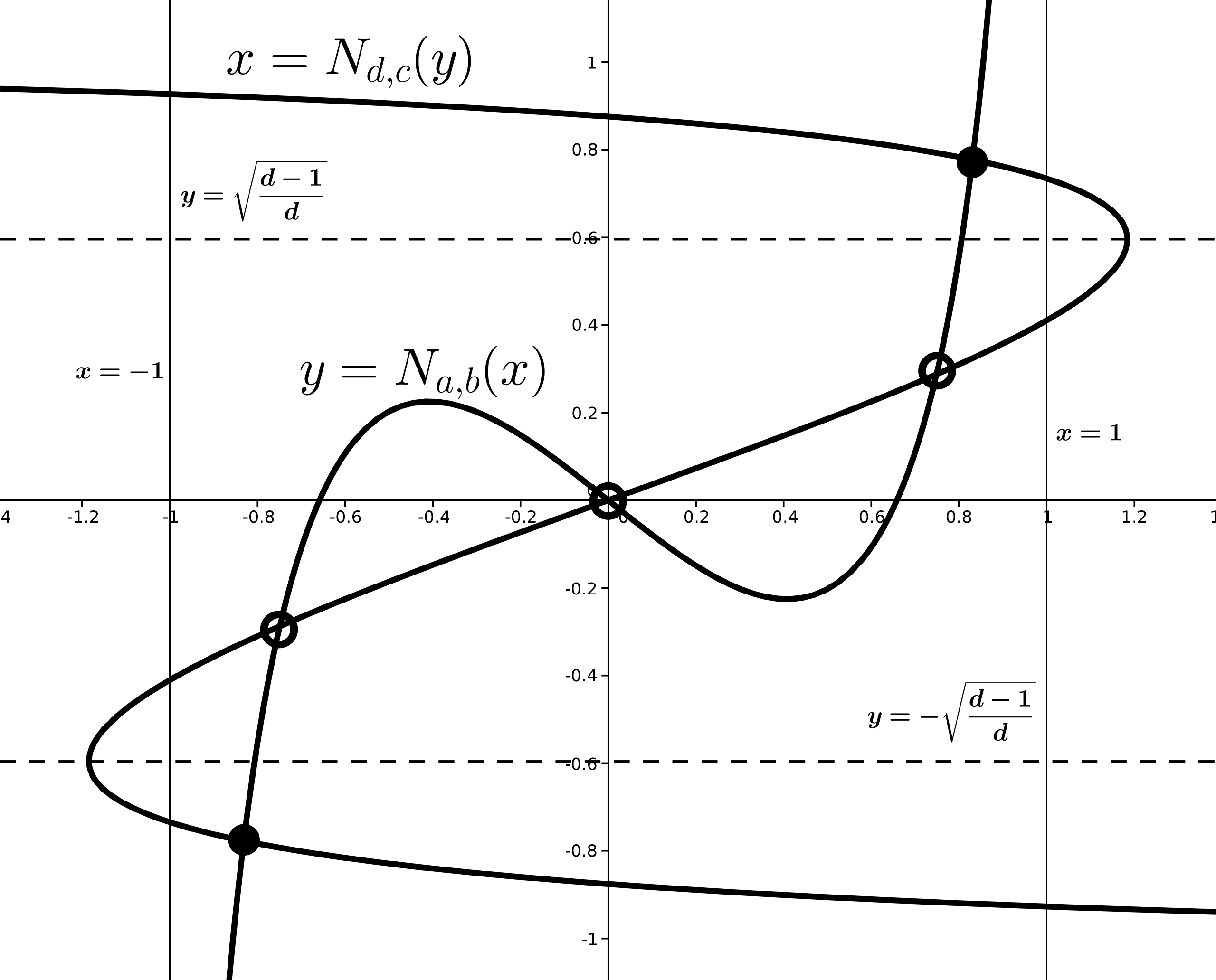}
    \end{minipage}
    \caption{In all panels, filled points denote stable attractors, while circles denote saddles or repellers. \textbf{Top left:} geometric representation of the sufficient condition \eqref{eq:origin_condit}; nullcline slopes on the origin are highlighted with different colours. \textbf{Top right:} illustration of the condition \eqref{eq:5fixpoint_condit}; the black dashed line acts as an upper bound ensuring that the maximum height of the peak of the $x$-nullcline (represented by the green dashed line) does not cross further the $y$-nullcline. \textbf{Centre left:} special case of 7 fixed points. Crosses indicate neutral fixed points where fold bifurcations take place. \textbf{Centre right:} an example of nullclines configuration holding the condition \eqref{eq:9fixpoint_condit} in the case of $bc > 0$. Both curves come out of the invariant square $[-1,1]^2$ with their humps, that guarantees 9 intersections, i.e., 9 fixed points. \textbf{Bottom left:} depiction of the sufficient condition \eqref{eq:1fixpoint_condit}; the origin is the only fixed point and it is unstable. \textbf{Bottom right:} illustration of the sufficient condition \eqref{eq:reverse5fixpoint_condit}.}
\label{fig:nullclines_fold} 
\end{figure}

In order to count the number of fixed points of the map \eqref{eq:2D_ESN}, we can draw its nullclines.
Defining function $N_{\alpha,\beta}(\eta) := \dfrac{1}{\beta}[-\alpha \eta +  \tanh^{-1}(\eta) ] $, the nullclines are given by
\begin{equation}
\label{eq:nullclines}
\begin{array}{l}
    y=N_{a,b}(x), \\
    x=N_{d,c}(y), \\
\end{array}
\end{equation}
and they represent the locus of points where $x$-dynamic / $y$-dynamic is stationary.
As a consequence, the solutions of the algebraic nonlinear system \eqref{eq:nullclines} coincide with the set of fixed points.
In the last part of this subsection, we show some sufficient conditions to control the number of fixed points assuming $a,d>1$, i.e., when both nullclines are folding. We refer to Fig. \ref{fig:nullclines_fold} for a graphical illustration.
\begin{itemize}
    \item Case $b c \geq 0$.\\
    \begin{equation}
        \label{eq:origin_condit}
        |N^{'}_{a,b}(0)| < |N^{'}_{d,c}(0)|^{-1} , \,\,\text{i.e. } (1-a)(1-d) < bc \quad  \Longrightarrow \quad  \text{there are exactly 3 fixed points.} 
    \end{equation}
    On the other hand, when $(1-a)(1-d) \geq bc$, there could exist 5, 7 or 9 fixed points.
    Critical points of the function $N_{\alpha,\beta}(\eta)$ are $\eta_{\pm} = \pm \sqrt{\frac{\alpha - 1}{\alpha}}$.
    Therefore, if $(1-a)(1-d) \geq bc$ holds then
    \begin{equation}
        \label{eq:5fixpoint_condit}
        \Biggl|N_{a,b}\Biggl( \sqrt{\frac{a - 1}{a}} \Biggl)\Biggl|< \sqrt{\dfrac{d - 1}{d}} \,\,\, \text{or} \,\,\, \Biggl|N_{d,c}\Biggl( \sqrt{\frac{d - 1}{d}} \Biggl)\Biggl|<\sqrt{\dfrac{a - 1}{a}} \quad \Longrightarrow \quad \text{there are exactly 5 fixed points.}
    \end{equation}
    Fig. \ref{fig:nullclines_fold} depicts a nullcline configuration where there are 5 intersections\footnote{Note that it holds $N_{\alpha,\beta}\Bigl(\pm\sqrt{\frac{\alpha - 1}{\alpha}}\Bigl)= -\frac{1}{\beta} w_{\pm}(\alpha) $.}.
    From that configuration, we can make the humps of $y=N_{a,b}(x)$ more pronounced by increasing the $\frac{a}{b}$ ratio, until a fold bifurcation occurs, giving rise to a new couple of pair of fixed points.
    The case of 7 fixed points is obtained after the fold bifurcation.
    
    \item Case $ b c < 0$. \\
    \begin{equation}
        \label{eq:1fixpoint_condit}
        \Biggl|N_{a,b}\Biggl( \sqrt{\frac{a - 1}{a}} \Biggl)\Biggl|< \sqrt{\dfrac{d - 1}{d}} \,\,\, \text{and} \,\,\, \Biggl|N_{d,c}\Biggl( \sqrt{\frac{d - 1}{d}} \Biggl)\Biggl|<\sqrt{\dfrac{a - 1}{a}} \quad \Longrightarrow \quad \text{there is exactly 1 fixed point.}
    \end{equation}
    From the above configuration, increasing the $d/c$ ratio makes stretch the humps of $N_{d,c}$ until a fold bifurcation occurs producing exactly three fixed points. Beyond the fold bifurcation, a new pair of fixed points is generated.
    \begin{equation}
        \label{eq:reverse5fixpoint_condit}
        \Biggl|N_{a,b}\Biggl( \sqrt{\frac{a - 1}{a}} \Biggl)\Biggl|< \sqrt{\frac{d - 1}{d}} \,\,\, \text{and} \,\,\, \Biggl|N_{d,c}\Biggl( \sqrt{\frac{d - 1}{d}} \Biggl)\Biggl| > 1 \quad \Longrightarrow \quad \text{there are exactly 5 fixed points.} \footnote{Due to symmetry, this condition holds also when inverting the role of $N_{d,c}$, $N_{a,b}$ and consequently $(a,b)$, $(d,c)$.}
    \end{equation}
\end{itemize}

In both cases, a simple sufficient condition to ensure the existence of 9 fixed points, see Fig. \ref{fig:nullclines_fold}, is 
\begin{equation}
    \label{eq:9fixpoint_condit}
    \Biggl|N_{a,b}\Biggl( \sqrt{\frac{a - 1}{a}} \Biggl)\Biggl|>1 \,\,\, \text{and} \,\,\, \Biggl|N_{d,c}\Biggl( \sqrt{\frac{d - 1}{d}} \Biggl)\Biggl|>1  \quad \Longrightarrow \quad \text{there are exactly 9 fixed points.}
\end{equation}
The maximum number of nullcline intersections is 9, setting the maximum number of fixed points that can be generated in a two-dimensional ESN map.

\clearpage
\section{Aggregation of fixed points via clustering}
\label{sec:fixed_point_aggregation}

The optimization procedure described in Sec. \ref{sec:optimization-fixed-points} provides a number of solutions equal to the number of initial conditions taken into account (assuming all initial conditions converge); however, such solutions might be similar up to a prescribed numerical precision.
In order to reduce the set of all solutions to a few effective fixed points, we run the $k$-means clustering algorithm and retain only the elements belonging the final clusters minimizing the distance w.r.t. the cluster centroids.
Instead of aggregating all solution at once, we first group such solutions according to their linear stability properties as indicated by the spectrum of the Jacobian matrix of the autonomous map \eqref{eq:autonomous_system}.
In particular, we form the group of linear stable fixed points, the group with one unstable direction and so on.
Finally, for each group, we run $k$-means with parameter $k$ identified according with the minimum of the Davies-Bouldin index \cite{arbelaitz2013extensive}.

\section*{Compliance with Ethical Standards}

\paragraph{Conflict of Interest}
Andrea Ceni, Peter Ashwin, and Lorenzo Livi declare that they have no conflict of interest.

\paragraph{Funding}
We gratefully acknowledge the support of NVIDIA Corporation with the donation of the Titan Xp GPU used for this research.
LL gratefully acknowledges partial support of the Canada Research Chairs program.
PA gratefully acknowledges partial support of the EPSRC Centre for Predictive Modelling in Healthcare via grant EP/N014391/1.

\paragraph{Human and Animal Rights}
This article does not contain any studies with human participants or animals performed by any of the authors.

\bibliographystyle{abbrvnat}
\bibliography{Bibliography.bib}

\end{document}